\documentclass[10pt,twocolumn,letterpaper]{article}

\usepackage[pagenumbers]{cvpr} 

\usepackage{subcaption}

\usepackage[dvipsnames]{xcolor}
\usepackage{acronym}
\usepackage[misc]{ifsym}
\usepackage[utf8]{inputenc} 
\usepackage[T1]{fontenc}    
\usepackage{url}            
\usepackage{booktabs}       
\usepackage{
    amsfonts,amsmath,amssymb,amsthm,mathbbol,mathtools
}       
\usepackage{amsmath,amssymb,mathbbol}
\usepackage{nicefrac}       
\usepackage{microtype}      
\usepackage{graphicx}

\usepackage{etoc}
\usepackage{lipsum}
\usepackage{adjustbox}
\usepackage{wrapfig}
\usepackage{multicol}
\usepackage{mdframed}
\usepackage{enumitem}
\usepackage{balance}
\usepackage{xspace}
\usepackage{setspace}
\usepackage[skip=3pt,font=small]{caption}
\usepackage{tabularx,colortbl,multirow,array,makecell}
\usepackage{pifont}
\usepackage{xr-hyper}
\usepackage{tcolorbox}
\tcbuselibrary{breakable}
\usepackage{makecell}
\usepackage{microtype}
\usepackage{adjustbox}
\newcommand{\cmark}{{\color{green!60!black}\ding{51}}} 
\newcommand{\xmark}{{\color{red!70!black}\ding{55}}}
\newcommand{\customfootnotetext}[2]{{
  \renewcommand{\thefootnote}{#1}
  \footnotetext[0]{#2}}}

\definecolor{cvprblue}{rgb}{0.21,0.49,0.74}
\usepackage[pagebackref,breaklinks,colorlinks,allcolors=cvprblue]{hyperref}

\def\paperID{2162} 
\def\confName{CVPR}
\def\confYear{2026}
\newcommand{\model}{{SAGE}\xspace}
\newcommand{\centercrop}[2][1.5]{%
  \pgfmathsetmacro{\clipfrac}{(1-1/#1)/2}%
  \adjustbox{scale=#1, Clip={\clipfrac\width} {\clipfrac\height} {\clipfrac\width} {\clipfrac\height}}{%
    #2%
  }%
}
\title{\model: Scalable Agentic 3D Scene Generation for Embodied AI}

\author{
Hongchi Xia$^{1,2}$\footnotemark[1]
\enspace
Xuan Li$^{1}$
\enspace
Zhaoshuo Li$^{1}$
\enspace
Qianli Ma$^{1}$
\enspace
Jiashu Xu$^{1}$
\enspace
Ming-Yu Liu$^{1}$
\enspace
Yin Cui$^{1}$
\\
Tsung-Yi Lin$^{1}$
\enspace
Wei-Chiu Ma$^{3}$
\enspace
Shenlong Wang$^{2}$
\enspace
Shuran Song$^{1,4}$
\enspace
Fangyin Wei$^{1}$
\vspace{2mm}\\
$^{1}$NVIDIA
\quad
$^{2}$University of Illinois Urbana-Champaign
\\
$^{3}$Cornell University
\quad
$^{4}$Stanford University
}

\begin{document}
\vspace{-10mm}
\twocolumn[{%
\renewcommand\twocolumn[1][]{#1}%
\maketitle
\vspace{-14mm}
\captionsetup{type=figure}
\begin{center}
\includegraphics[width=0.96\linewidth]{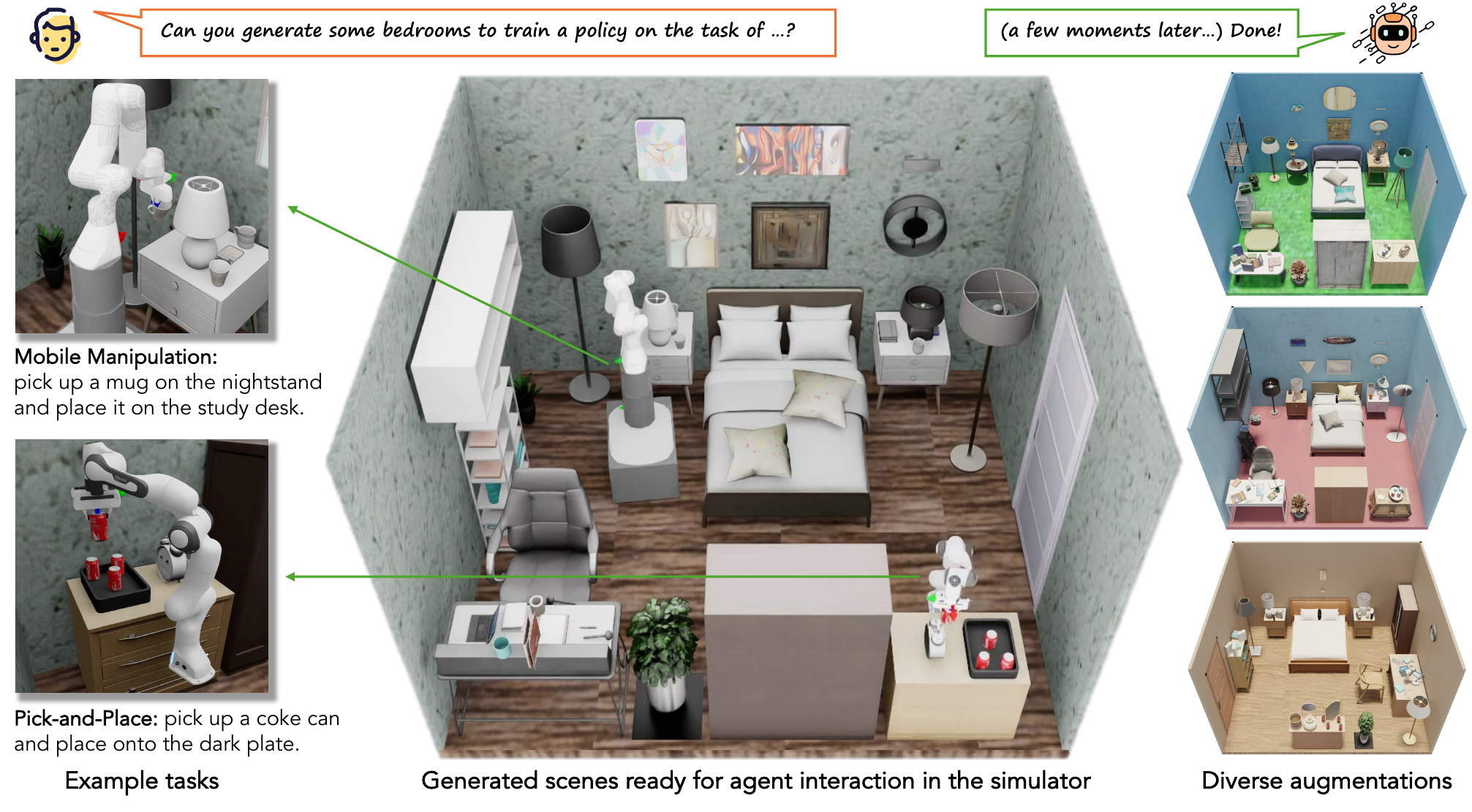}
    \hfill    
\end{center}
\vspace{-5mm}
\captionof{figure}{
\textbf{Overview and example outputs of \model}.
Given an open-ended user request, our system autonomously composes realistic, diverse, and simulation-ready 3D environments. The generated scenes are directly deployable in modern simulators, supporting embodied tasks such as Mobile Manipulation and Pick-and-Place. Through agent-driven reasoning, generator orchestration, and multi-level augmentation, the framework produces interactive environments at scale for robot policy learning.
}
\label{fig:teaser}
}
\vspace{3mm}
]

\customfootnotetext{*}{This work was done while Hongchi Xia was an intern at NVIDIA.}

\begin{abstract}
Real-world data collection for embodied agents remains costly and unsafe, calling for scalable, realistic, and simulator-ready 3D environments. However, existing scene-generation systems often rely on rule-based or task-specific pipelines, yielding artifacts and physically invalid scenes.

We present \model, an agentic framework that, given a user-specified embodied task (\eg, “pick up a bowl and place it on the table”), understands the intent and automatically generates simulation-ready environments at scale. The agent couples multiple generators for layout and object composition with critics that evaluate semantic plausibility, visual realism, and physical stability. Through iterative reasoning and adaptive tool selection, it self-refines the scenes until meeting user intent and physical validity.

The resulting environments are realistic, diverse, and directly deployable in modern simulators for policy training. Policies trained purely on this data exhibit clear scaling trends and generalize to unseen objects and layouts, demonstrating the promise of simulation-driven scaling for embodied AI.
Code, demos, and the SAGE-10k dataset can be found on the project page \href{https://research.nvidia.com/labs/dir/sage/}{here}.
\end{abstract}    
\section{Introduction}

\begin{table*}[h]
\centering
\setlength{\tabcolsep}{3pt}

\resizebox{0.99\textwidth}{!}{
\begin{tabular}{clcccccccccccc}
\toprule
 \multirow{3}{*}{Framework} & \multirow{3}{*}{Method} & \multicolumn{2}{c}{Input} & \multicolumn{4}{c}{Methodology} & \multicolumn{3}{c}{Simulation-Ready Output}  \\
\cmidrule(r){3-4} \cmidrule(r){5-8}\cmidrule(r){9-11}
  &  &  Open- &  \multirow{2}{*}{Modality} & Built-in Scene  & Grounded & Self-& Fine-grained  &Collision& w/ Physical &Simulator-  \\
&  &  Vocabulary &  &Generation &in 3D & Improvement&Control  & Check& Attributes&Validated  \\
\midrule
Rule-based & Infinigen~\cite{infinigen2024indoors}, ProcTHOR~\cite{procthor} &\xmark&Text&\cmark&\cmark& \xmark&\xmark&\xmark&\xmark&\xmark\\
\midrule
\multirow{3}{*}{\shortstack{3D Data-\\Driven}}  & \small{DiffuScene~\cite{tang2024diffuscene}, Text2Room~\cite{hollein2023text2room}}&\multirow{3}{*}{\xmark}&Text / 3D Scene&\multirow{3}{*}{\cmark}&\multirow{3}{*}{\cmark}&\multirow{3}{*}{\xmark}&\multirow{3}{*}{\xmark}&\multirow{3}{*}{\xmark}&\multirow{3}{*}{\xmark}&\multirow{3}{*}{\xmark}\\
& \small{ATISS~\cite{paschalidou2021atiss}, PhyScene~\cite{yang2024physcene}}  &&Floor Plan&&\\
& \small{CommonScenes~\cite{zhai2023commonscenes}, EchoScene~\cite{zhai2024echoscene}}  &&Scene Graph&&\\
\midrule
LLM/VFMs & \small{LayoutGPT~\cite{feng2023layoutgpt}, LayoutVLM~\cite{sun2025layoutvlm}, I-Design~\cite{ccelen2024design}} &\multirow{3}{*}{\cmark}&\multirow{3}{*}{Text}&\multirow{3}{*}{\cmark} & \xmark&\multirow{2}{*}{\xmark}&\multirow{3}{*}{\xmark}&\multirow{2}{*}{\xmark}&\multirow{3}{*}{\xmark}&\multirow{3}{*}{\xmark}\\
Static & \small{Holodeck~\cite{yang2024holodeck}, ArtiScene~\cite{gu2025artiscene}, Architect~\cite{wang2024architect}}& &&&\multirow{2}{*}{\cmark}&&&\\
Pipeline & \small{AnyHome~\cite{fu2024anyhome}, Scenethesis~\cite{ling2025scenethesis}, GenUSD~\cite{lin2024genusd}}& &&&&\xmark\kern-0.67em
\cmark (pose-only)&&\cmark\\
\midrule
Agent- & SceneWeaver~\cite{yang2025sceneweaver} &\multirow{2}{*}{\cmark}&  Text&\xmark&\multirow{2}{*}{\cmark}&\multirow{2}{*}{\cmark}&\multirow{2}{*}{\cmark}&\multirow{2}{*}{\cmark}&\xmark&\xmark\\
based& \model (Ours) & &Text (+Image)&\cmark&&&&&\cmark&\cmark\\
\bottomrule
\end{tabular}
}
\caption{\textbf{Comparison of scene generation methods.} Our agent-based method uniquely fulfills all criteria, enabling scalable generation of simulator-validated data essential for robotic applications.  }
\label{tab:capability_comparison}
\end{table*}

Embodied AI is on the hunt for data. Unfortunately, while the web has powered large vision and language models \cite{radford2021learningtransferablevisualmodels, brown2020languagemodelsfewshotlearners}, neither the web nor the real world can resolve the pressing data needs of embodied agents. 
Real-world embodied data collection is slow and costly \cite{wang2025embodiedreameradvancingreal2sim2realtransfer}, and is fundamentally constrained by the need for {\it interactive environments} \cite{shen2021igibson}. 
Simulation thus emerges as the natural alternative, providing scalable, interactive, and low-cost embodied environments with safety guarantees \cite{isaaclab}.

To effectively support embodied AI, simulated data must satisfy four desiderata: 
(i) {\bf Realism}: the geometry, appearance, semantic structure, and physics of the simulation should resemble the real world closely enough that policies learned in simulation can reliably transfer to reality; 
(ii) {\bf Diversity}: simulation should encompass a wide range of assets, environments, and tasks to prevent overfitting and support generalization;
(iii) {\bf Simulation-readiness}: objects and scenes must be physically stable, interactable, and directly compatible with modern simulators for large-scale training; (iv) {\bf Task-awareness}: simulation environments should adapt to and facilitate the training of various targeted embodied tasks. For instance, a data engine should provide various kitchens for household robots, fire-hazard scenes for rescue robots, surgical rooms for medical robots, \etc.

Unfortunately, existing methods for generating 3D simulated data fall short on one or more of these criteria. 
While Real2Sim approaches~\cite{drawer, video2game, holoscene, phystwin} prioritize realism by reconstructing digital twins from the real world, the high costs of data capture and reconstruction make them difficult to scale. In contrast, generative approaches offer a more scalable alternative, but still face significant bottlenecks.
For example, rule-based systems~\cite{infinigen2024indoors,procthor} ensure physical plausibility but sacrifice flexibility and diversity. 
Data-driven approaches~\cite{paschalidou2021atiss,tang2024diffuscene,hollein2023text2room} improve realism, but their limited 3D training data prevents them from generalizing to new room types, handling open-vocabulary prompts, or supporting fine-grained layout control. 
Pipelines based on foundation models ~\cite{feng2023layoutgpt,sun2025layoutvlm,yang2024holodeck} enable text-driven generation but lack 3D grounding, yielding physically invalid scenes. 
Perhaps most importantly, these systems are \emph{static}: their ``computational graph'' is fixed, preventing adaptive reasoning and self-correction.
Recently, a concurrent work, SceneWeaver~\cite{yang2025sceneweaver}, takes a step toward agentic scene generation.
Yet, it is not directly deployable in simulation due to missing physical attributes and the absence of simulator-in-the-loop verification.
There are still major gaps in (1) physical grounding for interaction and (2) compatibility with robot simulators, as shown in Tab.~\ref{tab:capability_comparison}.

In this work, we present \model, a novel agentic framework for scalable 3D scene generation that produces simulation-ready environments directly from arbitrary user prompts. The agent, operating over Model Context Protocal (MCP)~\cite{model_context_protocol}, adaptively orchestrates generators for floor plans, structured layouts, and text-to-3D assets, while two complementary critics provide continuous feedback: a visual critic for semantic/spatial coherence and a physics critic with simulator-in-the-loop validation (Isaac Sim~\cite{isaacsim}) for stability under gravity and collisions. This closed loop enables self-correction without hard-coded tool order and yields scenes that are both realistic and interactable.

We scale these scenes for embodied AI by applying multi-level augmentation (object configuration, object category, and layout) to generate diverse yet task-consistent environments. On top of these scenes, we automatically synthesize action data with grasp pose proposals \cite{yuan2023m2t2}, collision-aware Inverse Kinematics (IK) \cite{curobo}, and navigation planning, and train a Diffusion Policy \cite{chi2024diffusionpolicyvisuomotorpolicy}. Experiments show clear scaling trends with increasing scene diversity and demonstration count, and improved generalization to unseen objects and layouts.

In summary, our contributions are twofold:

\begin{itemize}
    \item 
    \textbf{An agentic scene-generation framework} that unifies multiple generators with visual and physics critics under MCP, enabling adaptive tool use, self-improvement, and simulator-validated stability. It generates open-vocabulary scenes with state-of-the-art realism and physical validity.
    \item 
    \textbf{An embodied AI data pipeline} that automatically generates unlimited scenes and teacher demonstrations for imitation learning, yielding clear scaling benefits and stronger generalization to unseen environments.
\end{itemize}

\begin{figure*}[t]
    \centering    
    \includegraphics[width=0.98\textwidth,trim={0 0 0 0}, clip]{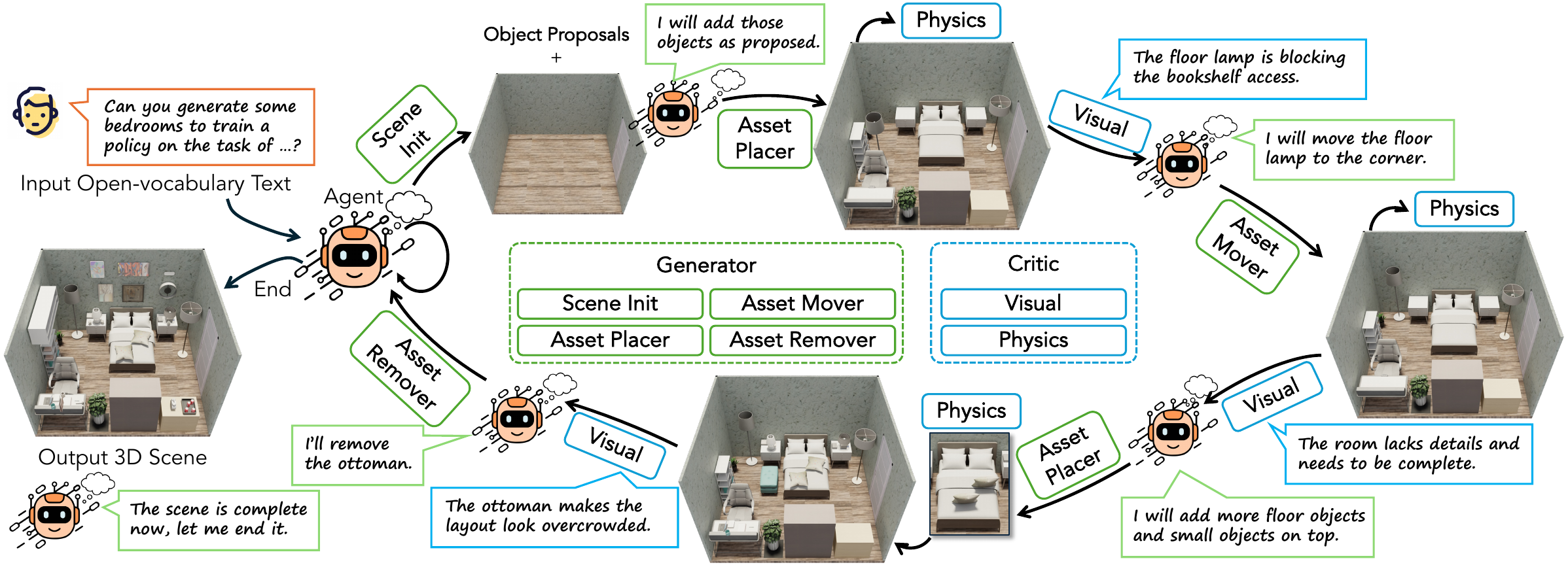}
    \caption{
    \textbf{Overview of \model scene generation.} Our system converts open-vocabulary text prompts into simulation-ready 3D scenes by orchestrating multiple generator tools and critics. The agent dynamically calls generators (Scene Init, Asset Placer/Mover/Remover) to construct and refine layouts, while visual and physics critics provide iterative feedback for self-improvement. The visual critic suggests semantic corrections (\eg, missing or misplaced objects), and the physics critic validates stability via Isaac Sim. For example, after applying physics critic in the bottom image, the newly added pillows on the bed fall flat. 
    This self-improvement process ends when the agent considers that the generated scene meets the input user requirements.
    The resulting scenes can be further scaled via augmentation and used for embodied policy learning.
    }
    \label{fig:overview}
    \vspace{-5mm}
\end{figure*}
\section{Related Works}
\paragraph{3D Indoor Scene Synthesis} Early work synthesized indoor scenes with procedural systems such as ProcTHOR~\cite{procthor} and Infinigen Indoors~\cite{infinigen2024indoors}, assembling layouts via rules and grammars. This paradigm grounds geometry and scales well but fixed recipes constrain open vocabulary, fine-grained control, and self-improvement, and rarely expose physics-rich attributes beyond hand-coded rules. Data-driven methods such as ATISS~\cite{paschalidou2021atiss} and DiffuScene~\cite{tang2024diffuscene}, along with graph-aware CommonScenes~\cite{zhai2023commonscenes} and EchoScene~\cite{zhai2024echoscene}, learn strong spatial priors and yield realistic layouts with neural 3D generators; nevertheless they inherit closed taxonomies, offer limited object-level levers, and do not attach explicit physical properties or run simulator-side validation. Motivated by advances in long-horizon reasoning for LLM agents~\cite{weiemergent,Yao2022ReActSR,gao2025survey}, recent methods shift to language- or vision-assisted scene synthesis. Holodeck~\cite{yang2024holodeck} and ArtiScene~\cite{gu2025artiscene} leverage LLMs or image intermediaries to widen semantic coverage and stylistic control; however, they usually follow static, pre-ordered modules, provide little self-improvement, have sparse 3D grounding, and omit physics validation such as collision or stability. Some works~\cite{fu2024anyhome, ling2025scenethesis} combine LLM planning with vision-guided refinement under constraints, but the refinement is only on asset pose. Building further on agent capabilities for iterative decision making and self-reflection~\cite{shinn2023reflexion,huang2023large}, agentic pipelines~\cite{lin2024genusd, yang2025sceneweaver} generate complete layouts and refine them through feedback. SceneWeaver~\cite{yang2025sceneweaver} further leverages tool use for relation/collision-aware placement. Despite these advances, orchestration often outweighs end-to-end simulation deliverables: physical attributes are not systematically attached and simulator-validated outputs are not the default artifact. Our method \model pushes from ``semantically plausible'' to task-aligned, simulation-ready: we support open-vocabulary prompts (with optional images), generate scenes natively, attach object-level physical properties, and perform in-simulator validation (collision and stability), so outputs are directly deployable for embodied policy training.
\vspace{-3mm}
\paragraph{Simulation Environment for Embodied AI}

In recent years, numerous simulation benchmark environments have been developed to provide safe, scalable, and controllable platforms for evaluating robot learning algorithms before real-world deployment.
LIBERO \cite{liu2023libero}, CALVIN \cite{mees2022calvin} and SimplerEnv \cite{li2024evaluating} focus on tabletop manipulation tasks such as pick-and-place and object rearrangement using simple robotic arm setups. 
iGibson \cite{li2021igibson}, Habitat \cite{szot2021habitat}, BEHAVIOR-1K \cite{li2023behavior}, ManiSkill \cite{tao2024maniskill3}, RoboCasa \cite{nasiriany2024robocasa}, AI2Thor \cite{kolve2017ai2}, VirtualHome \cite{puig2018virtualhome}, ThreeDWorld \cite{gan2020threedworld} extend the scope to home-scale embodied activities that combine object interaction and motion planning within complex 3D indoor scenes. 
Some recent works also aim to unify and scale up embodied AI benchmarking. RoboGen \cite{wang2023robogen} combines automatic object generation and task generation. RoboVerse \cite{geng2025roboverse} integrates multiple simulators and benchmarks to support seamless transitions across environments.  
Underlying these platforms are diverse physics engines such as PyBullet \cite{coumans2016pybullet}, MuJoCo \cite{todorov2012mujoco}, Isaac Sim \cite{isaacsim}, SAPIEN \cite{xiang2020sapien}, Genesis \cite{Genesis}, which are specifically developed for robotics simulation, and Unity \cite{unity}, Unreal \cite{unrealengine}, which are designed for game developments.
While many of these frameworks provide rich, realistic, and ready-to-use environments for embodied AI, they often require substantial manual effort for scene and asset creation. Others rely on procedurally retrieved assets, which can limit diversity when scaling up. And the simulation is often only enabled when the creation is complete. In contrast, \model integrates 3D generation frameworks, enabling open-vocabulary asset creation with high flexibility and minimal manual intervention. Furthermore, by incorporating simulation directly into the generation loop, \model supports iterative self correction and improvement.
\section{Method}

Given a user demand with a robot task description, our goal is to generate diverse 3D scenes ready to run embodied agents in simulation for scalable policy learning.
At the core of \model is an agent-driven scene generation framework, detailed in Sec.~\ref{sec:scene_generation}. Based on the feedback from visual and physical \textit{critics}, the agent effectively self-improves the generation with multiple editing operations supported by scene \textit{generator} tools.
In Sec.~\ref{sec:policy_learning}, we describe how scene generation can be easily scaled up with object-level and scene-level augmentation, which can then be used for action generation to train embodied AI policy.

\subsection{Agent-driven Scene Generation}
\label{sec:scene_generation}
\model operates under the Model Context Protocol (MCP), a standardized protocol for seamless interaction with external tools~\cite{model_context_protocol}. In this setup, the agent acts as the MCP client, while each tool (\eg, layout generator, physics simulator) is hosted behind an MCP server. 
The agent takes a chain of actions to improve the scene until the scene is considered visually realistic and physically stable.
In each iteration, when the agent identifies the need for a specific capability, \eg, generating a floor plan or validating physical stability, it sends a structured request through MCP. The server executes the corresponding operation, returns the result, which the agent incorporates as feedback to decide the next action. This setup enables adaptive, tool-driven scene generation without hard-coded logic.
Fig.~\ref{fig:overview} summarizes our approach.

\subsubsection{Generator}
The scene is constructed through a set of generator tools that the agent dynamically invokes via MCP. Each tool performs a specific operation -- initializing layouts, adding new assets, or adjusting existing ones -- and can be flexibly composed based on the agent’s reasoning and critic feedback. This allows for iterative, adaptive scene construction with fine-grained control over content and structure.
\vspace{-3mm}
\paragraph{Scene Initializer} 
takes the scene specification as input and is responsible for generating an empty 3D room (with only floor and walls). It also outputs a list of proposed objects, each with a text description, estimated physical attributes, and placement constraints (\eg, relationship to other objects, boundaries). 
The textures of floor and wall are generated by MatFuse~\cite{matfuse}, with a size assigned from the LLM prediction.
The object list reflects the scene demands or robot task descriptions. For example, if the user asks about learning a task of ``pick an apple and place it to a bowl'', the generator will include the required apple and bowl in the object list to be returned to the agent.
\vspace{-3mm}
\paragraph{Asset Placer} 
takes in a text string describing the placement requirement for a few objects. Its task is to generate and place these objects into the 3D scene. 
We generate objects with TRELLIS~\cite{trellis} given a text description. To make the object simulation-ready, we also leverage a VLM to estimate physical properties, including height to rescale the unit-sized object, mass for physical simulation, and metallic/roughness values for physics-based rendering (PBR).
For each object, we use an LLM to analyze the input placement condition and choose one of the three categories: floor, wall, and on-top. This further guides the placement sequence and constraints of each placement. We adopt depth first search with collision avoidance to place the objects to possible location candidates following the sequence, and choose the placement the best satisfies the constraints.
\vspace{-3mm}
\paragraph{Asset Mover} 
locates and moves the object specified in the input text string. This is achieved by first removing the object and then reusing the placement planner from the Asset Placer to reloate the object. Movement instructions typically come from the visual critic.
\vspace{-3mm}
\paragraph{Asset Remover} 
removes the object described in the input text, using LLM-based reasoning to locate the object in the scene. It is usually called when the critics (described later) gives feedback to remove an object.

\subsubsection{Critic for Self-Improvement}
Naively stacking different generators causes error accumulation from individual imperfections. 
Errors generally fall into two types: visual artifacts (\eg, missing or misplaced objects) and physical violations (\eg, instability or collisions). To mitigate this, we introduce a visual critic that assesses semantic and spatial coherence, as well as a physics critic that enforces stability and simulator readiness. Together, they guide iterative self-correction high-quality scene generation.
\vspace{-5mm}
\paragraph{Visual Critic}
To evaluate and improve the layout quality and scene completeness, we introduce a visual critic. The critic takes the current scene configuration as input, \ie, object placements and multi-view renderings (top-down and four corner views), and proposes new objects to place as well as adjustments to existing placements. Its feedback is then incorporated to aid the agent’s decision-making of the most appropriate generator to invoke next. By integrating these visual feedback signals, the agent gains a more comprehensive understanding of the current scene state and can decide which generator to invoke next for optimal refinement.

\paragraph{Physics Critic}
Beyond visual realism, a generated scene must also be physically stable and simulation-ready to support embodied learning. To achieve this, we employ simulation-in-the-loop validation during generation. After each object addition, movement, or removal. the scene is loaded into Isaac Sim~\cite{isaacsim} to test its physical stability. We measure the pose change after simulation and reject placements that cause instability or collisions, retaining only candidates that preserve global stability. If the generator fails to find a stable configuration, the critic reports the failure to the agent, suggesting alternative actions such as using smaller objects or adjusting placement target locations.
Through this iterative validation loop, \model maintains near-perfect physical stability, ensuring that the resulting scenes are directly deployable for downstream embodied learning tasks.

\subsection{Scaling the Scene for Embodied AI}
\label{sec:policy_learning}
To train a robust and generalizable policy, it is not sufficient to use a single generated scene; instead, we must generate a diverse set of simulation-ready scenes that follow the user’s task specification.
To this end, we introduce a scene augmentation strategy that systematically expands each generated environment into numerous variants while preserving task semantics.
These augmented scenes are then used to generate corresponding robot actions via motion planning, followed by imitation learning to train task-specific embodied policies.

\subsubsection{Scene Augmentation}
To scale one generated scene into a diverse yet task-consistent set of variants, we apply (1) task-relevant object augmentation (configuration, category) that diversifies key objects while preserving task semantics, and (2) task-irrelevant scene augmentation to enrich the rest of the scenes.
\vspace{-3mm}
\paragraph{Object Configuration-level} 
The pose of each task-relevant object (\eg, the target to be picked, placed on, or navigated toward) is resampled within the current scene to create variations in object placement. This process increases spatial diversity while preserving overall scene semantics.
\vspace{-3mm}
\paragraph{Object Category-level} 
Given the text description of each task-relevant object from the generation stage, we employ an LLM-based text augmentation to produce variations in geometry and texture (\eg, shape, color, material, or finish) while maintaining the original object category.
We then use TRELLIS~\cite{trellis} to synthesize corresponding 3D assets from these augmented descriptions, which are placed into the scene to enrich visual and physical diversity across instances.
\vspace{-3mm}
\paragraph{Scene Layout-level} 
While the above augmentations modify task-related objects, the background environment remains unchanged.
For tasks requiring full-scene exploration or navigation, we introduce layout-level augmentation, where the background scene, including room geometry and all task-irrelevant objects, is regenerated through the agent-driven scene generation.
This process produces diverse scene layouts sharing the same task specification, enabling learning policies that generalize across spatial configurations.
\vspace{-3mm}
\paragraph{Simulation-ready Validation} 
After each augmentation step, we call the physics critic to ensure the stability and physical plausibility of all placements.
Object-level physical properties such as mass and PBR parameters are estimated by a VLM as before.
This guarantees that every augmented scene remains physically valid and immediately deployable for policy training.

\begin{table*}[t]
\centering
\resizebox{\textwidth}{!}{
\setlength{\tabcolsep}{0pt}
\begin{tabular}{@{}cccccc@{}}
\rotatebox{90}{\quad Holodeck~\cite{yang2024holodeck}} &
\includegraphics[width=0.25\linewidth,trim=110 20 110 0, clip]{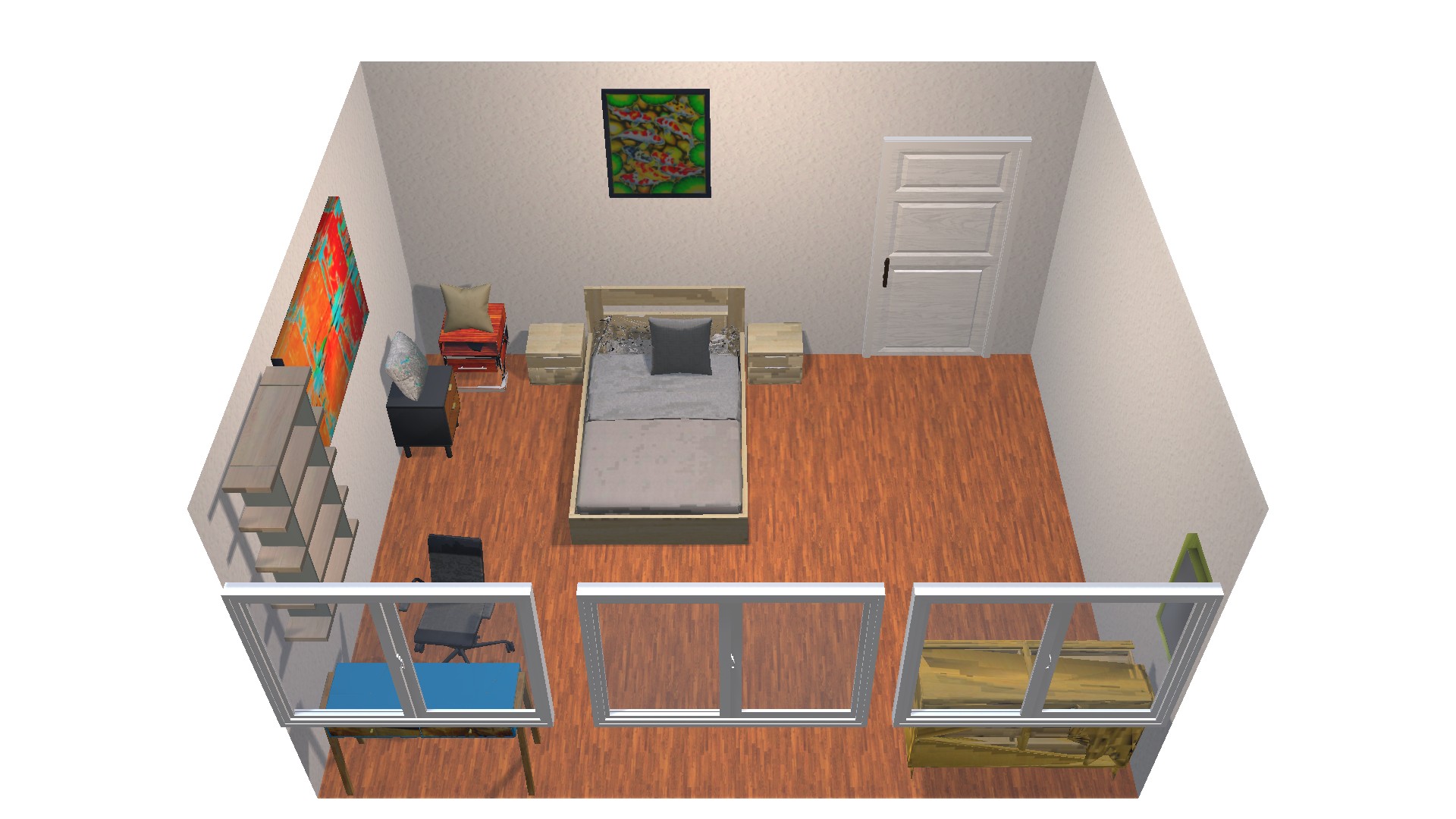} &
\includegraphics[width=0.25\linewidth,trim=110 20 110 0, clip]{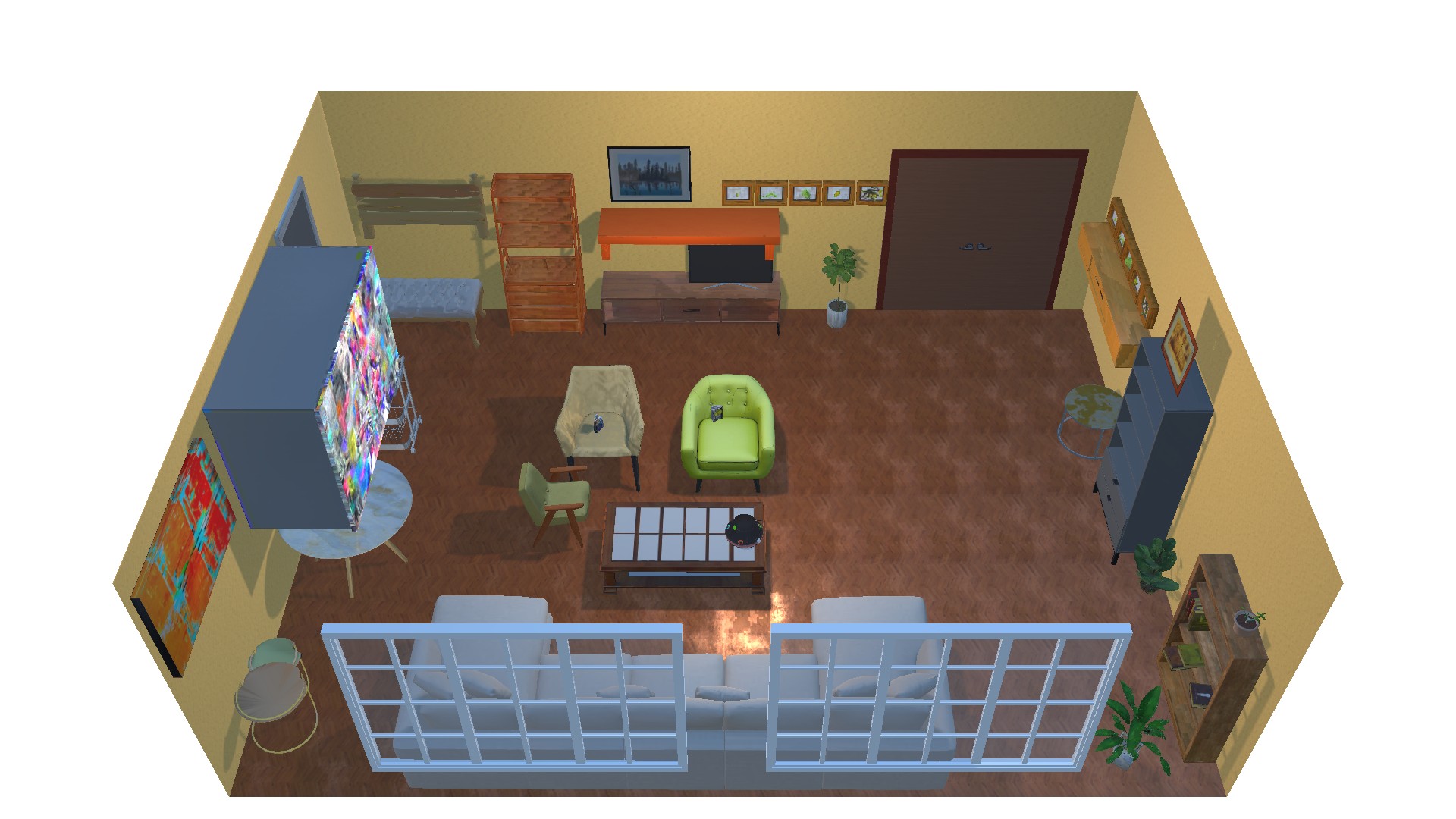} &
\includegraphics[width=0.25\linewidth,trim=100 20 100 10, clip]{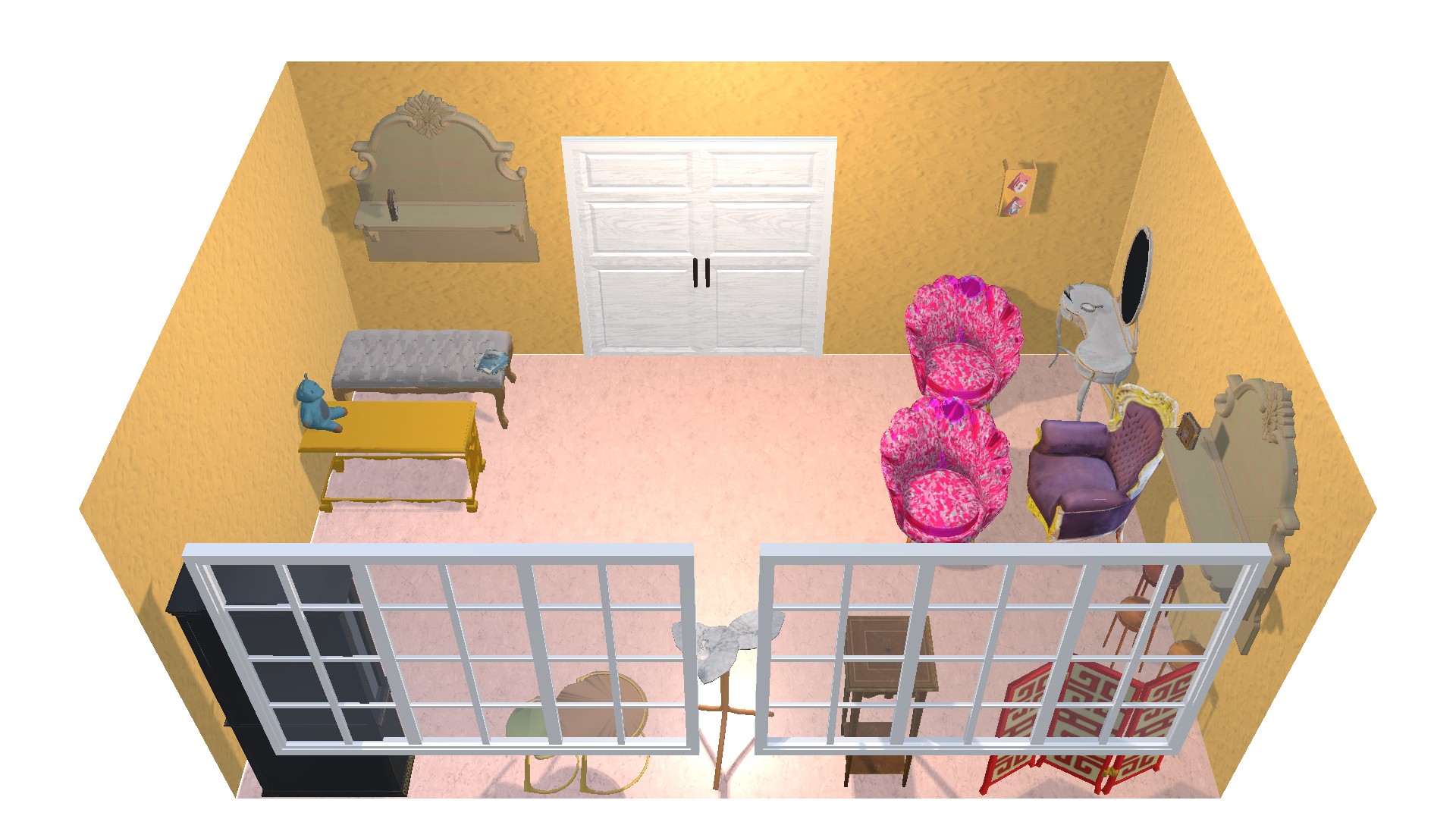} &
\includegraphics[width=0.25\linewidth,trim=100 20 100 40, clip]{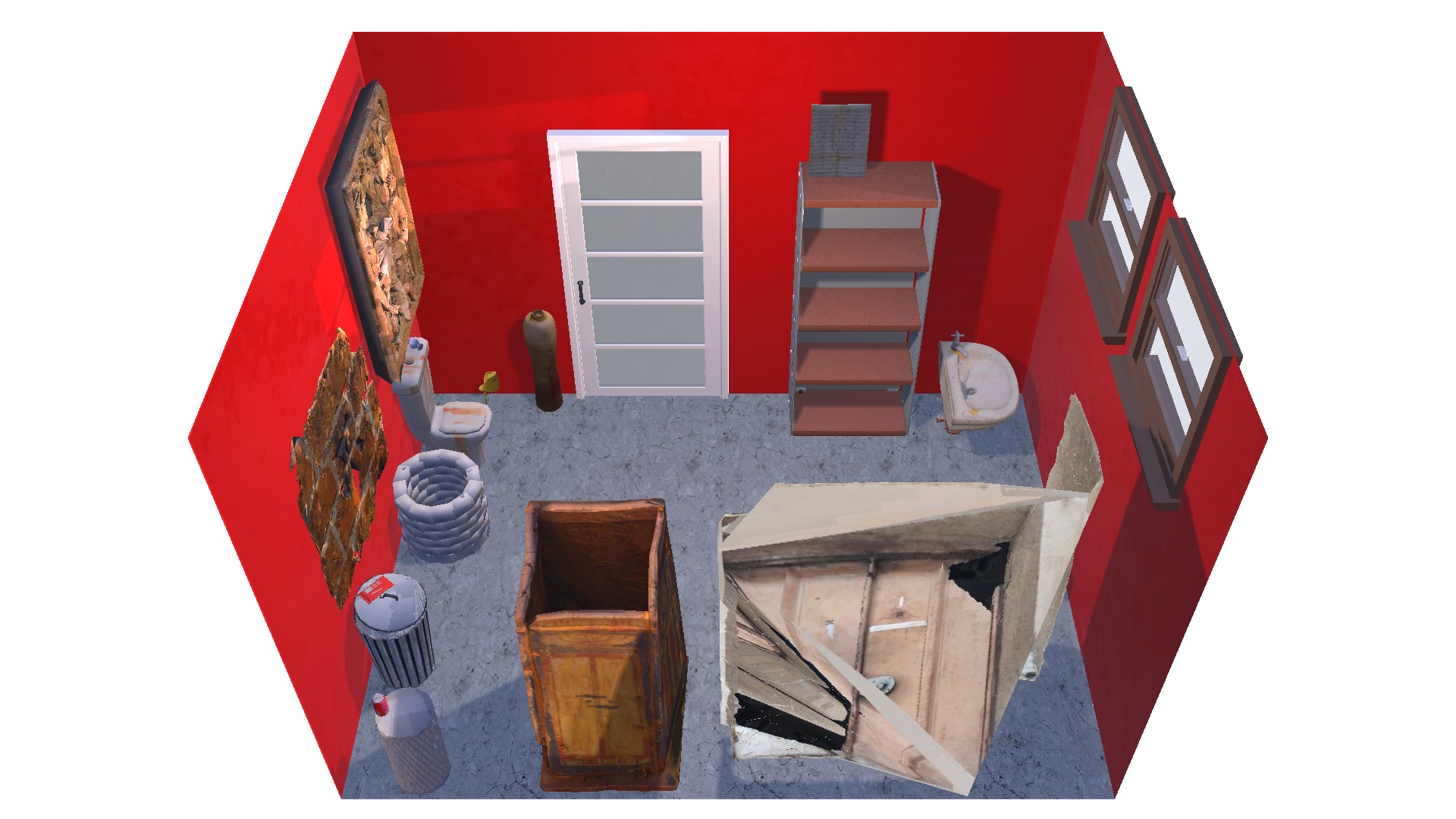} 
\\[1.1mm]
\rotatebox{90}{\enspace SceneWeaver~\cite{yang2025sceneweaver}} &
\includegraphics[width=0.25\linewidth,trim=110 20 110 20, clip]{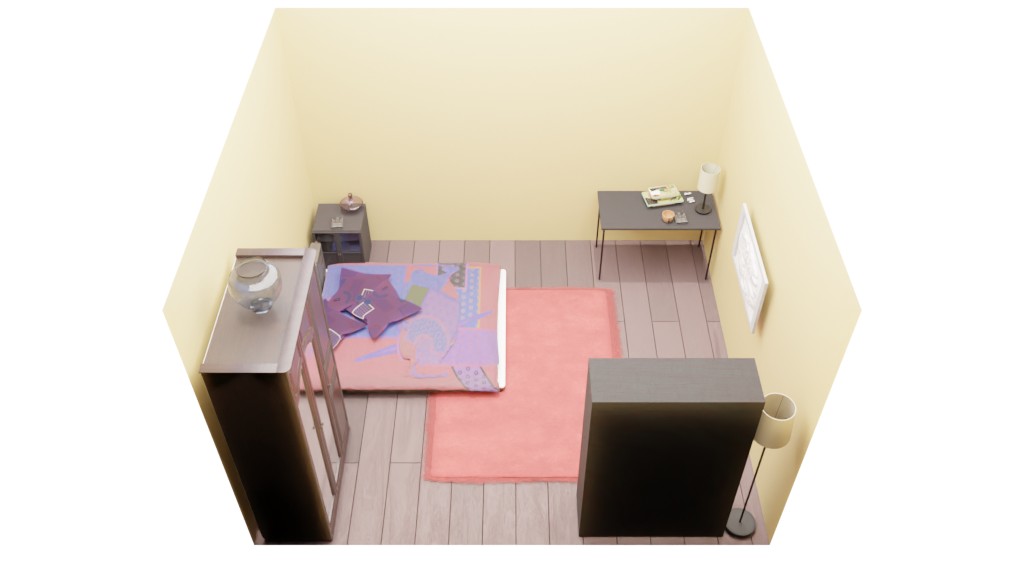} &
\includegraphics[width=0.25\linewidth,trim=110 20 110 20, clip]{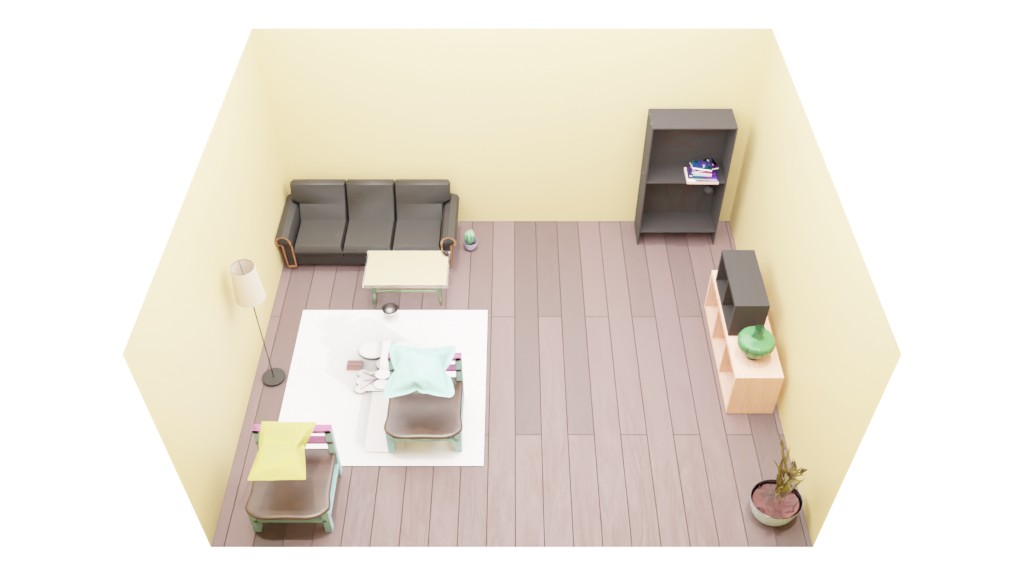} &
\includegraphics[width=0.25\linewidth,trim=110 20 110 20, clip]{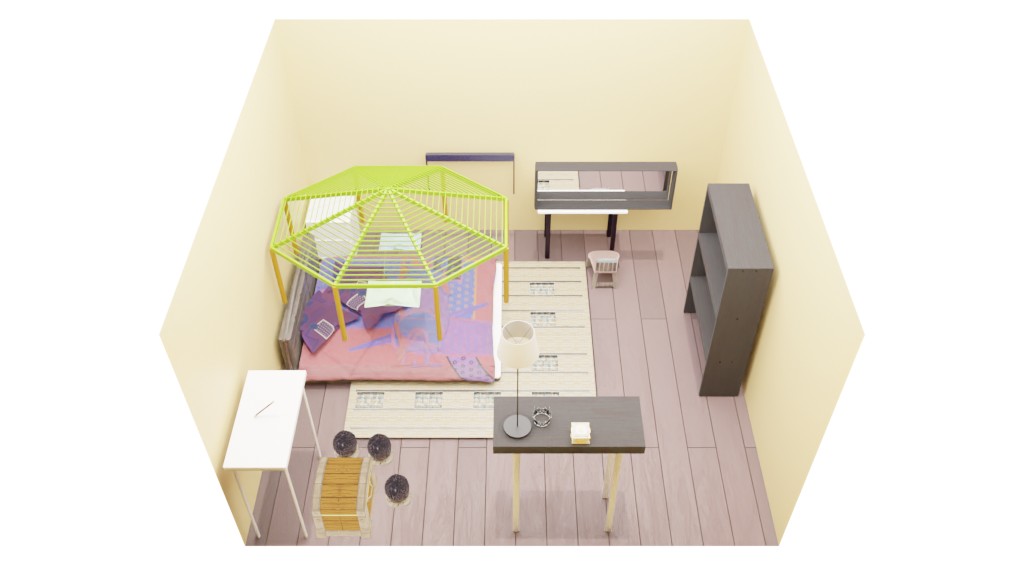} &
\includegraphics[width=0.25\linewidth,trim=110 20 110 20, clip]{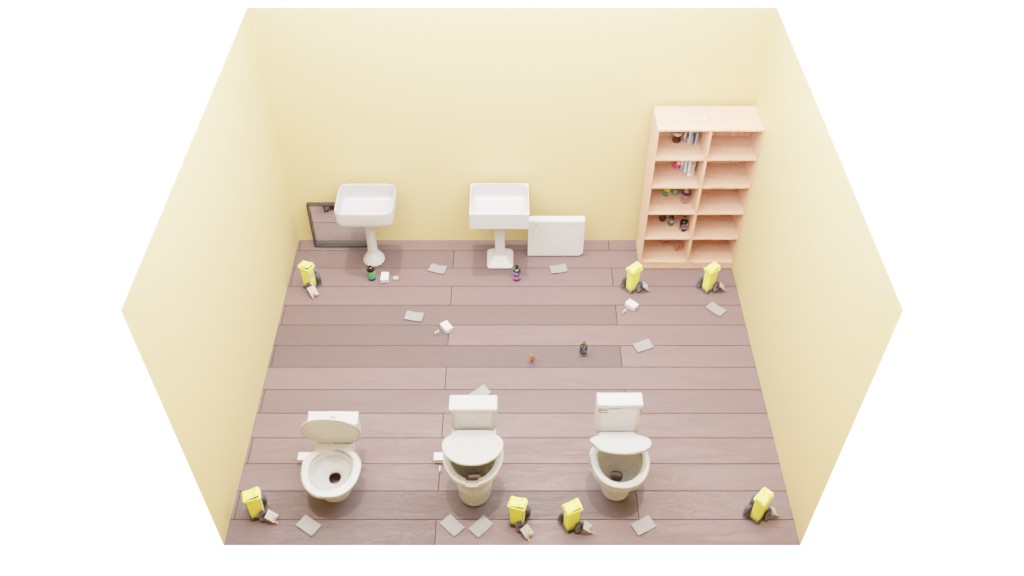} 
\\
\rotatebox{90}{ \quad \model (Ours)} & 
\includegraphics[width=0.25\linewidth,trim=200 20 200 40, clip]{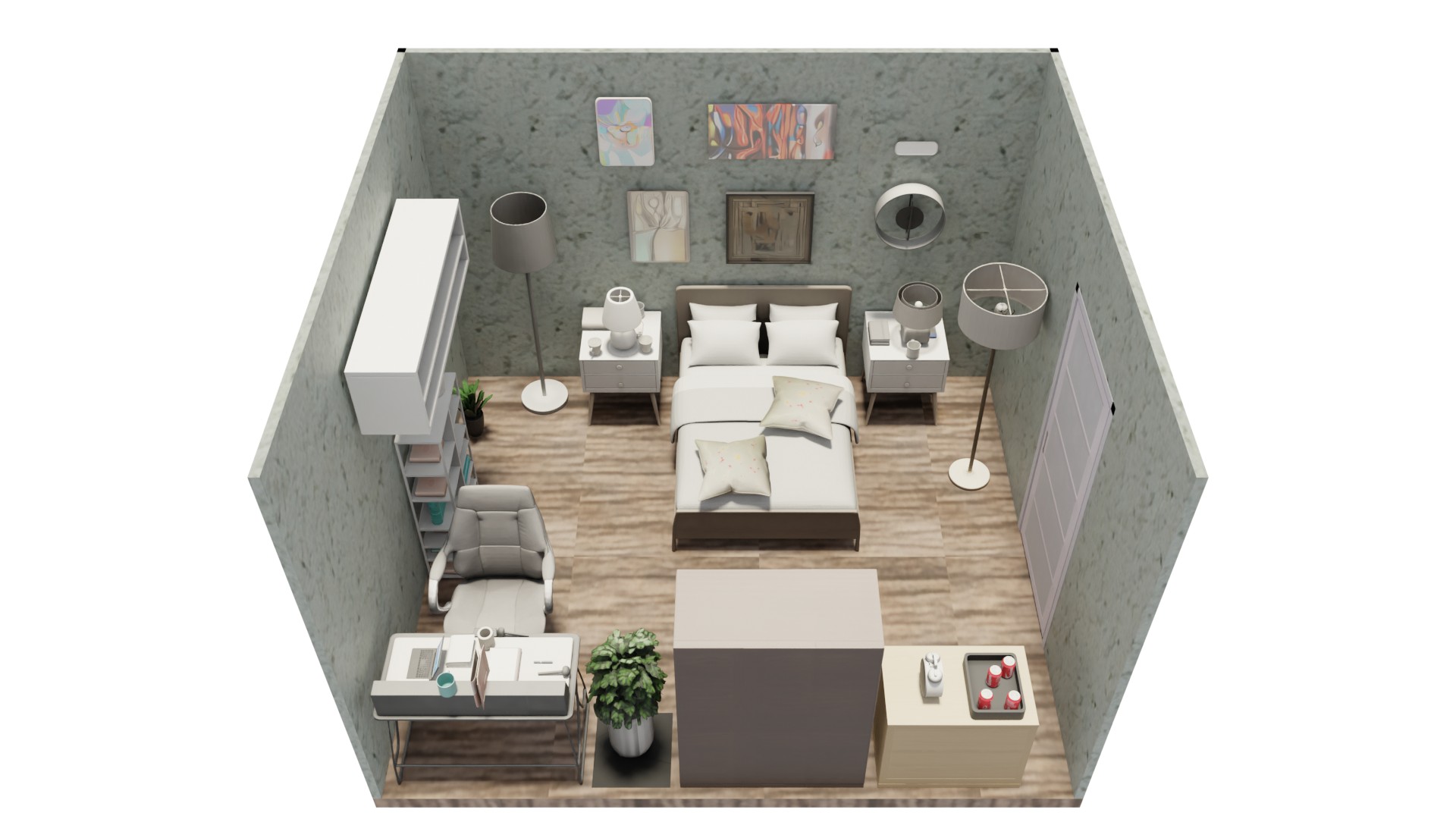} &
\includegraphics[width=0.25\linewidth,trim=200 20 200 40, clip]{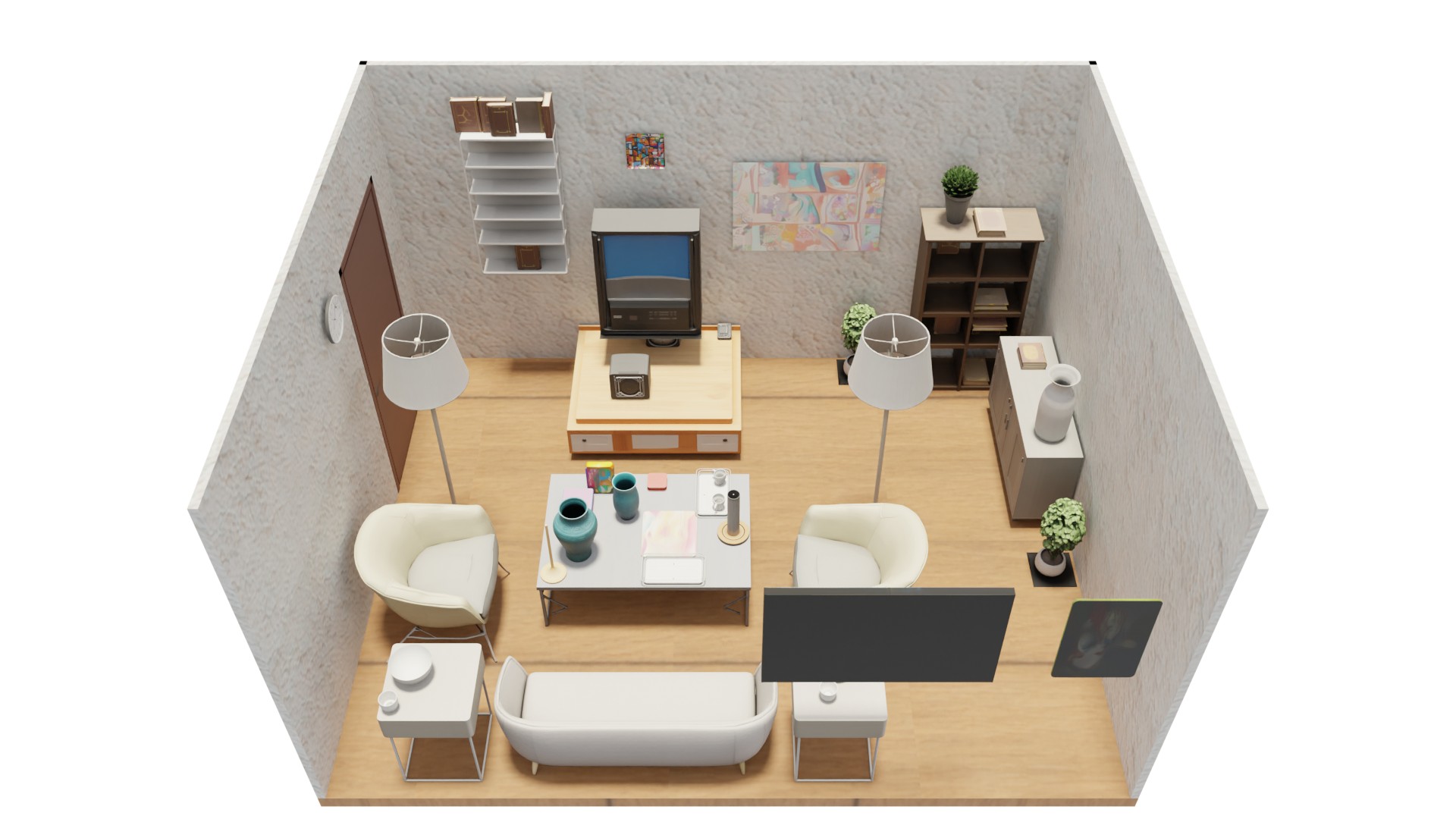} &
\includegraphics[width=0.25\linewidth,trim=200 20 200 40, clip]{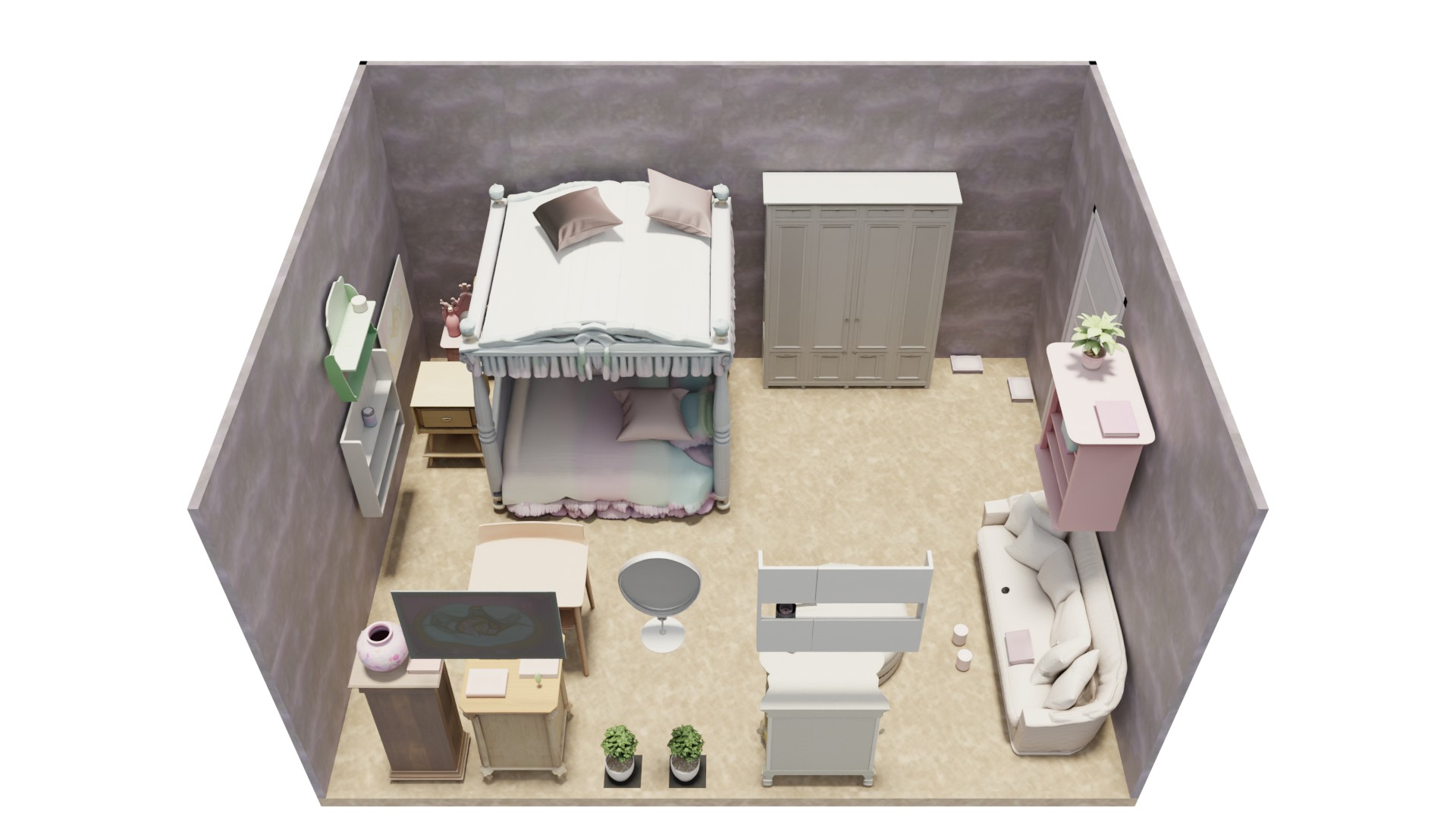} &
\includegraphics[width=0.25\linewidth,trim=190 20 190 40, clip]{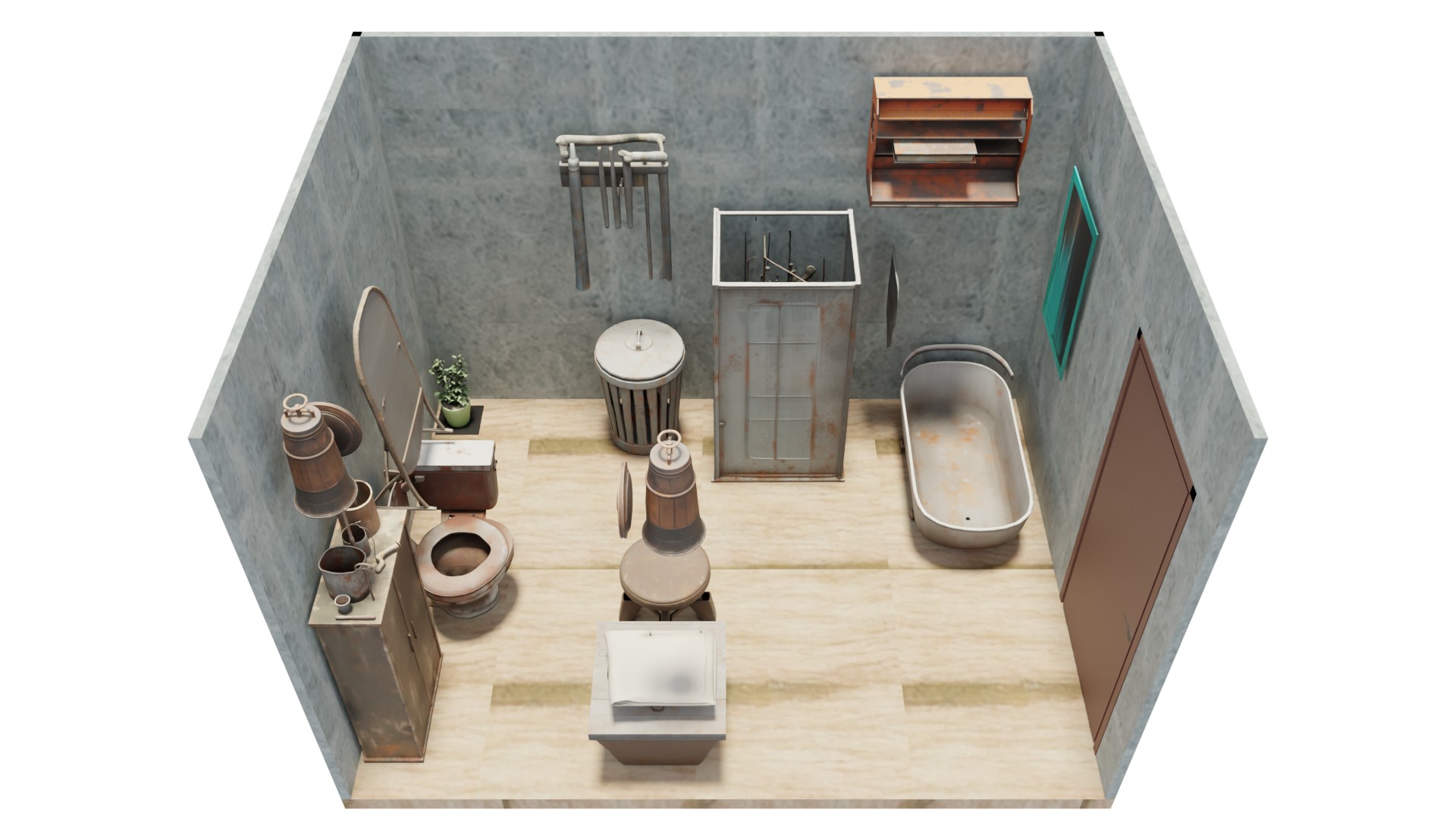} 
\\

&``Bedroom''&``Living room''&``Fairy-tale princess room''&``Rusty and dusty restroom''\\
\end{tabular}
}
\captionof{figure}{\textbf{Common and open-vocabulary scene generation comparison.} Compared with baselines, \model produces more complete scenes with more realistic layouts on common room types, while following the style prompts more faithfully on open-vocabulary queries.}
\label{fig:qual_cmp}
\end{table*}

\subsubsection{Action Generation}
~\label{sec:action}
The generated environments are now ready for downstream embodied AI tasks. We demonstrate their utility on two representative tasks, \ie, Pick-and-Place and Mobile Manipulation, and automatically generate large-scale action demonstrations using established motion planning techniques. This enables scalable policy learning and highlights the benefits of simulation-ready scene generation.
\vspace{-3mm}
\paragraph{Pick-and-Place} 
For grasping actions, we use M2T2~\cite{yuan2023m2t2} to generate grasp pose candidates from rendered depth images.
Collision-free trajectories are computed by integrating Curobo~\cite{curobo} into the motion planning and inverse kinematics pipeline, ensuring feasible and stable grasp execution.
\vspace{-3mm}
\paragraph{Mobile Manipulation} This task is composed of navigation with object pick-and-place in between.
For the navigation motions, we adopt RRT~\cite{rrt} for robot path planning, generating collision-free trajectories between designated start and target positions.
\vspace{-3mm}
\paragraph{Failure Modes} Motion planning is not always successful, often due to inaccurate grasp pose predictions, unreachable target configurations, or unexpected collisions during trajectory execution. We filter out failed examples by performing collision checks and verifying whether each manipulated object reaches its expected location.

\subsubsection{Policy Learning}
Given the large-scale action data generated from motion planning, we employ imitation learning to train generalizable policies from these demonstrations.
Specifically, we use Diffusion Policy~\cite{chi2024diffusionpolicyvisuomotorpolicy} for policy learning.
The model takes as input RGB and depth images from multiple camera views along with corresponding end-effector trajectories, and outputs continuous actions for next-step execution.

\begin{table}[t]
\centering
\setlength{\tabcolsep}{0.5pt}
\resizebox{0.48\textwidth}{!}{
\begin{tabular}{cl ccccc cc}
\toprule
\multirow{2}[1]{*}{\shortstack{Room \\Type}}&\multirow{2}[2]{*}{Method} 
& \multicolumn{5}{c}{Visual} & \multicolumn{2}{c}{Physics} \\ 
\cmidrule(r){3-7} \cmidrule(r){8-9}
&& { \#Obj $\uparrow$ }
& { Real. $\uparrow$} 
& { Func. $\uparrow$ }
& { Lay.$\uparrow$ }
& { Comp. $\uparrow$ }
& { Coll. \% $\downarrow$ }
& { Stab. \% $\uparrow$ } \\ 
\midrule
\multirow{3}{*}{Bedroom\enspace}&Holodeck\cite{yang2024holodeck} & 
 28.5  &  7.4 & 6.8  & 5.0  & 6.1 & 29.1  &  51.0 \\
&SceneWeaver\cite{yang2025sceneweaver} & 
 17.5  &  9.0 &  9.7 &  7.8 &  7.5  & 31.0  & 58.8  \\
&\model (Ours) &
 \textbf{48.3}  & \textbf{9.0}  & \textbf{10.0}  & \textbf{8.0}  &  \textbf{9.5} & \textbf{2.3}  &  \textbf{99.8} \\
\midrule
\multirow{3}{*}{Kitchen}&Holodeck\cite{yang2024holodeck} & 
 28.5  & 6.7  & 6.1  &  4.4 &  6.2 & 16.0  &  73.8 \\
&SceneWeaver\cite{yang2025sceneweaver} & 
 37.5  & 8.2 & 7.7 & 6.8  & 7.2  &  28.0 & 67.0  \\
&\model (Ours) &
 \textbf{47.6}  & \textbf{8.5}  &  \textbf{9.0} & \textbf{7.8}  & \textbf{7.6}  & \textbf{0.7}  & \textbf{100.0}  \\
\midrule
\multirow{3}{*}{\shortstack{Living \\Room}} & Holodeck\cite{yang2024holodeck} & 
 34.0  &  8.3 &  7.3 &  5.7 &  7.3 &  20.8 & 66.5  \\
& SceneWeaver\cite{yang2025sceneweaver} & 
  18.1 &  8.5  &  9.3  &  7.2 & 6.8  & 39.5  & 77.2  \\
& \model (Ours) &
 \textbf{48.8}  & \textbf{8.8}  & \textbf{9.5}  & \textbf{7.8}  & \textbf{7.4}  &  \textbf{2.7} &  \textbf{100.0} \\
\midrule
\multirow{3}{*}{Average} & Holodeck\cite{yang2024holodeck} & 
 30.3  & 7.5  & 6.7  & 5.0  & 6.5  & 22.0  & 63.8  \\
&SceneWeaver\cite{yang2025sceneweaver} & 
 24.4  & 8.6  & 8.9  & 7.3  & 7.2  & 32.8  & 67.7  \\
& \model (Ours) &
 \textbf{48.2}  & \textbf{8.8}  & \textbf{9.5}  & \textbf{7.9}  & \textbf{8.2}  & \textbf{1.9}  & \textbf{99.9}  \\
\bottomrule
\end{tabular}
}
\caption{\textbf{Scene generation evaluation on common scene types.} Scores averaged across 10 scenes per room type. \model consistently outperforms prior methods across all categories.}
\vspace{-10pt}
\label{tab:exp_common}
\end{table}

\section{Experiments}

We evaluate scene generation in Sec.~\ref{sec:exp_scene_generation}. In Sec.~\ref{sec:result_learning}, we demonstrate how scalable scene and action data generated by our framework improves policy generalization.

\begin{table*}[t]
\centering
\resizebox{\textwidth}{!}{
\setlength{\tabcolsep}{0pt}
\begin{tabular}{@{}cccc@{}}

\includegraphics[width=0.25\linewidth,trim=140 10 140 0, clip]{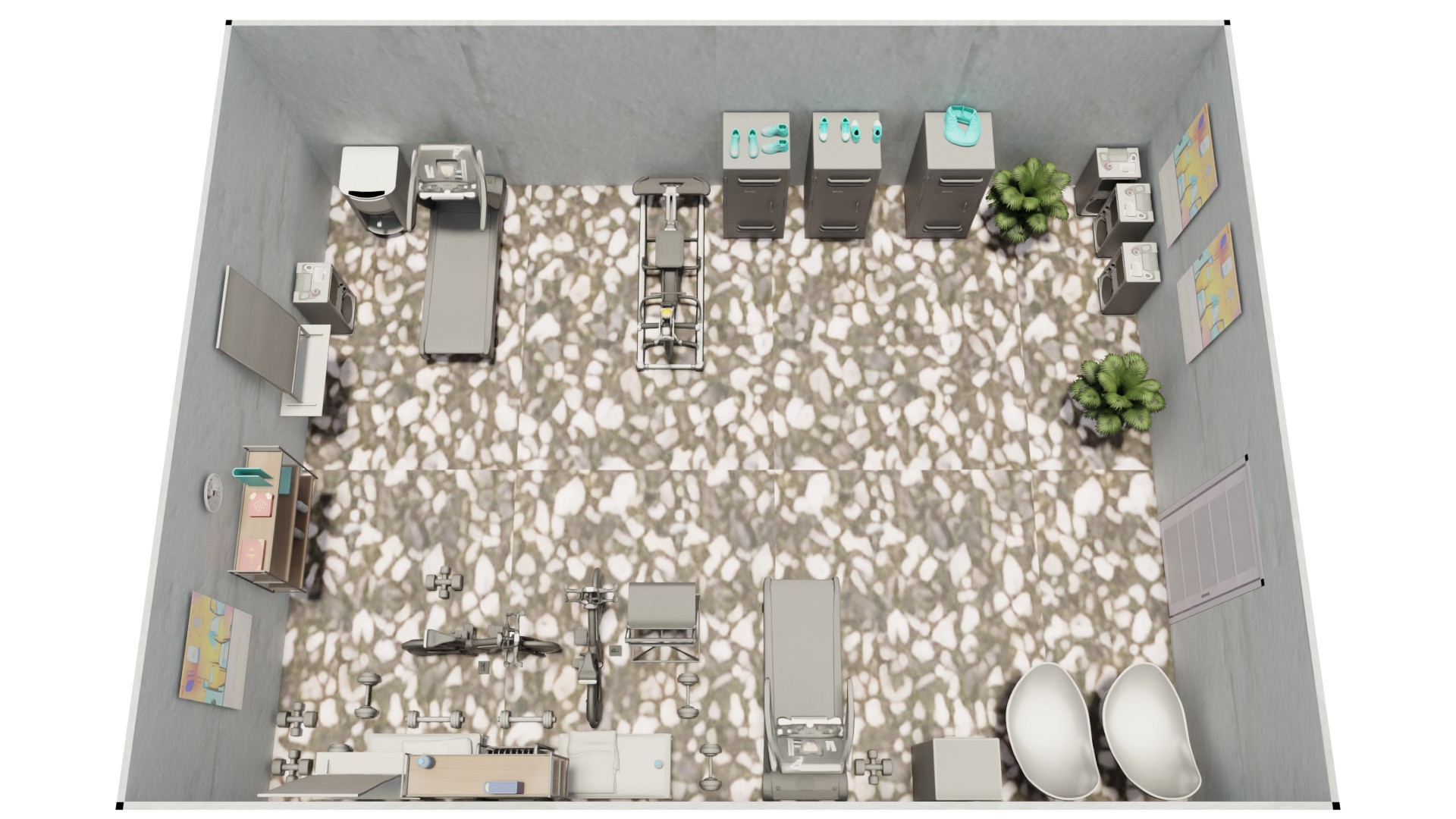} &
\centercrop[1.1]{
\includegraphics[width=0.25\linewidth,trim=190 40 190 0, clip]{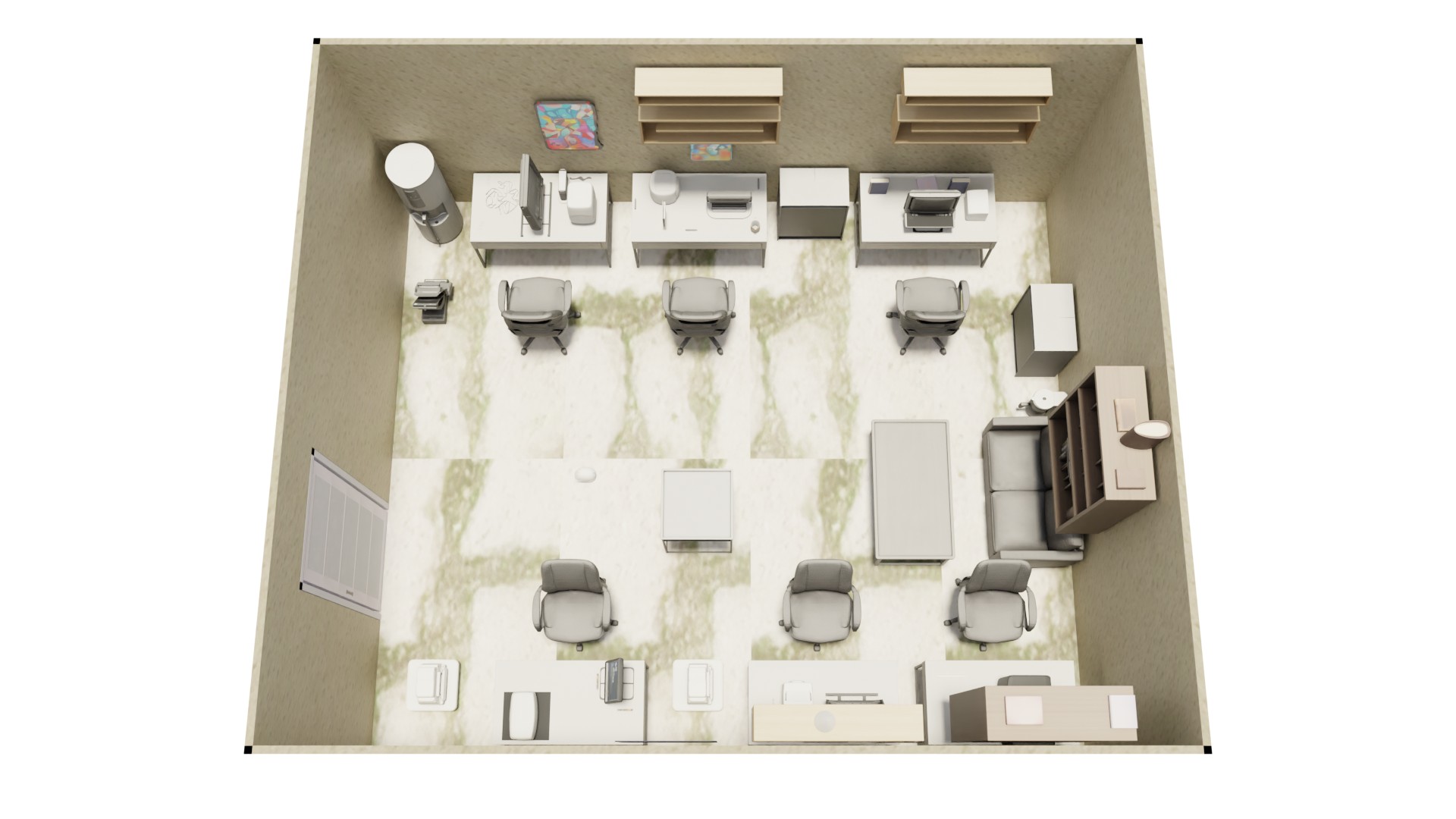}
} &
\includegraphics[width=0.25\linewidth,trim=190 40 190 0, clip]{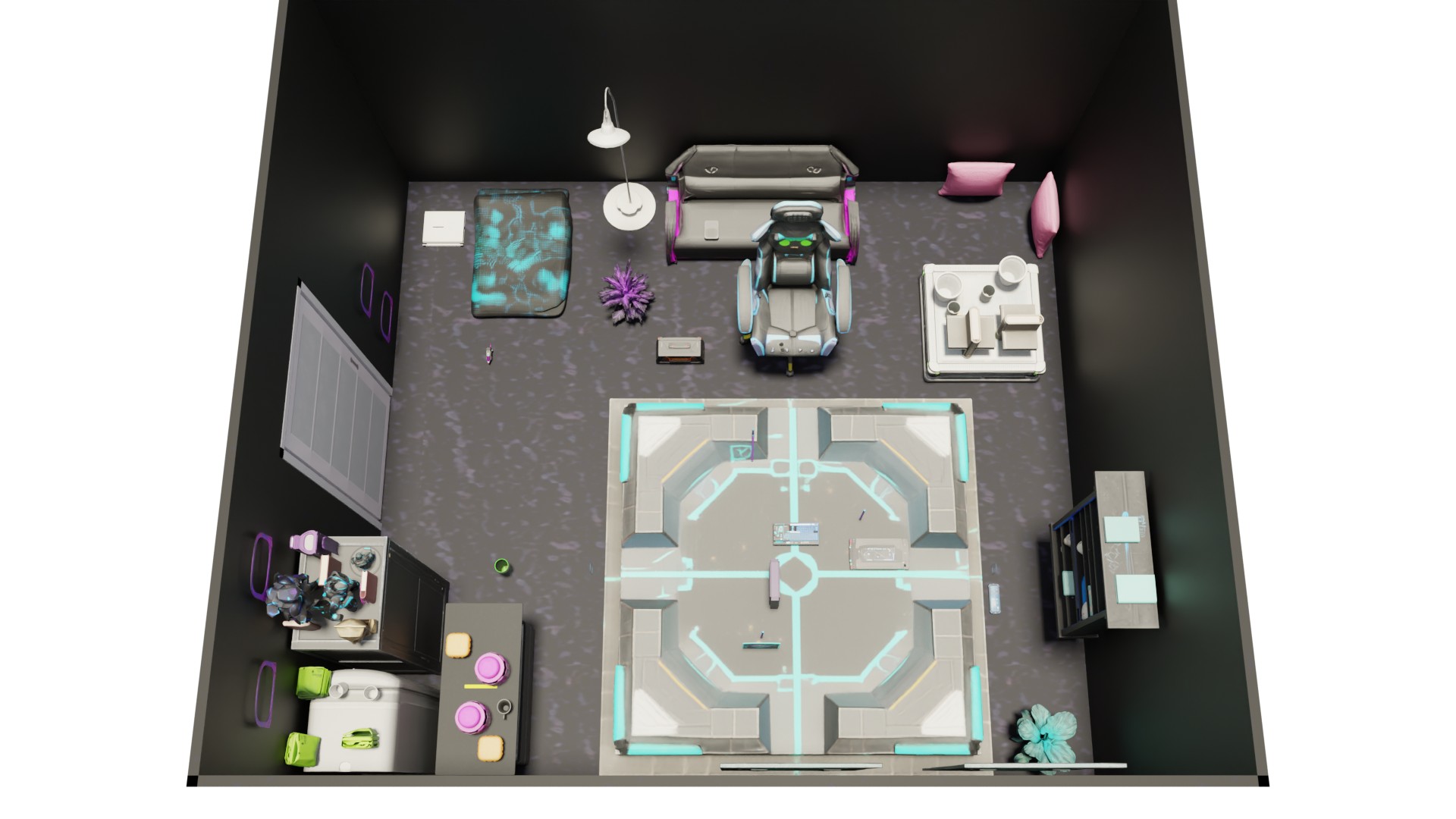} &
\includegraphics[width=0.25\linewidth,trim=180 40 180 20, clip]{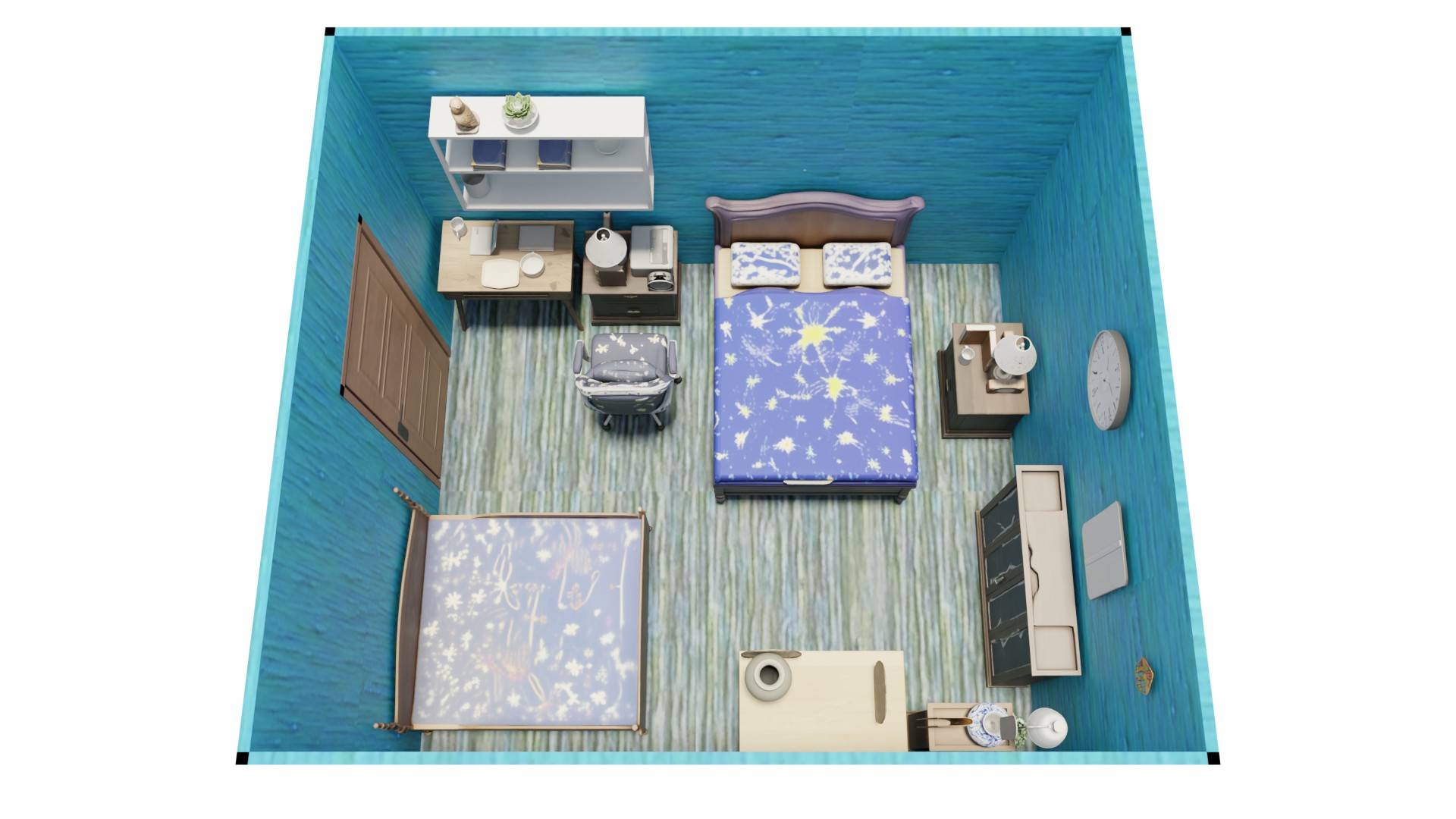} \\
``Gym'' & ``Office'' & ``Cyberpunk game den'' & ``Starry-night bedroom'' \\

\includegraphics[width=0.25\linewidth,trim=140 10 140 0, clip]{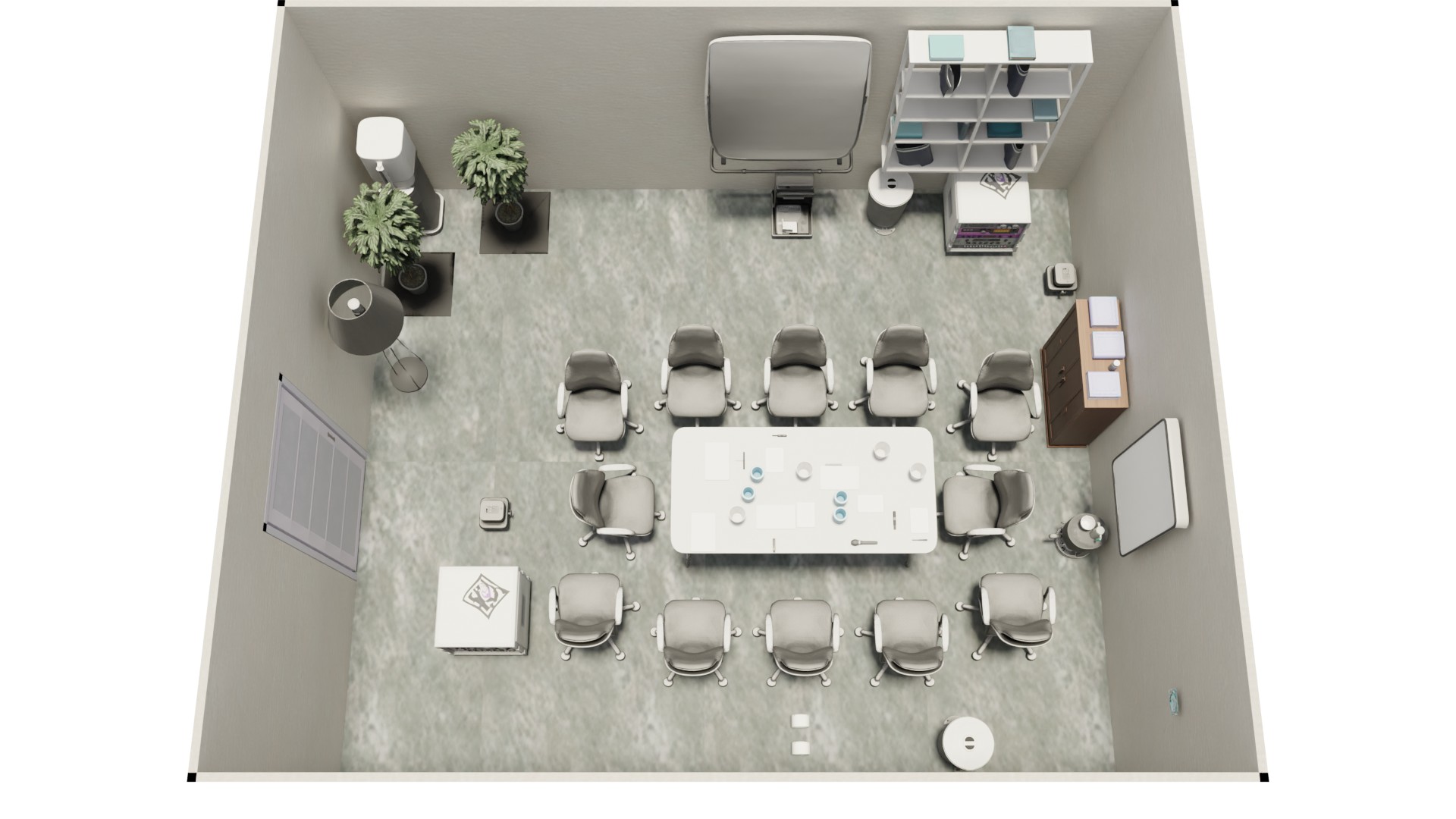} &
\centercrop[1.1]{
\includegraphics[width=0.25\linewidth,trim=190 40 190 0, clip]{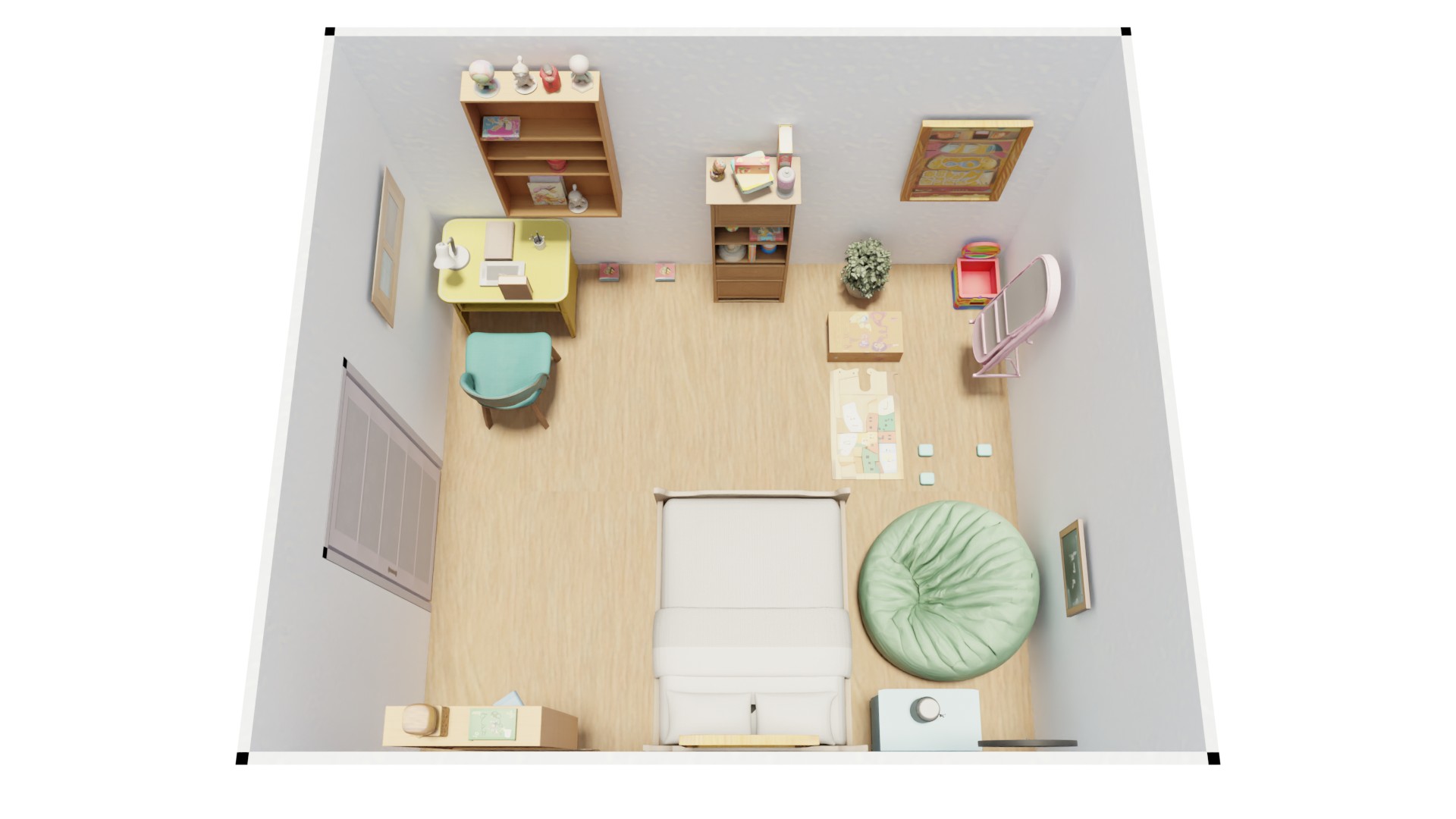}
} &
\includegraphics[width=0.25\linewidth,trim=190 40 190 0, clip]{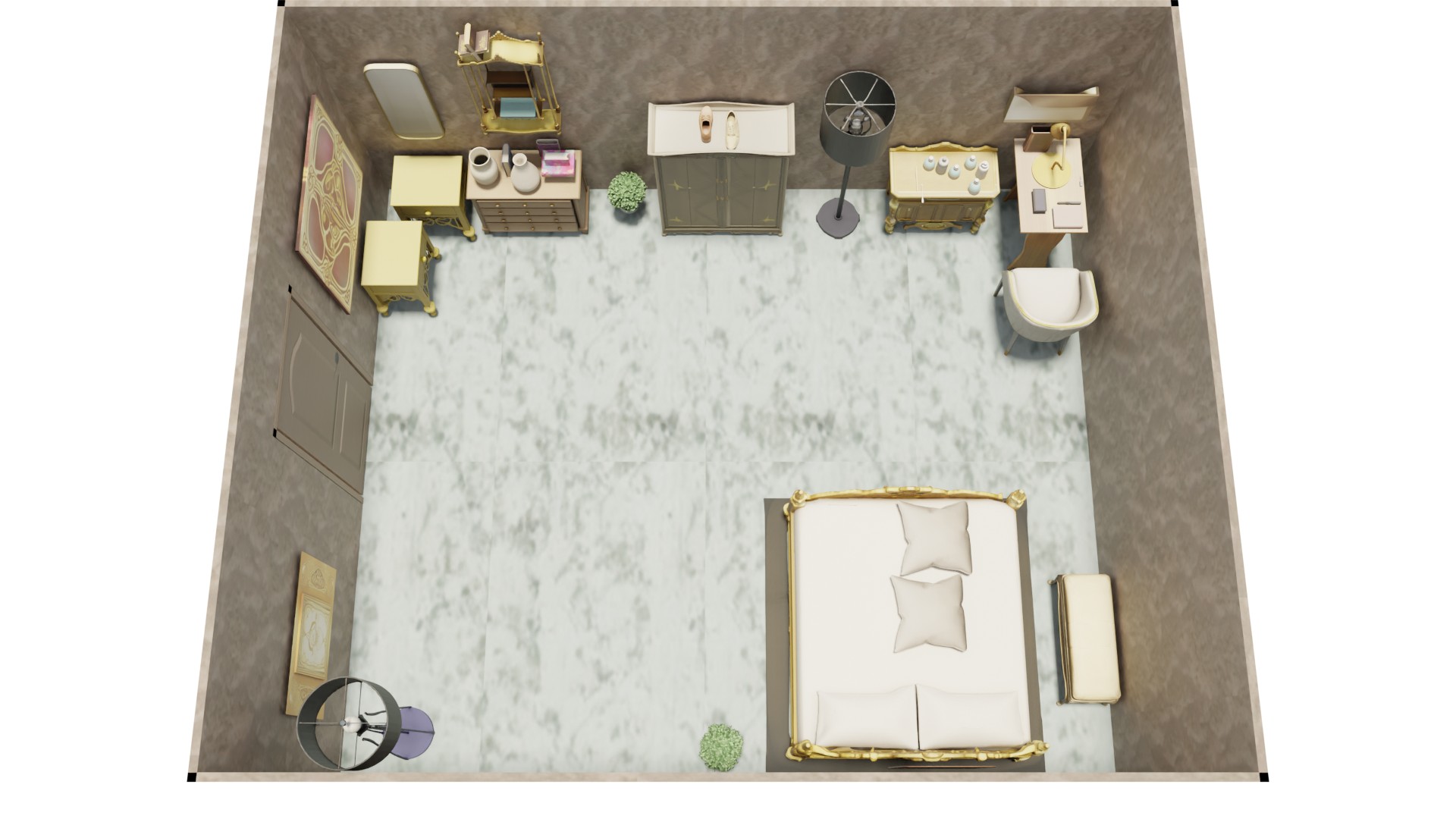} &
\includegraphics[width=0.25\linewidth,trim=180 40 180 20, clip]{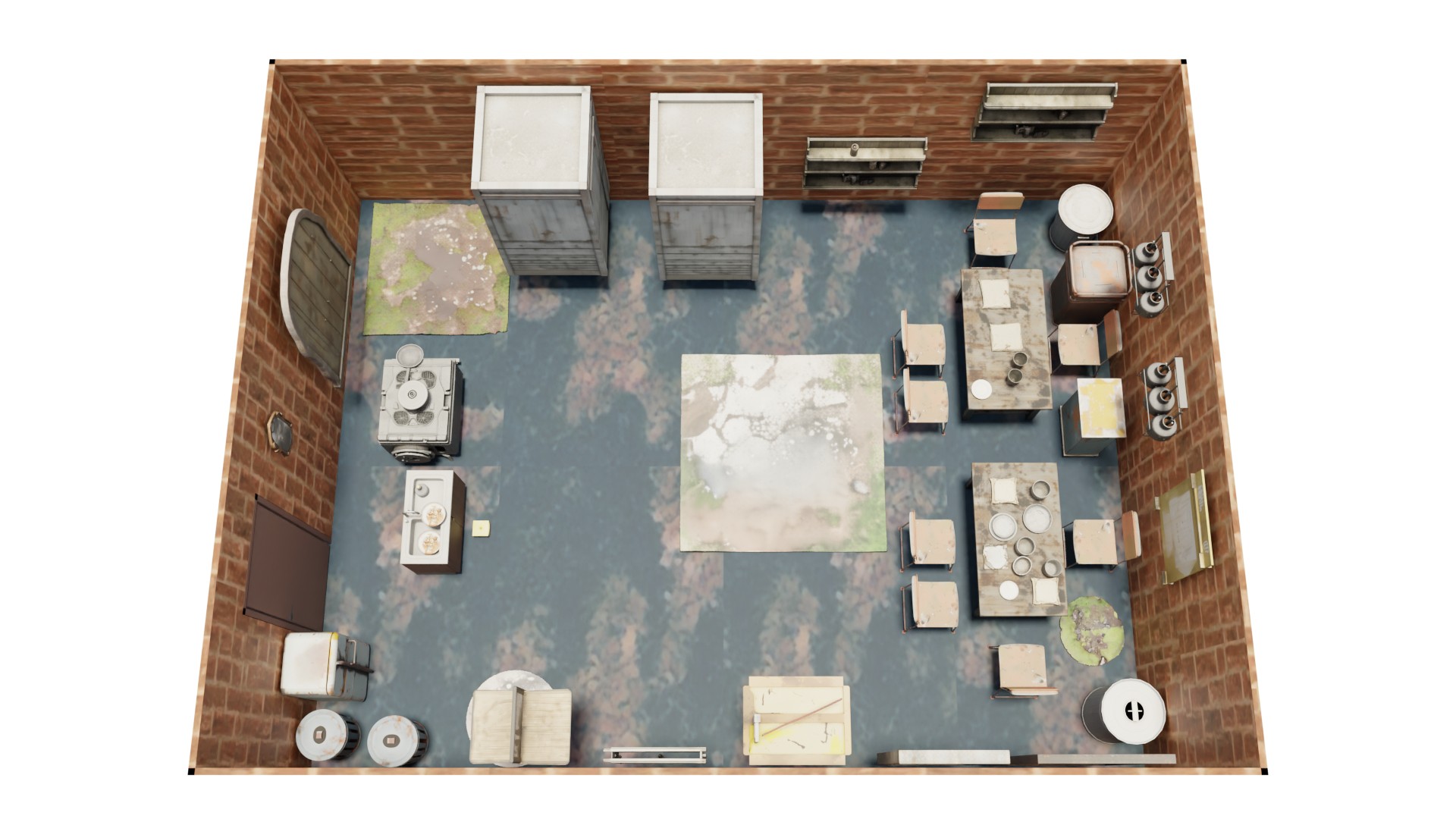} \\
``Meeting room'' & ``Children room''  & ``Golden and luxury bedroom'' & ``Muddy and dirty dining room'' \\

\end{tabular}
}
\captionof{figure}{\textbf{Additional open-vocabulary generation.} \model produces diverse, semantically coherent scenes spanning various styles and functionalities, from Gym and Office spaces to creative themes like ``Cyberpunk game den'' and ``Starry-night bedroom''.}
\label{fig:open_vocab}
\end{table*}

\begin{table}[h]
\centering
\resizebox{0.48\textwidth}{!}{
\setlength{\tabcolsep}{0pt}
\begin{tabular}{@{}c @{\hspace{3pt}} ccc@{}}
& Holodeck \cite{yang2024holodeck} & SceneWeaver \cite{yang2025sceneweaver} & \model (Ours) \\
{\rotatebox{90}{\qquad  Before}} &
\includegraphics[width=0.2\textwidth,trim=0 0 0 0, clip]{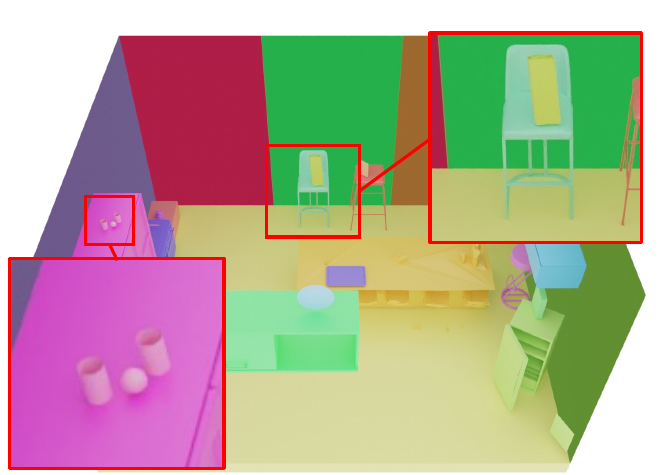} &
\includegraphics[width=0.2\textwidth,trim=0 0 0 0, clip]{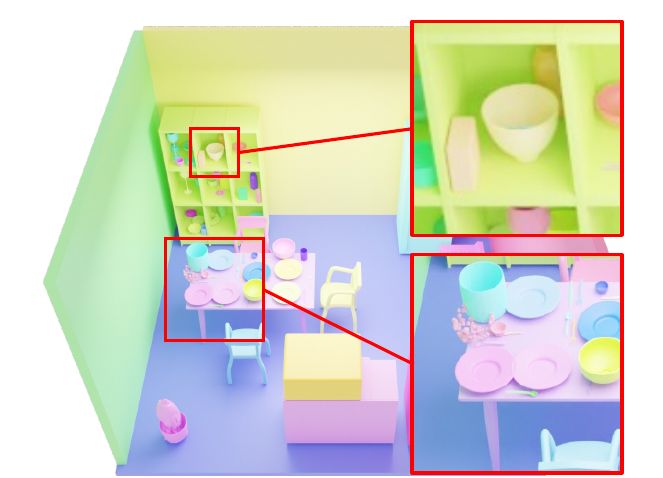} &
\includegraphics[width=0.17\textwidth,trim=300 0 300 0, clip]{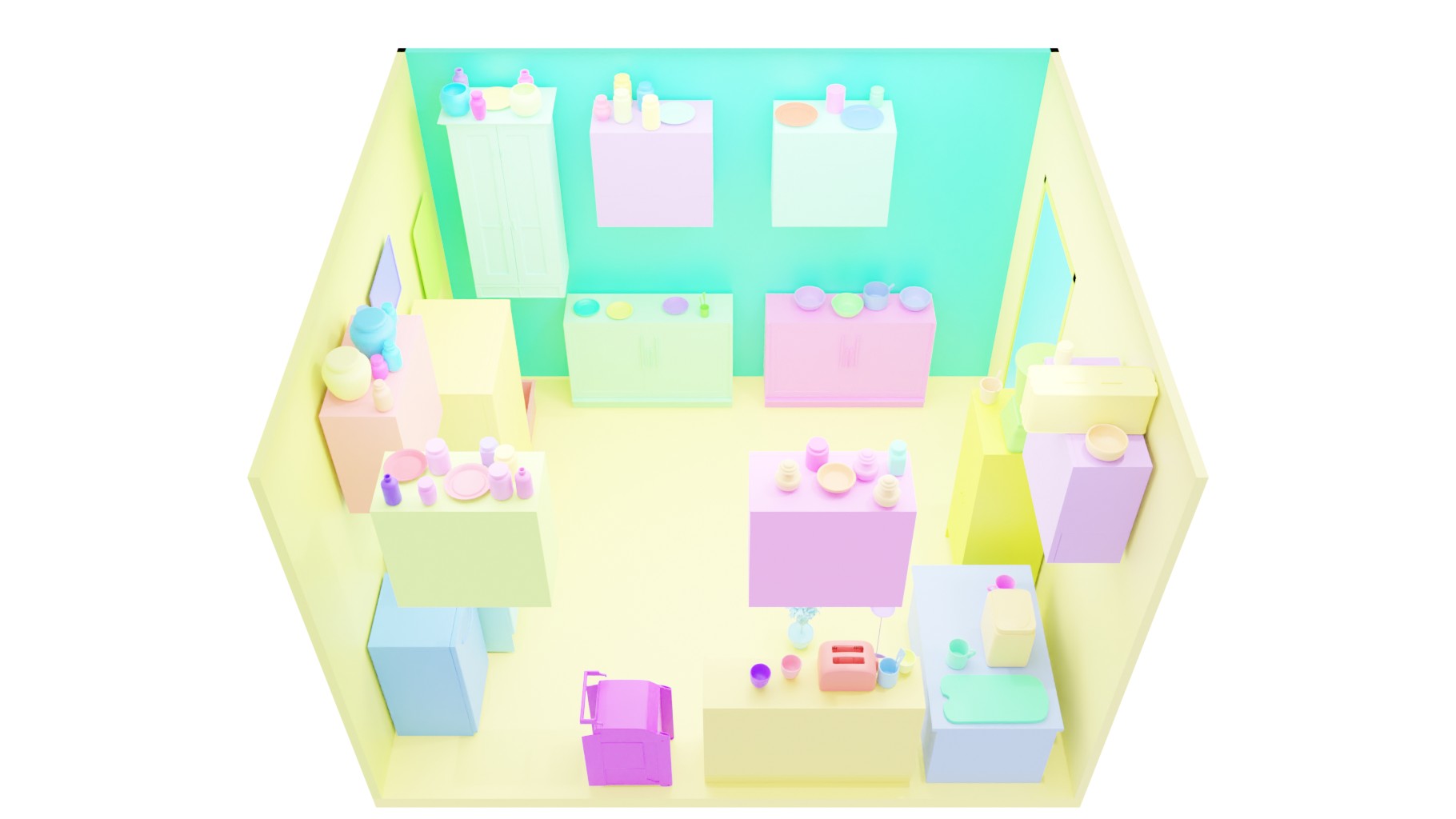} \\
\rotatebox{90}{\qquad  After} &
\includegraphics[width=0.2\textwidth,trim=0 0 0 0, clip]{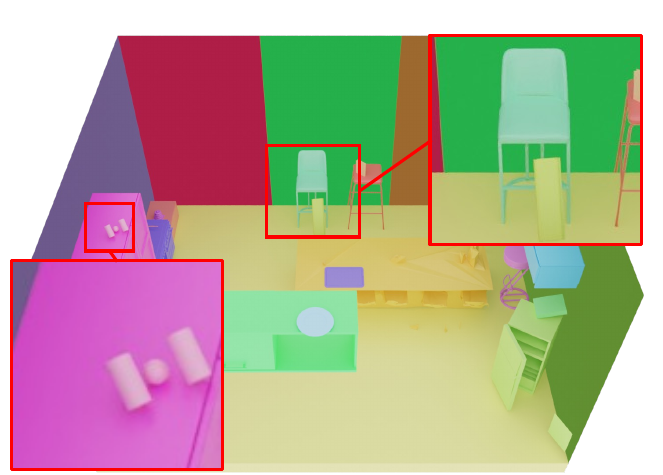} &
\includegraphics[width=0.2\textwidth,trim=0 0 0 0, clip]{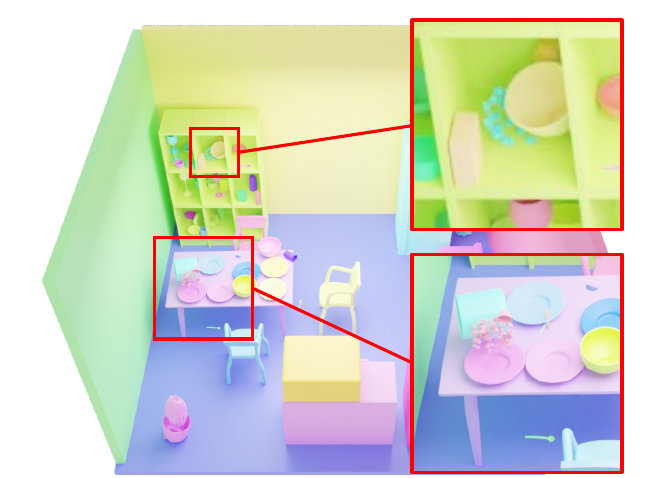} &
\includegraphics[width=0.17\textwidth,trim=300 0 300 0, clip]{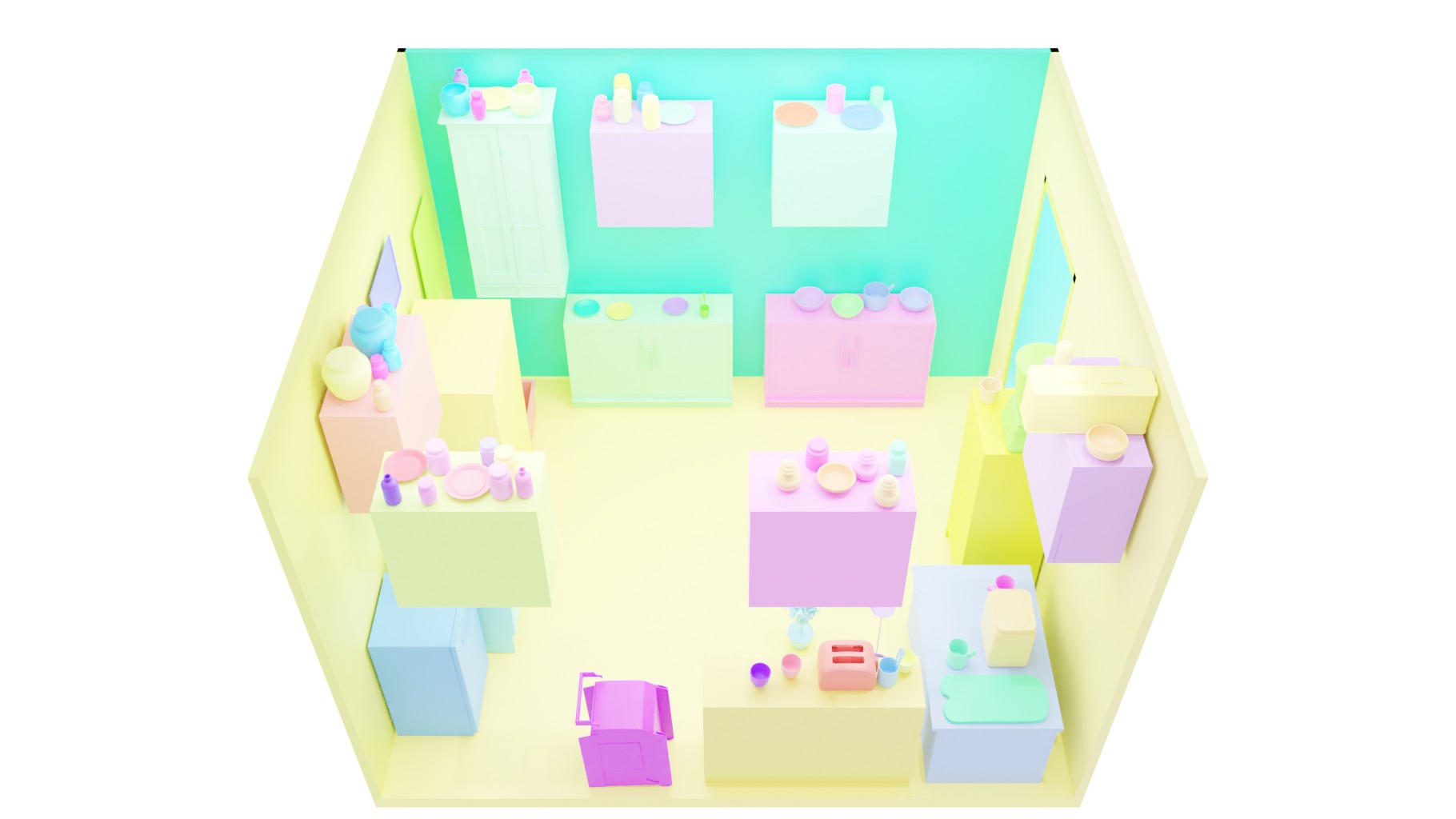} \\
\end{tabular}
}
\captionof{figure}{\textbf{Stability verification.} Generated scenes are loaded into IsaacSim for physical validation. Both baselines exhibit displaced objects due to instability, whereas  \model preserves scene stability before and after simulation.}
\label{fig:stability}
\end{table}
\begin{table*}[h]
\centering
\resizebox{0.96\textwidth}{!}{
\setlength{\tabcolsep}{3pt}

\begin{tabular}{c}
\includegraphics[width=\textwidth]{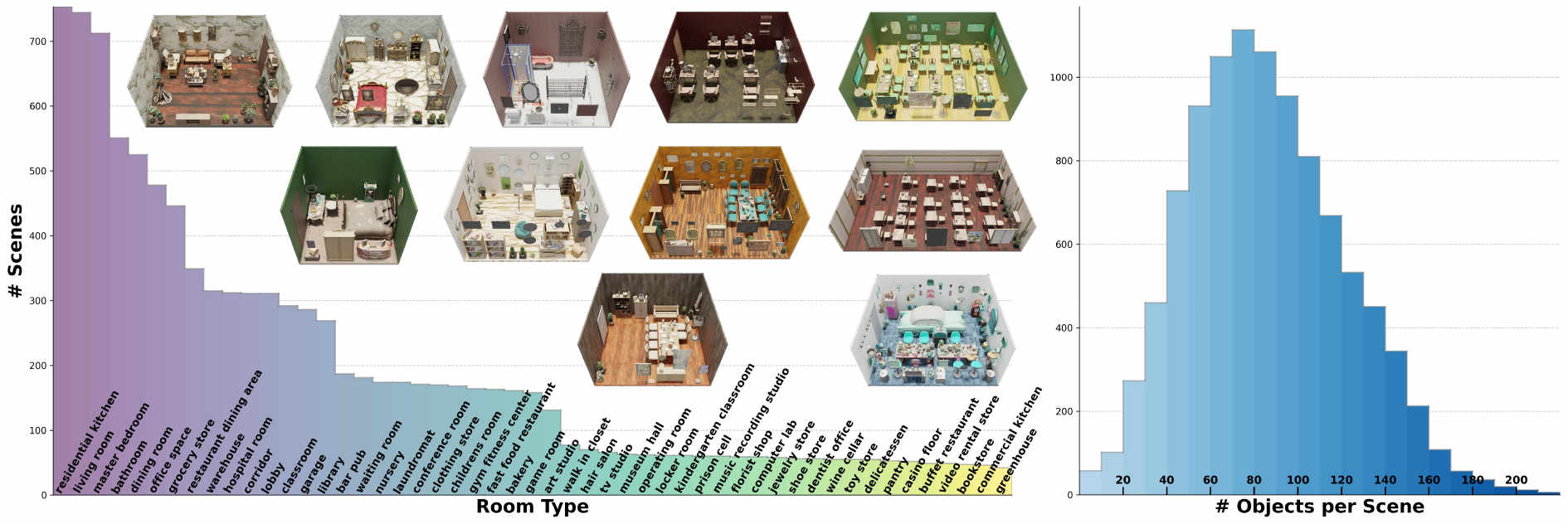} \\

\end{tabular}
}
\captionof{figure}{
\textbf{SAGE-10k Dataset}: We pre-generated a 10k-scene dataset named SAGE-10k Dataset across 50 room types and 50 styles, including 565K uniquely generated 3D objects.  
We include the statistics of room types, room examples, and objects per scene in the figure as well.
The dataset can be accessed via this 
\href{https://huggingface.co/datasets/nvidia/SAGE-10k}{link}.
}
\label{fig:sage10k}
\end{table*}

\begin{table*}[t]
\centering
\resizebox{\textwidth}{!}{
\setlength{\tabcolsep}{0pt}
\begin{tabular}{@{}ccc@{}}

\includegraphics[width=0.33\textwidth]{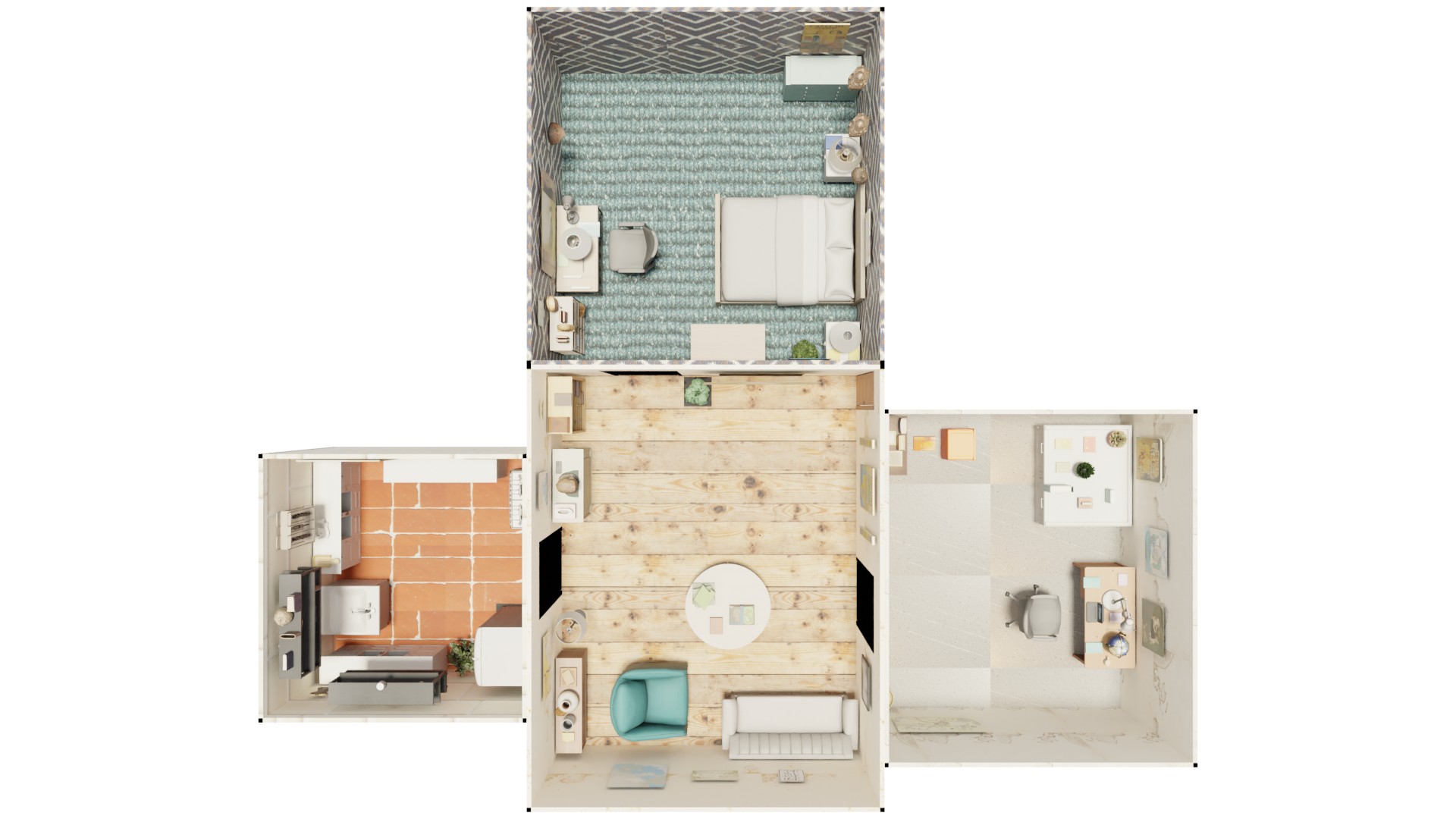} &
\includegraphics[width=0.33\textwidth]{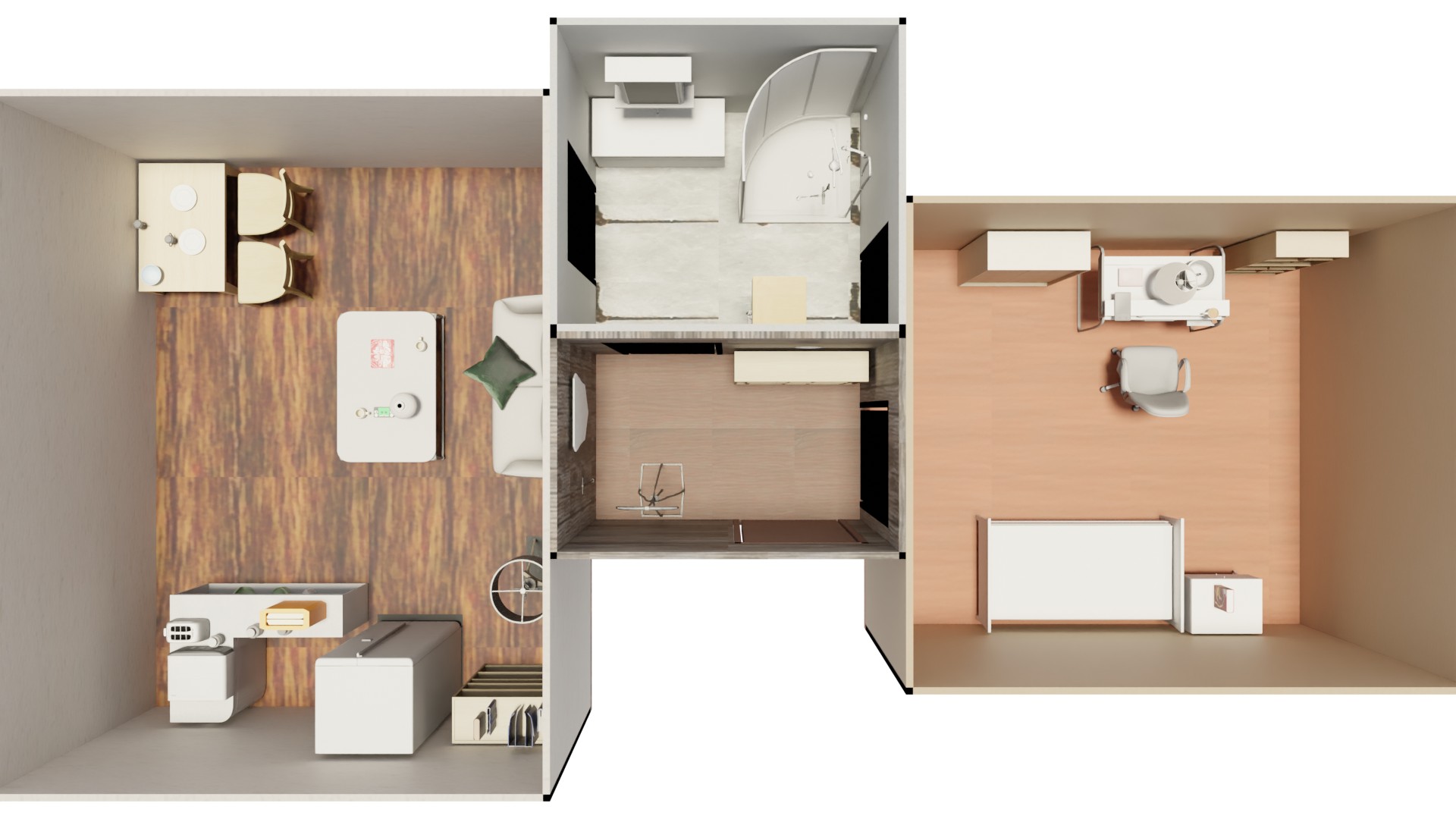} &
\includegraphics[width=0.33\textwidth]{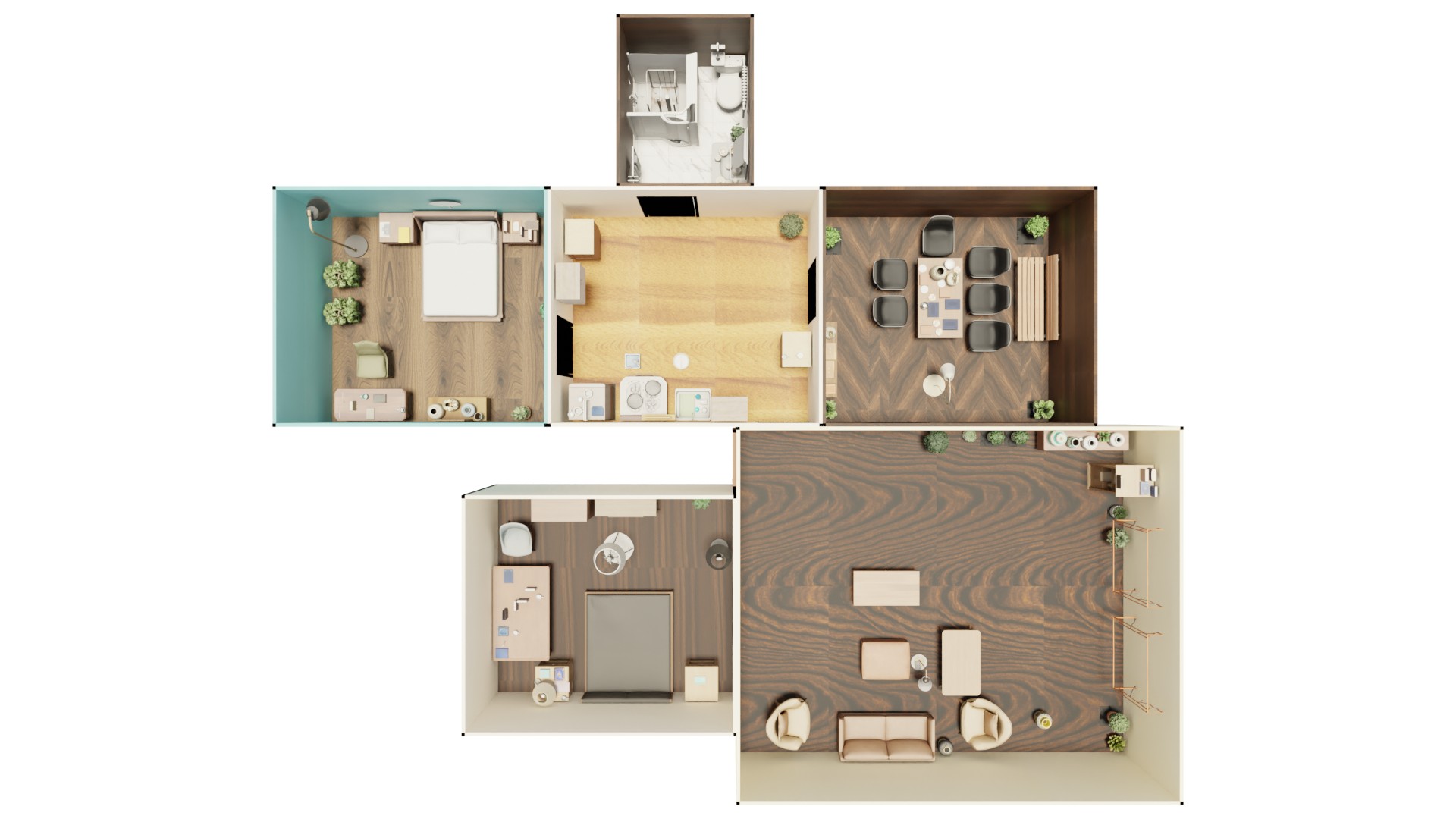}\\

``Multilingual teacher's apartment'' & 
``A student apartment with one bedroom'' & 
``Mid-century modern family home'' \\

\end{tabular}
}
\captionof{figure}{\textbf{Multi-room open-vocabulary generation.} \model can be extended to generate multi-room scenes at scale easily by generating the floor plan and then calling generator MCP tools to fill in multiple rooms in parallel.}
\label{fig:open_vocab_multi_room}
\end{table*}

\begin{table}[t]
\centering
\resizebox{0.5\textwidth}{!}{
\setlength{\tabcolsep}{2pt}
\begin{tabular}{@{}c@{\hspace{3pt}} cc@{}}
{\rotatebox{90}{\quad Ref Image}}&
{\includegraphics[width=0.22\textwidth,trim=0 0 0 0, clip]{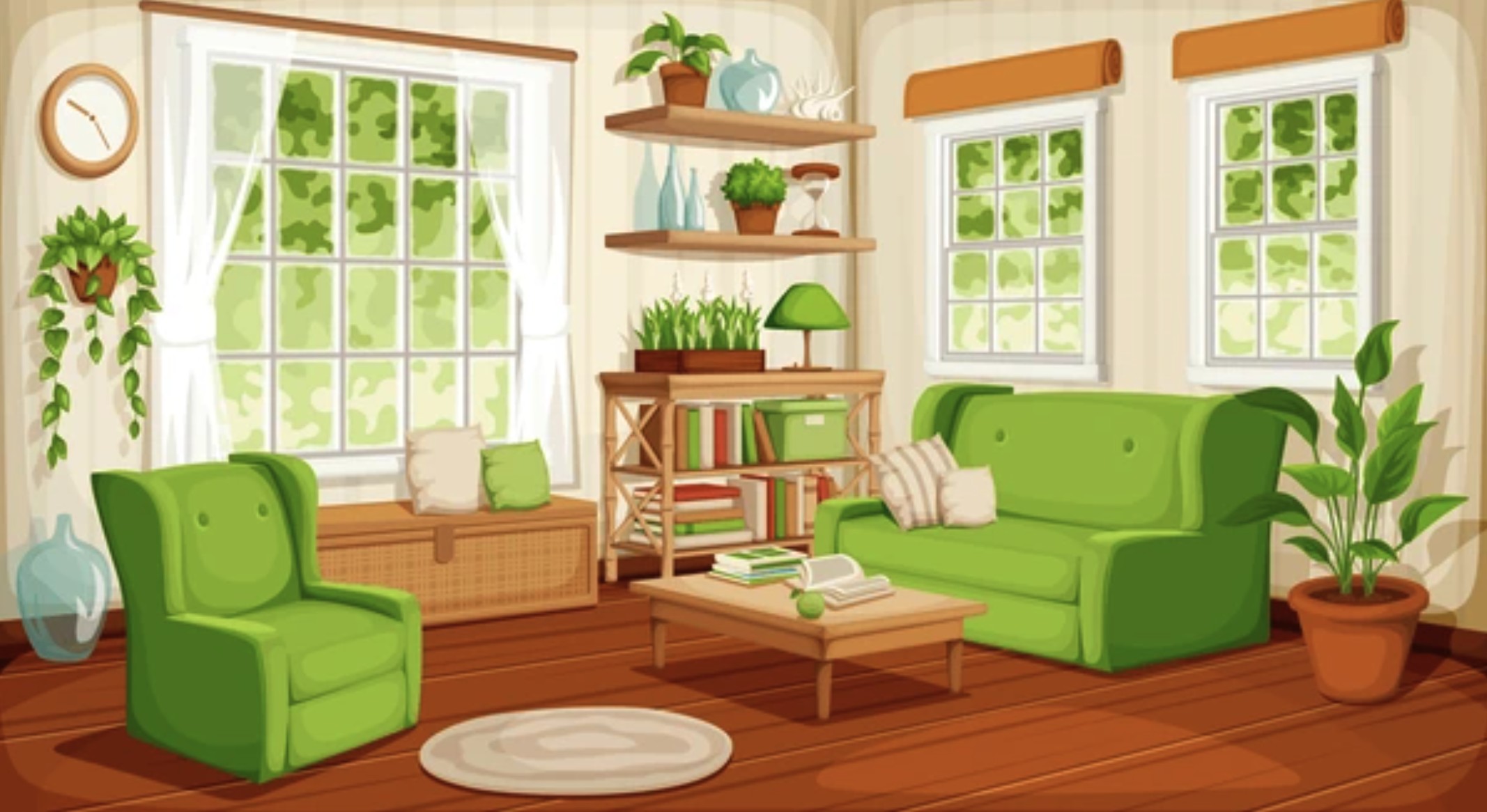}} &
{\includegraphics[width=0.22\textwidth,trim=0 0 0 235, clip]{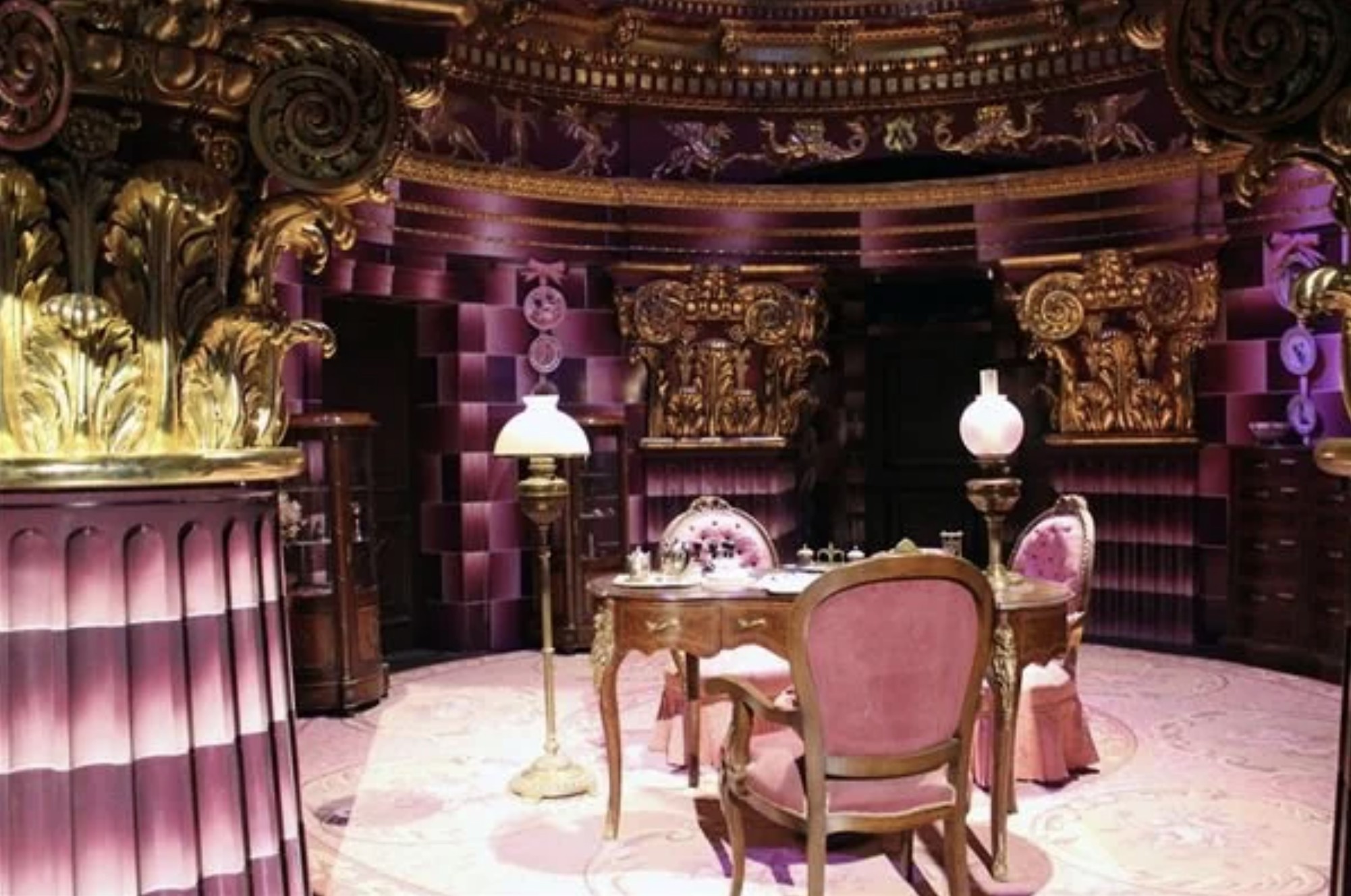}} \\
{\rotatebox{90}{\quad Generated Scene}}&
{\includegraphics[width=0.22\textwidth,trim=280 0 280 0, clip]{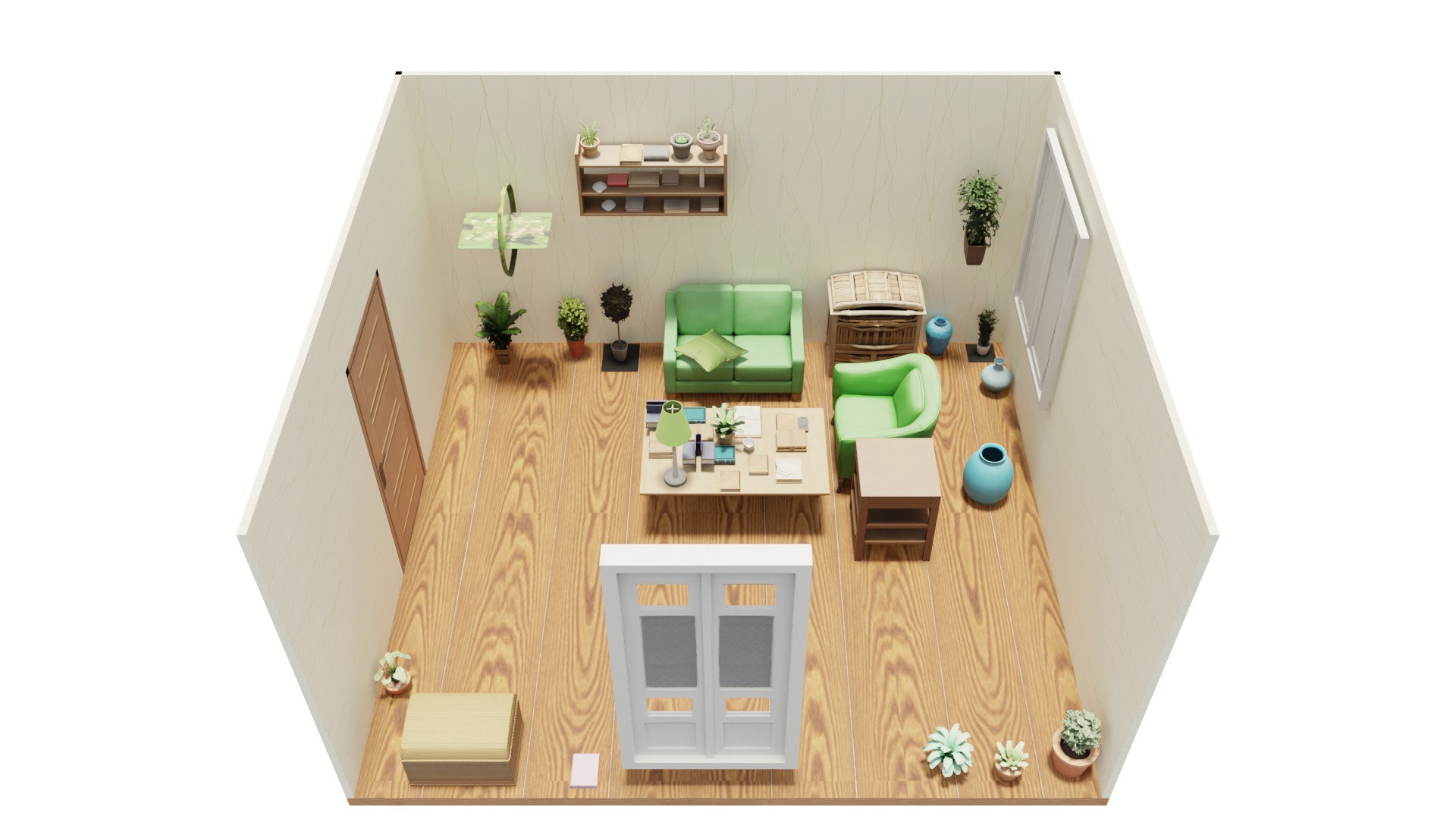}} &

{\includegraphics[width=0.22\textwidth,trim=280 0 280 0, clip]{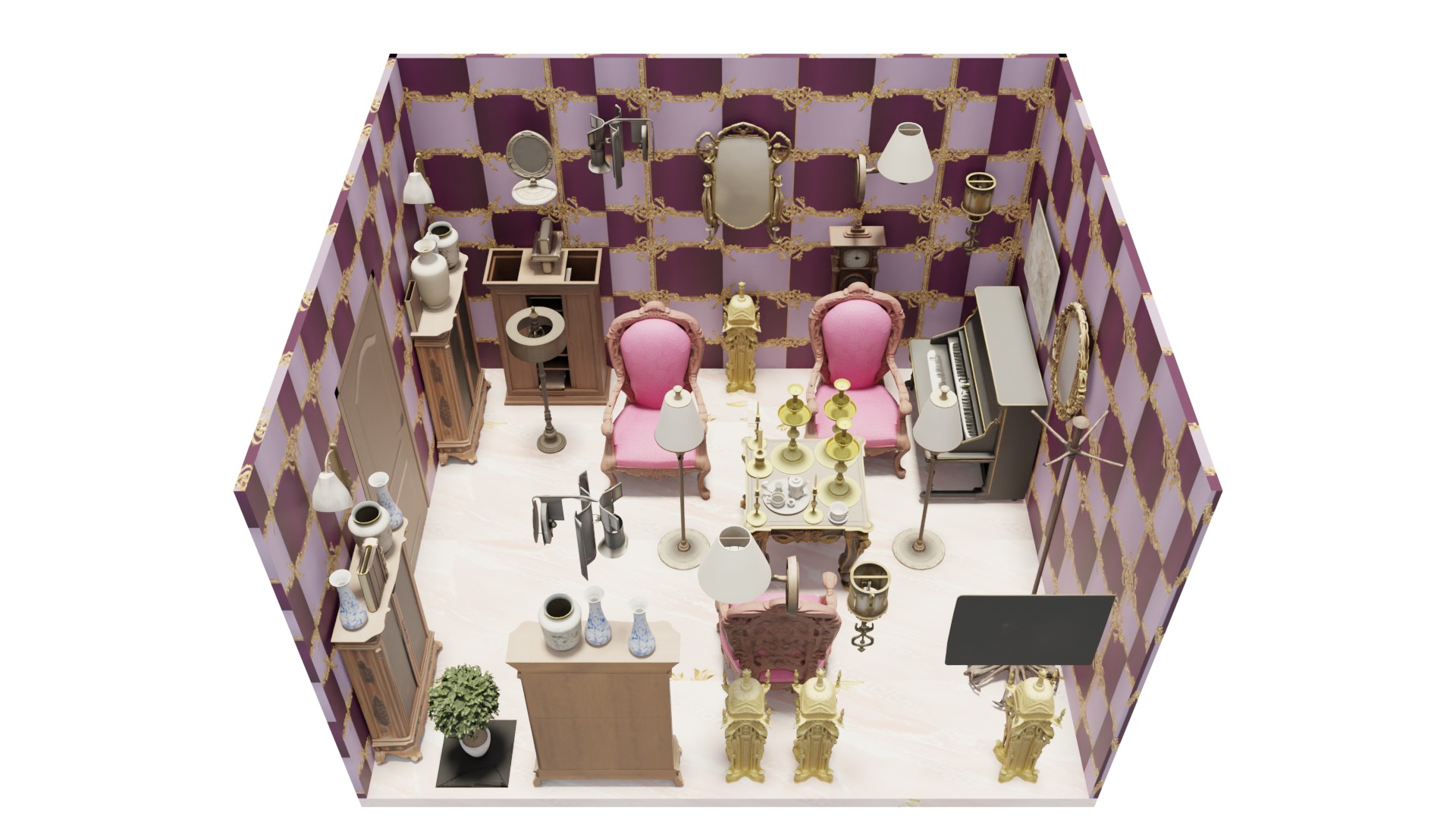}} \\

\end{tabular}
}
\captionof{figure}{\textbf{Image-conditioned scene generation.} 
Using Qwen3-VL \cite{qwen3}, \model extracts style and object attributes from reference images to enable image-conditioned scene generation without architectural modifications. The generated scenes are not pixel-aligned but remain semantically consistent with the reference images.
}
\label{fig:open_vocab_image_ref}
\end{table}

\begin{table*}[h]
\centering
\resizebox{\textwidth}{!}{
\setlength{\tabcolsep}{6pt}
\begin{tabular}{c}
\includegraphics[width=0.48\textwidth ]{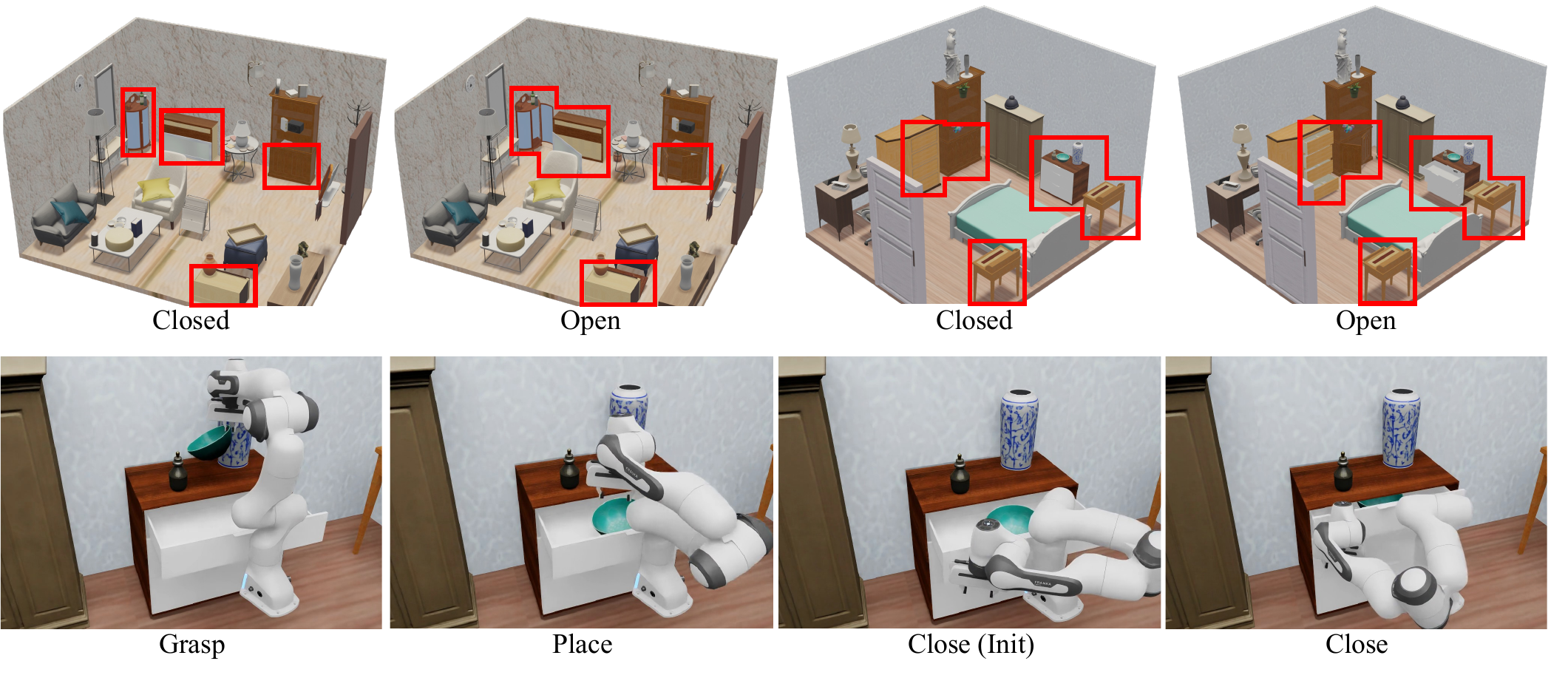} \\
\end{tabular}
}
\captionof{figure}{\textbf{Articulated Objects}: \model can be extended with articulated objects using retrieval from PartNet-Mobility \cite{xiang2020sapien}. \textit{Top}: we show two scenes with multiple articulated objects at closed and open states. \textit{Bottom}: an action sequence generated with grasp pose prediction and motion planning(~Sec.\ref{sec:action}) for ``pick up the bowl, place it in the drawer, and close the drawer''. Please visit website for full video. } 
\label{fig:art}
\end{table*}
\begin{table}[t]
\centering
\setlength{\tabcolsep}{2pt}
\resizebox{0.48\textwidth}{!}{
\begin{tabular}{cc | ccccc cc}
\toprule
\multicolumn{2}{c|}{Critics Setting} & \multicolumn{5}{c}{Visual} & \multicolumn{2}{c}{Physics} \\
\cmidrule(r){1-2} \cmidrule(lr){3-7} \cmidrule(r){8-9}
Visual  & Physics 
& { \#Obj $\uparrow$ }
& { Real. $\uparrow$} 
& { Func. $\uparrow$ }
& { Lay.$\uparrow$ }
& { Comp. $\uparrow$ }
& { Coll. \% $\downarrow$ }
& { Stab. \% $\uparrow$ } \\ 
\midrule
\xmark & \xmark  &
 35.3  & 8.5  & 9.2  & 7.5  & 7.3  & 7.8  &  80.3 \\
\cmark & \xmark &
 \underline{50.1}  & \textbf{8.9}  & \underline{9.5}  & \textbf{8.1}  & \underline{8.1}  & 3.7  &  84.1 \\
\xmark & \cmark &
 36.8  & 8.8  &  9.3 & 7.7  & 7.8  & \underline{1.9}  &  \underline{99.6}  \\
\cmark & \cmark &
 \textbf{53.7}  & \textbf{8.9}  &  \textbf{9.6} & \underline{8.0}  & \textbf{8.2}  & \textbf{0.8}  &  \textbf{100.0} \\
\bottomrule
\end{tabular}
}
\caption{\textbf{Ablation study on critics.} Results are averaged over five scenes for each of three room types. Adding both visual and physics critics leads to the best results, confirming their joint importance in scene generation. Best is in \textbf{bold}, second best is \underline{underlined}.}
\vspace{-5mm}
\label{tab:critic_ablation}
\end{table}

\begin{table*}[t]
\centering
\resizebox{0.8\textwidth}{!}{
\setlength{\tabcolsep}{3pt}
\begin{tabular}{cc}
\includegraphics[height=66px]{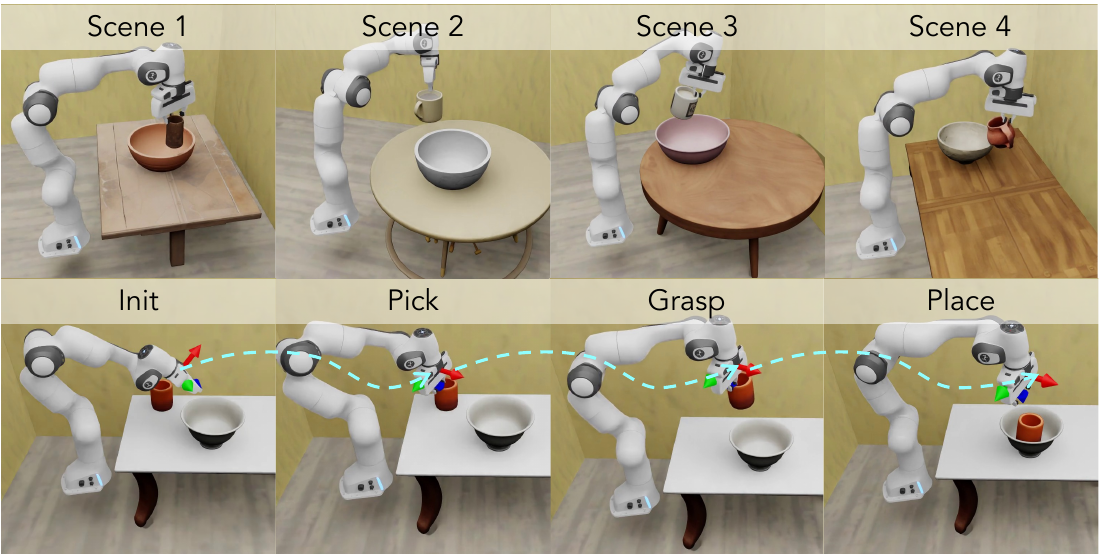} &
\includegraphics[height=66px]{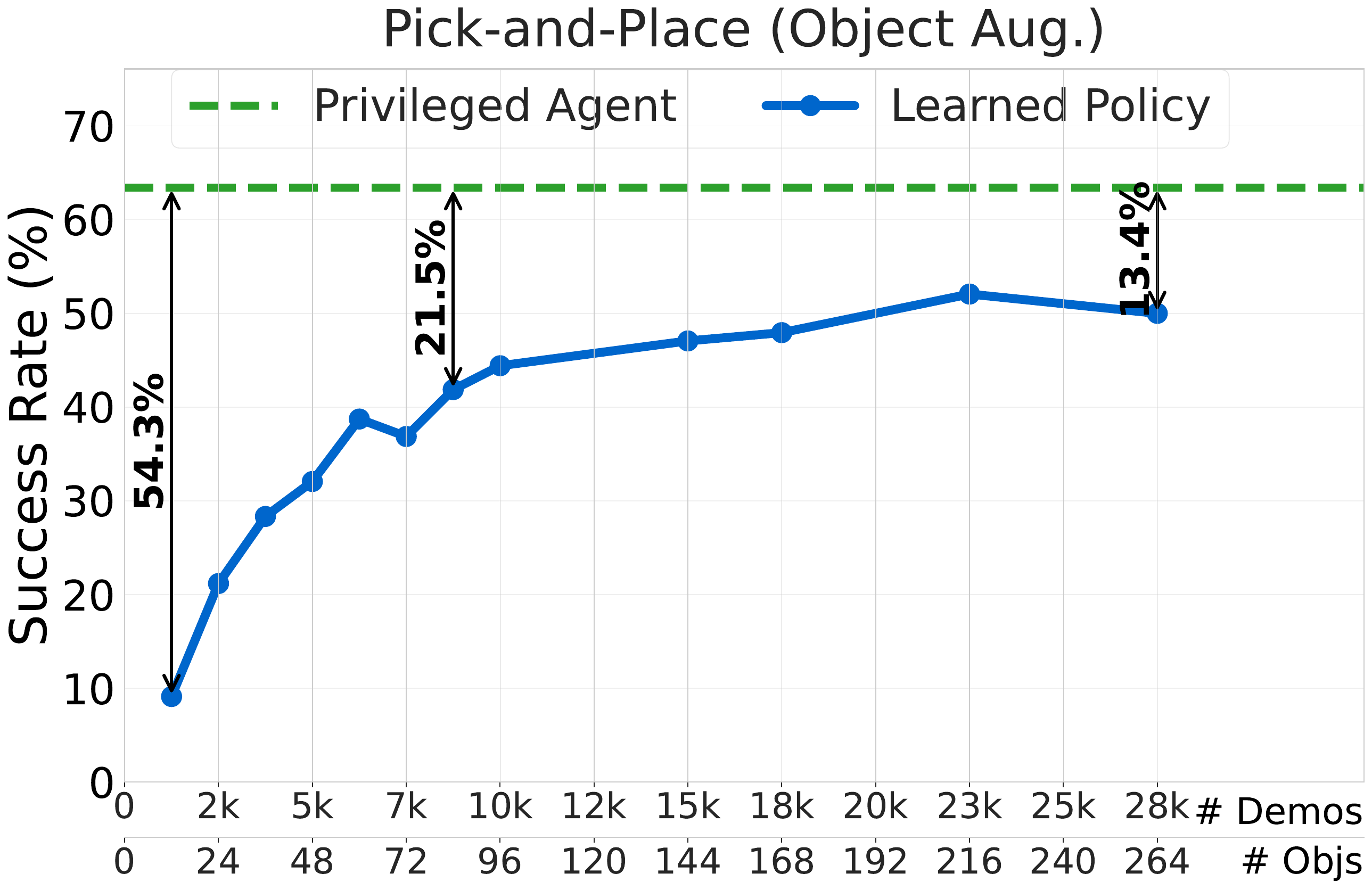} \\
\end{tabular}
}
\captionof{figure}{\textbf{Examples and scaling curve on Pick-and-Place.} \textit{Top left}: diverse generation. \textit{Bottom left}: example trajectory. \textit{Right}: success rate \textit{w.r.t.} demo/object counts. More diverse object augmentations improve policy success, narrowing the gap to the privileged agent. }
\label{fig:robot_graph_pnp}
\end{table*}

\begin{table*}[t]
\centering
\resizebox{\textwidth}{!}{
\setlength{\tabcolsep}{3pt}
\begin{tabular}{cc}
\includegraphics[height=66px]{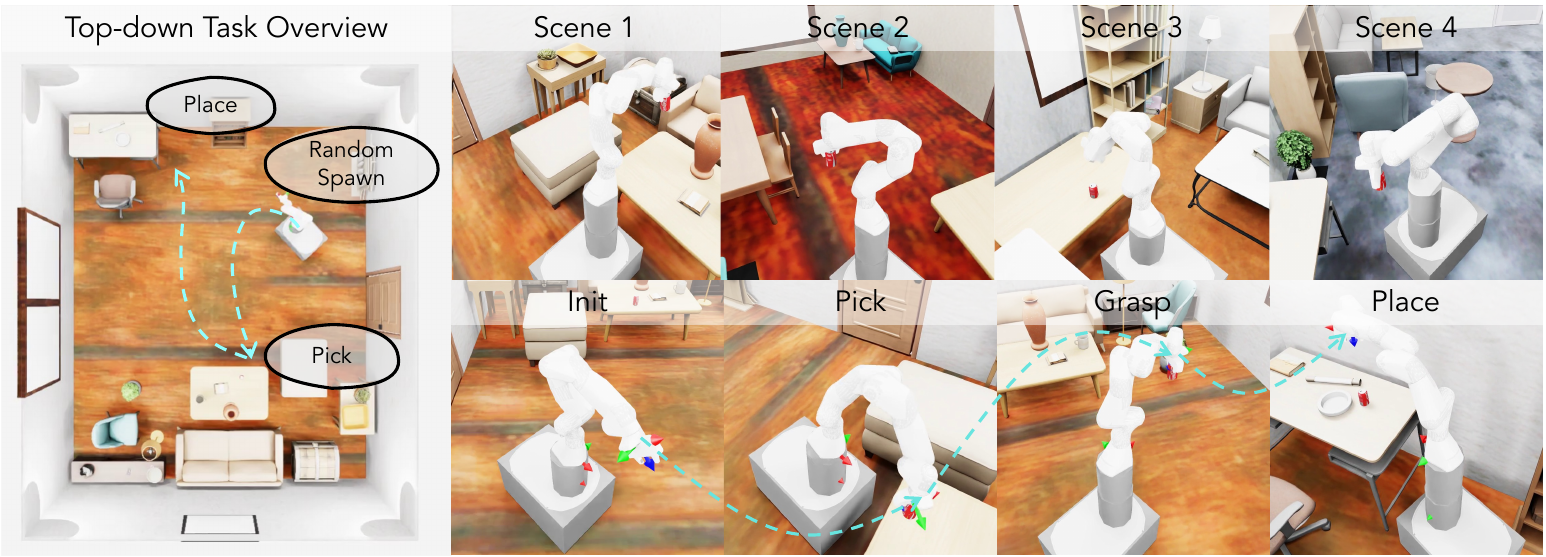} &
\includegraphics[height=66px]{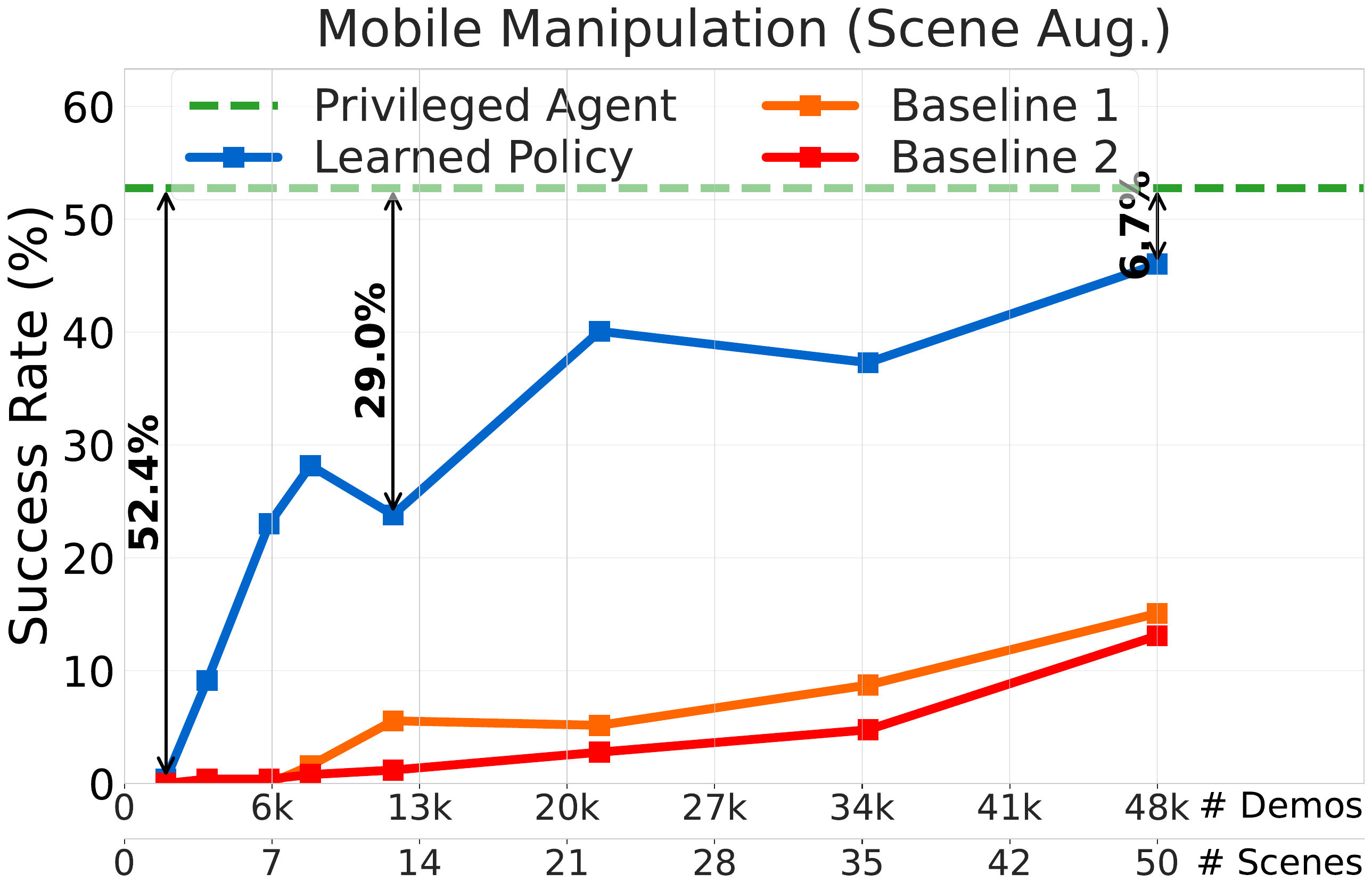} \\
\end{tabular}
}
\captionof{figure}{\textbf{Examples and scaling curve on Mobile Manipulation.} \textit{Left}: task overview. \textit{Mid-top}: diverse generation. \textit{Mid-bottom}: example trajectory. \textit{Right}: success rate \textit{w.r.t.} demo/scene counts. Both baselines omit physics critic and replace text-to-3D object synthesis with retrieval: \textbf{Baseline 1} mimics SceneWeaver~\cite{yang2025sceneweaver}; \textbf{Baseline 2} further replaces the agent with a fixed pipeline, resembling Holodeck~\cite{yang2024holodeck}. Diverse \model-augmented scenes boost the \textbf{learned policy}'s success and close the gap to the privileged agent, while removing physics critic and object synthesis degrades performance (\textbf{Baseline 1}). Replacing the agent with a static pipeline further reduces success rate (\textbf{Baseline 2}).}
\label{fig:robot_graph_mm}
\vspace{-4mm}
\end{table*}

\begin{table}[h]
\centering
\resizebox{0.4\textwidth}{!}{
\begin{tabular}{lcccc}
\toprule
\multirow{2}{*}{Task} & \multicolumn{2}{c}{Privileged Agent} & \multicolumn{2}{c}{Policy Rollout} \\
\cmidrule(r){2-3} \cmidrule(r){4-5}
 & Train & Test & Train & Test \\
\midrule
Pick-and-Place & 65.3 & 57.7 & 63.4 & 50.0 \\
Mobile Manipulation & 68.4 & 52.8  & 54.6 & 46.0 \\
\bottomrule
\end{tabular}
}
\caption{\textbf{Comparison of training and test success rates for policy rollout and motion planning.} The learned policy generalizes well from training to testing, and policy rollout performance approaches that of the privileged motion planning.}
\label{tab:train_mp_metric}
\end{table}

\begin{table}[h]
    \centering
    \setlength\tabcolsep{5pt}
    \resizebox{0.4\textwidth}{!}{
    \begin{tabular}{lccc}
        \toprule
        \multirow{2}{*}{\shortstack[c]{\\Training\\Source}} & \multicolumn{3}{c}{Test Source Succuss Rate (\%)} \\
        \cmidrule{2-4}
                   & Baseline 1~\cite{yang2025sceneweaver} & Baseline 2~\cite{yang2024holodeck} & SAGE \\
        \midrule
        Baseline 1~\cite{yang2025sceneweaver} & 13.2 & 9.3 & 14.4 \\
        Baseline 2~\cite{yang2024holodeck} & 13.5 & 16.2 & 13.1 \\
        \textbf{SAGE (Ours)} & \textbf{39.1} & \textbf{24.7} & \textbf{46.0} \\
        \bottomrule
    \end{tabular}
    }
    \captionof{table}{{\bf Cross-Evaluation.} \model policies generalize better on out-of-distribution scenes. \model even achieves higher success rates on baseline-generated scenes.}
    \label{tab:cross_eval}
\end{table}

\subsection{Scene Generation}
\label{sec:exp_scene_generation}

\subsubsection{Setup}

\paragraph{Implementation Details}
 \model integrates a series of large foundation models, including LLMs, VLMs, and specialized generators for 3D objects and background textures.
We use open-source models hosted via self-managed APIs to ensure reproducibility.
Specifically, we adopt gpt-oss-120b~\cite{gptoss} as both the agent LLM and the integrated LLM, and Qwen3-VL-30B-A3B-Instruct~\cite{qwen3} for vision–language reasoning.
\vspace{-3mm}
\paragraph{Metrics}
Following~\cite{yang2025sceneweaver}, we evaluate the visual quality and physical plausibility with scores averaged across 10 generated scenes per room type.
Visual metrics cover Realism, Functionality, Layout, and Completeness based on GPT-4.1~\cite{openai2024api}.
Physical validity includes the collision ratio of 3D object meshes using trimesh~\cite{trimesh}, and the ratio of stable objects within Isaac Sim~\cite{isaacsim}.
An object is considered unstable if its relative translation exceeds 0.2 meter or its rotation exceeds 8 degrees after 120 simulation steps.
\vspace{-3mm}
\paragraph{Baselines}
We compare against two state-of-the-art methods using their official codebases.
\textbf{Holodeck}~\cite{yang2024holodeck} is LLM-driven but lacks self-improvement due to its fixed generation pipeline.
\textbf{SceneWeaver}~\cite{yang2025sceneweaver} is agent-based yet omits simulator validation, yielding non–simulation-ready scenes.

\subsubsection{Experiment Results}

\paragraph{Common Types}
We evaluate  \model and the baselines on three common indoor scene types: Bedroom, Kitchen, and Living Room. 
Quantitative results are reported in Tab.~\ref{tab:exp_common}, with qualitative comparisons in Fig.~\ref{fig:qual_cmp}.
 \model achieves the best results across all metrics, in both visual quality and physical stability.
Holodeck~\cite{yang2024holodeck} generates fewer objects with lower realism and functionality, and exhibits frequent collisions due to its rigid, predefined generation pipeline.
SceneWeaver~\cite{yang2025sceneweaver} achieves moderate visual quality but exhibits high collision rates and low stability, mainly due to its absence of simulator-based validation.
As illustrated in Fig.~\ref{fig:stability}, baseline scenes show displaced or fallen objects after simulation, whereas  \model remains fully stable.
\vspace{-3mm}
\paragraph{Open-Vocabulary Types}
We further demonstrate the open-vocabulary generation capability of \model across a wide range of scene types. As shown in Fig.~\ref{fig:qual_cmp} and Fig.~\ref{fig:open_vocab}, our method successfully generates highly diverse and stylized spaces (\eg, Gym, Office, Cyberpunk game den, Starry-night bedroom).
Unlike retrieval-based methods, our text-to-3D object synthesis enables the creation of long-tail, semantically coherent scenes with faithful adherence to user prompts and distinctive visual styles.

\vspace{-3mm}
\paragraph{Ablation Study}
We conduct an ablation study to assess the impact of each critic design. As shown in Tab.~\ref{tab:critic_ablation}, adding the visual critic substantially improves visual quality, while the physics critic greatly reduces collisions from $7.8\%$ to $1.9\%$ and raises stability to $99.6\%$. Combining both critics yields the best overall performance across all metrics, confirming that visual feedback and simulation-in-the-loop validation are complementary and critical for generating realistic and physically stable scenes.

\paragraph{SAGE-10k Dataset}
To support community research at scale, we pre-generated a 10k-scene dataset with our approach, named SAGE-10k Dataset. It's across 50 room types in diverse styles, including 565K uniquely generated 3D objects. The preview image and statistics of the dataset are shown in Fig.~\ref{fig:sage10k}.

\subsubsection{Additional Capabilities and Extensions}
In this subsection, we present several straightforward extensions of \model that demonstrate its flexibility across diverse generation settings. While these capabilities are not the primary focus of this paper, they arise naturally from the modular design of our framework and require minimal adaptation. These results highlight the generality of \model and suggest that further improvements in generation quality can be achieved by strengthening individual components, which we leave as a promising direction for future work.

\paragraph{Multi-room Layout}
We showcase the multi-room generation capability of \model across different kinds of scene layouts, as Fig.~\ref{fig:open_vocab_multi_room} shows. Connected floor plans are created in the {Scene Initializer}. Additionally, the agent can execute parallel updates with {Asset Placer/Mover/Remover}, and the generators are updated to accept multiple room IDs and conditions within a single MCP tool call.

\paragraph{Image-conditioned Scene Generation}
We demonstrate that \model can be extended with image-conditioned generation without architectural changes by utilizing Qwen3-VL \cite{qwen3} to extract style and object attributes from reference images, as shown in Fig. \ref{fig:open_vocab_image_ref}. While not pixel-aligned, the generated scenes remain semantically coherent.

\paragraph{Articulated Objects}
In addition to rigid-body objects, our modular design can easily extend the current text-to-3D generation with object retrieval. As Fig. \ref{fig:art} shows, \model can be integrated with articulated assets from PartNet-Mobility \cite{xiang2020sapien} in the generated scenes. Also, robot actions can be generated by grasp pose prediction and motion planning.

\subsection{Embodied AI Learning}
\label{sec:result_learning}
We evaluate whether the diversity of our generated scenes enables effective scaling by measuring policy improvements under increasing scene and demonstration counts.

\subsubsection{Setup}
\paragraph{Implementation Details} We study two representative tasks: (1)
Pick-and-Place with a Franka Emika Panda robot, (2) Mobile Manipulation using a Franka Emika Panda arm mounted on an Omron LD-60 mobile base~\cite{robocasa, robosuite}. 
All action data generation and policy rollout are conducted in Isaac Lab~\cite{isaaclab} on NVIDIA L40S GPUs with parallel simulation.
Policies are trained via the imitation learning framework Robomimic~\cite{robomimic}. Pick-and-Place uses a single model, whereas the long-horizon Mobile Manipulation is decomposed into four sequential policies. Each policy is responsible for one stage of the task (see task description) and also predicts a termination signal for stage transition.
We report success rates averaged across three held-out scenes, each evaluated with 90 random robot spawning poses and object configurations.
\vspace{-3mm}
\paragraph{Task Description}
For Pick-and-Place, the robot must pick up a mug from the table and place it into a bowl.
For Mobile Manipulation, the robot is initialized at a random position in the scene. It needs to navigate to a table with a coke can, pick it up, move to another desk, and place the can on it.
\vspace{-3mm}
\paragraph{Scene and Action Generation}
 \model can automatically generate large-scale scene and action data starting from the task descriptions.
Object-level and scene-level augmentation is applied to Pick-and-Place and Mobile Manipulation, respectively.
We collect over 28k demonstrations for Pick-and-Place with 264 unique objects
and nearly 50k demonstrations for Mobile Manipulation across 50 diverse scenes.

\subsubsection{Policy Learning}

\paragraph{Scaling Trend with \model-generated Data}
As shown in Fig.~\ref{fig:robot_graph_pnp} and Fig.~\ref{fig:robot_graph_mm}, policy success rates increase with the number of scenes. Despite that the learned policy only has access to partial visual observations, it gradually converges towards the privileged agent which has full access to the entire 3D scene information.
To contextualize performance relative to task difficulty, we also compare the success rate between policy rollout and motion planning in Tab.~\ref{tab:train_mp_metric}.
Our results show that the learned policy achieves success rates approaching those of the underlying motion planner, confirming both the fidelity of the generated data and the effectiveness of imitation learning at scale.

\vspace{-4mm}
\paragraph{Comparison with Baselines}
Since existing methods rely on heterogeneous simulation setups and implementations, directly integrating them into a unified environment is nontrivial. To enable fair comparison, we design two baselines that mimic the methodology of prior works by removing components unique to our framework and substituting them with their counterparts.
As shown in Fig.~\ref{fig:robot_graph_mm}, both baselines improve slowly with scale--achieving less than one-third of the final success rate of the policy trained on \model-generated data, even with the same number of demonstrations and scenes.
To further demonstrate the generalizability of our trained policy, we test on baseline-generated held-out scenes as in Fig. \ref{fig:robot_graph_mm}. As the results in Tab. \ref{tab:cross_eval}, despite a distribution shift favoring baselines, our trained policies achieve higher success rates, indicating robust generalization.
This comparison highlights that the unique design of \model, including agentic orchestration, physics validation, and text-to-3D synthesis, is critical for generating effective training data. Our method not only converges significantly faster than the baselines but also approaches the performance of the privileged agent, demonstrating that high-quality, simulation-validated scenes can efficiently drive policy learning. These results suggest a promising path toward scalable, simulation-driven learning for embodied AI.

\section{Conclusion}

We present \model, an agentic framework that turns open-vocabulary text prompts into \emph{simulation-ready} indoor environments by orchestrating layout and asset generators with visual and physics critics via MCP. The same pipeline scales them for embodied learning through multi-level scene augmentation and automatic action synthesis. On two representative tasks, \ie, Pick-and-Place and Mobile Manipulation, policies trained purely on our generated data show clear scaling with scene diversity and demonstration count, improving generalization to unseen objects and layouts.

\vspace{-4mm}
\paragraph{Limitations and future work.}
Our current scope emphasizes indoor scenes and rigid-body physics, and extending to outdoor settings as well as articulated and deformable objects is promising. Action generation presently targets flexible compositions of pick, place, and navigation. Incorporating additional tasks is a natural next step. Beyond imitation, coupling the generator with online RL and real-robot closed-loop validation could further boost performance.

{
    \small
    \bibliographystyle{ieeenat_fullname}
    \bibliography{main}
}
\clearpage

\maketitlesupplementary

\setcounter{section}{0}
\setcounter{footnote}{0}
\setcounter{figure}{0}
\setcounter{table}{0}

\appendix
\section*{Appendix}

\begin{table*}[t]
\centering
\resizebox{\textwidth}{!}{
\setlength{\tabcolsep}{0pt}
\begin{tabular}{@{}cccc@{}}
\adjustbox{valign=c}{\includegraphics[width=0.35\linewidth,trim=120 0 120 0, clip]{figs/qual_vis_cmp/ours/bedroom_2.jpg}} &
\adjustbox{valign=c}{\includegraphics[width=0.35\linewidth,trim=120 0 120 0, clip]{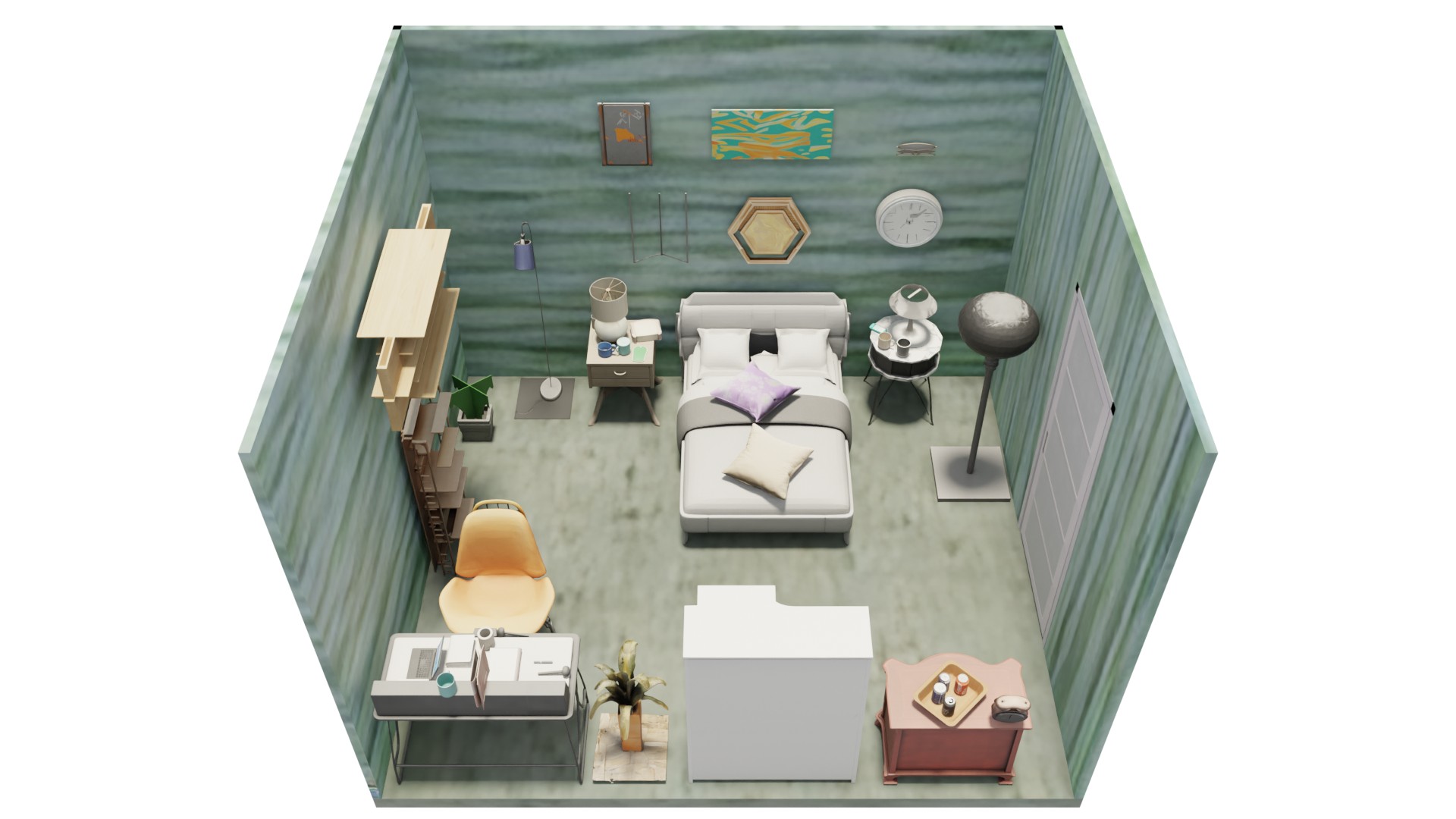}} &
\adjustbox{valign=c}{\includegraphics[width=0.35\linewidth,trim=120 0 120 0, clip]{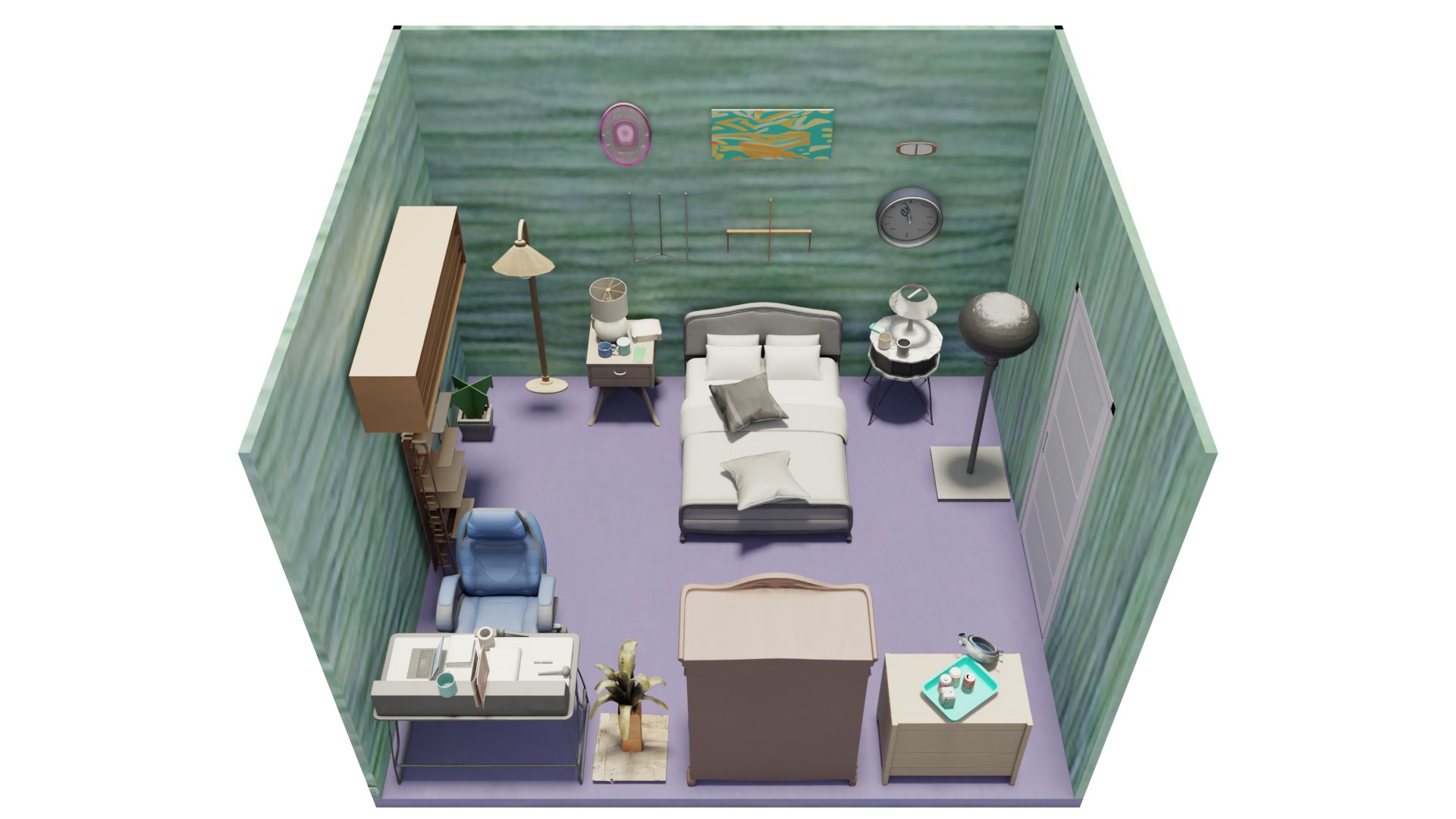}} &
\adjustbox{valign=c}{\includegraphics[width=0.35\linewidth,trim=120 0 120 0, clip]{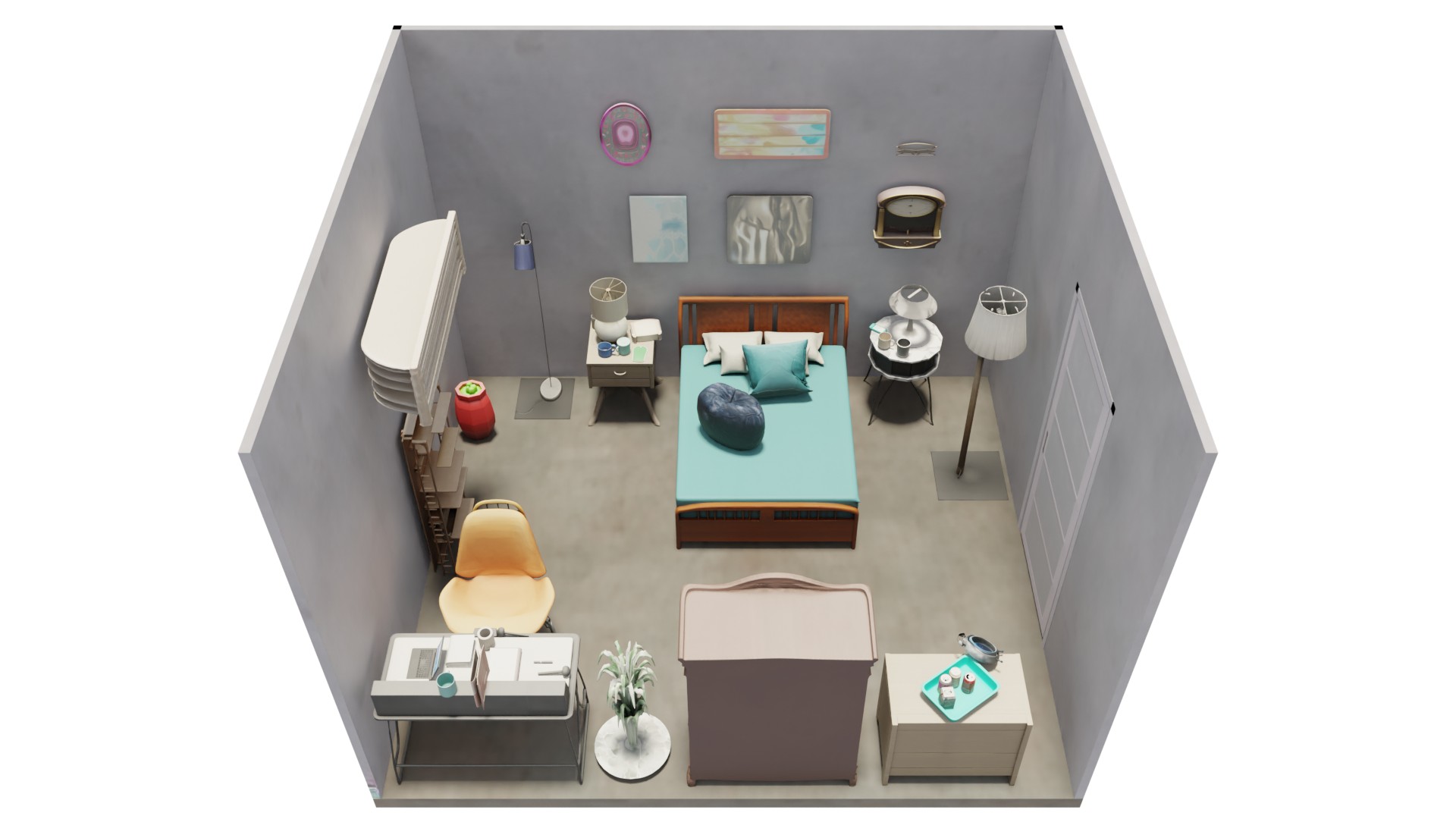}} \\

\adjustbox{valign=c}{\includegraphics[width=0.35\linewidth,trim=120 0 120 0, clip]{figs/qual_vis_cmp/ours/livingroom_2.jpg}} &
\adjustbox{valign=c}{\includegraphics[width=0.35\linewidth,trim=120 0 120 0, clip]{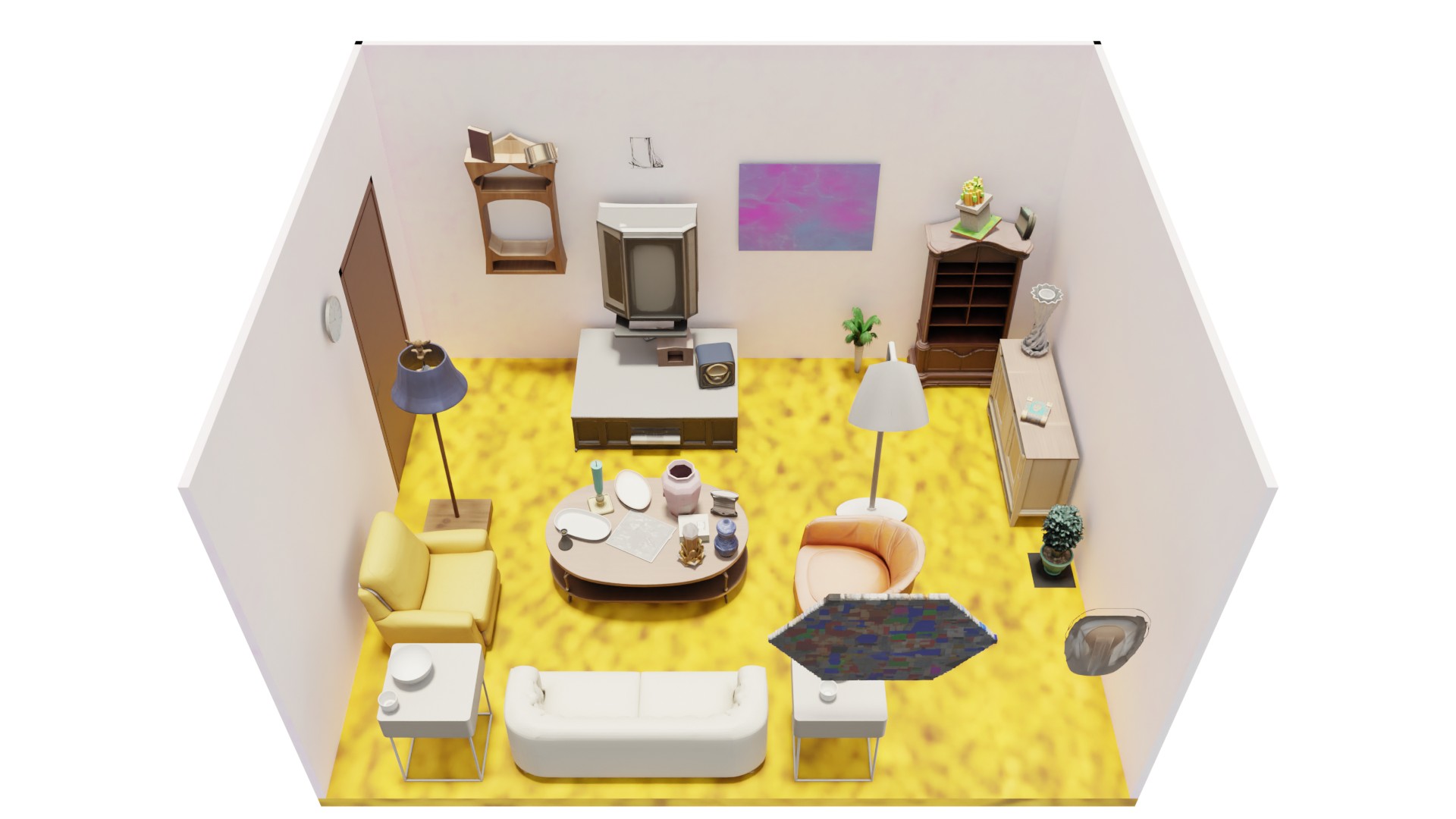}} &
\adjustbox{valign=c}{\includegraphics[width=0.35\linewidth,trim=120 0 120 0, clip]{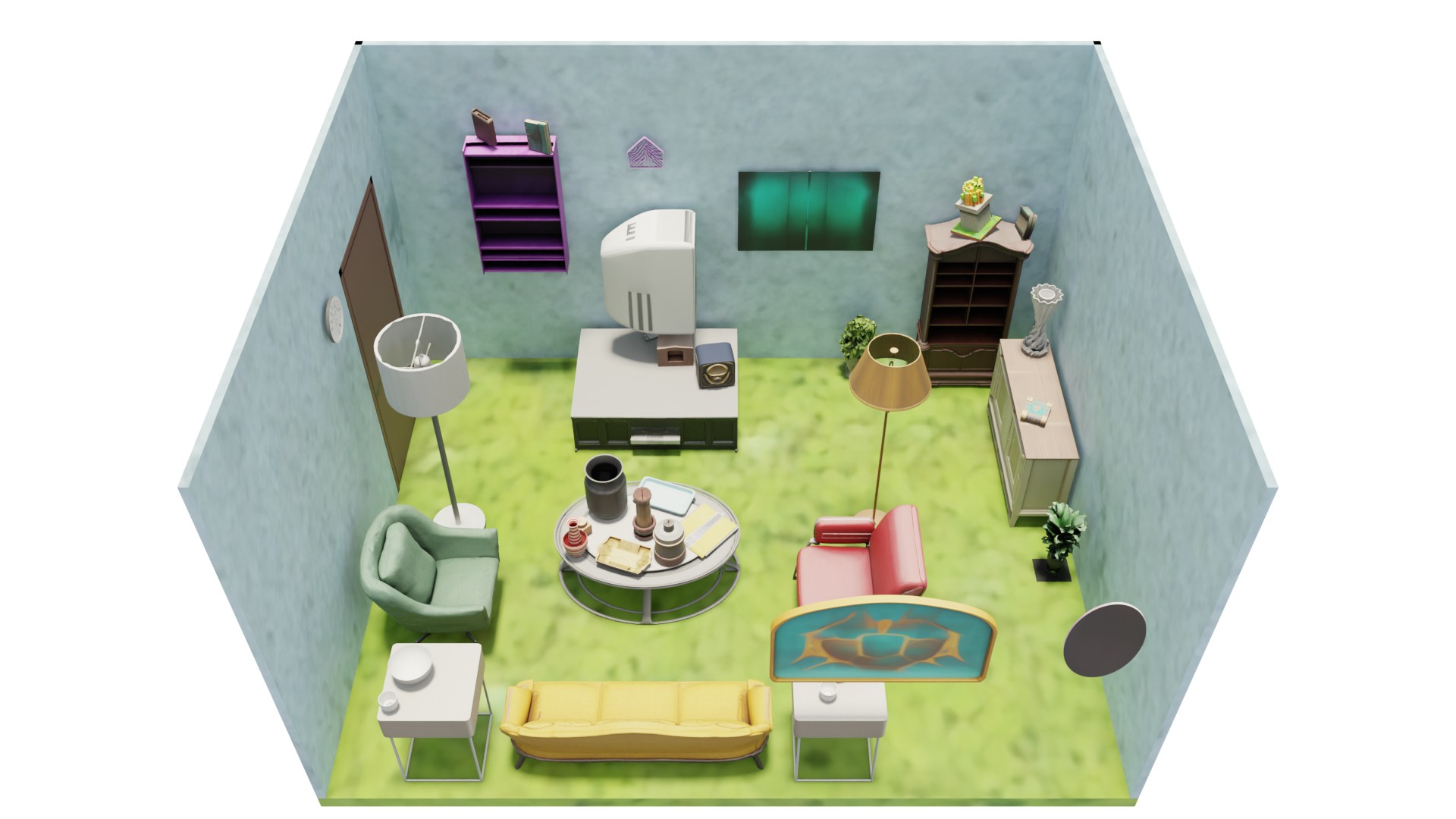}} &
\adjustbox{valign=c}{\includegraphics[width=0.35\linewidth,trim=120 0 120 0, clip]{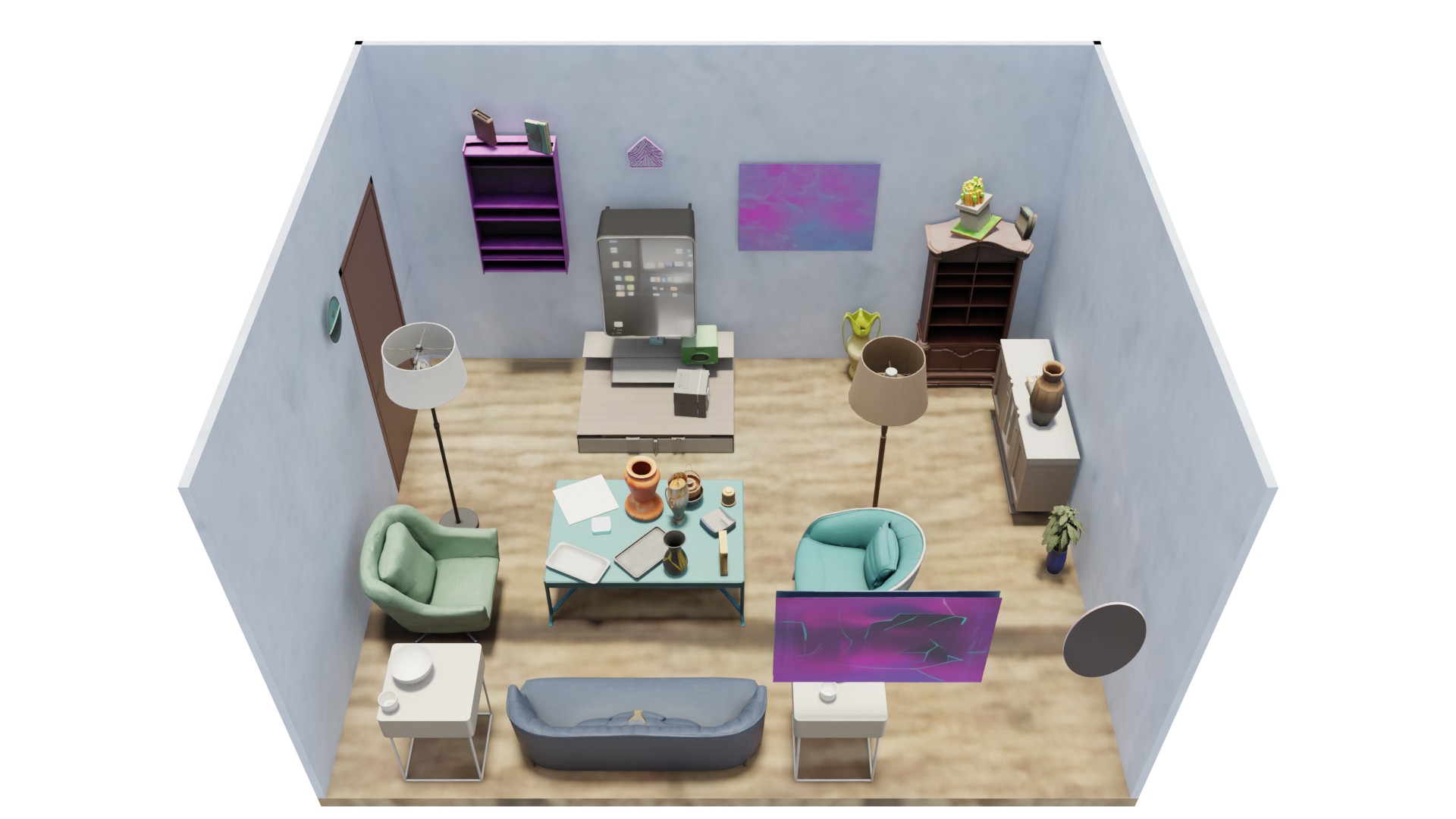}} \\

\adjustbox{valign=c}{\includegraphics[width=0.35\linewidth,trim=120 0 120 0, clip]{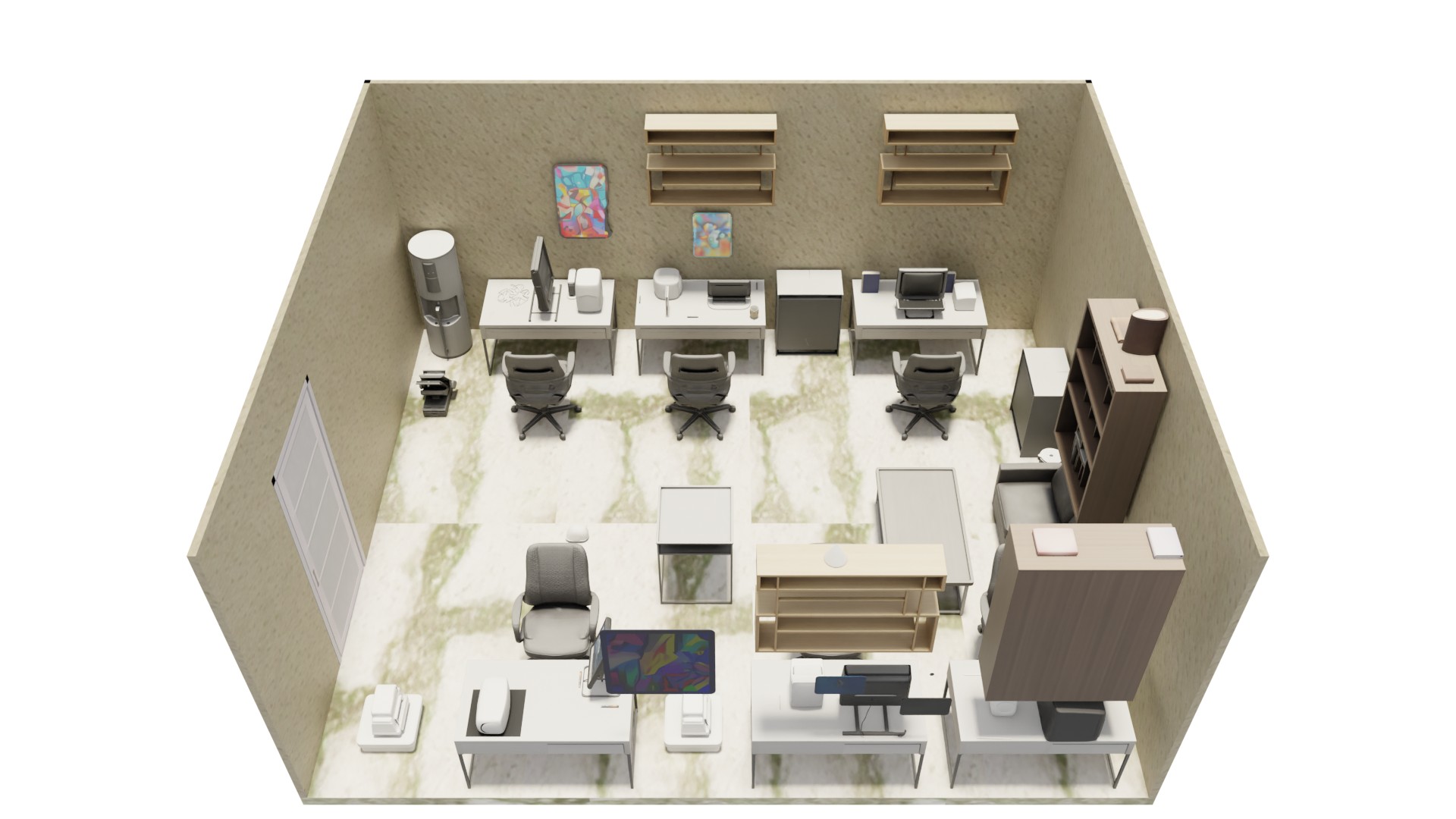}} &
\adjustbox{valign=c}{\includegraphics[width=0.35\linewidth,trim=120 0 120 0, clip]{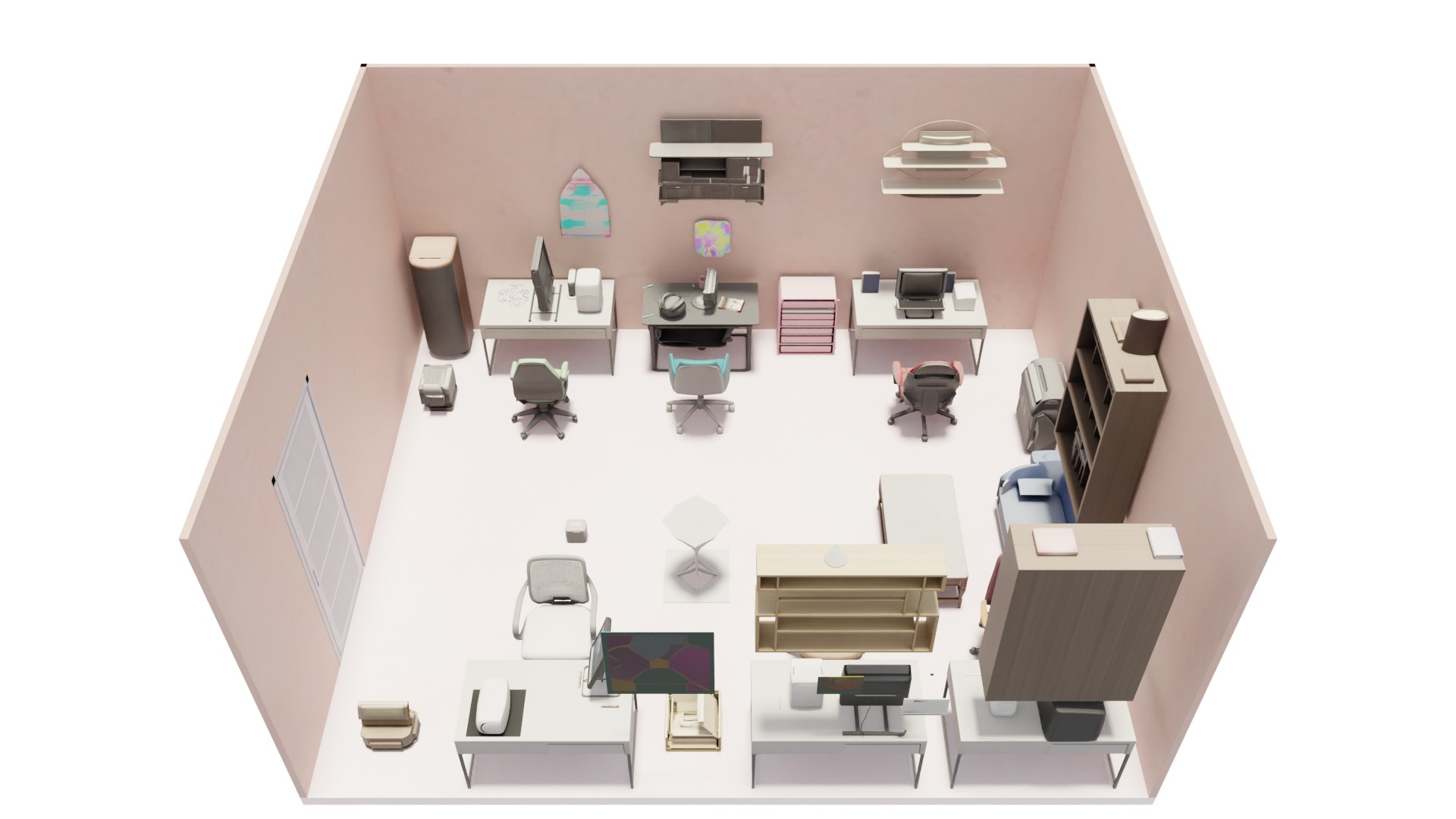}} &
\adjustbox{valign=c}{\includegraphics[width=0.35\linewidth,trim=120 0 120 0, clip]{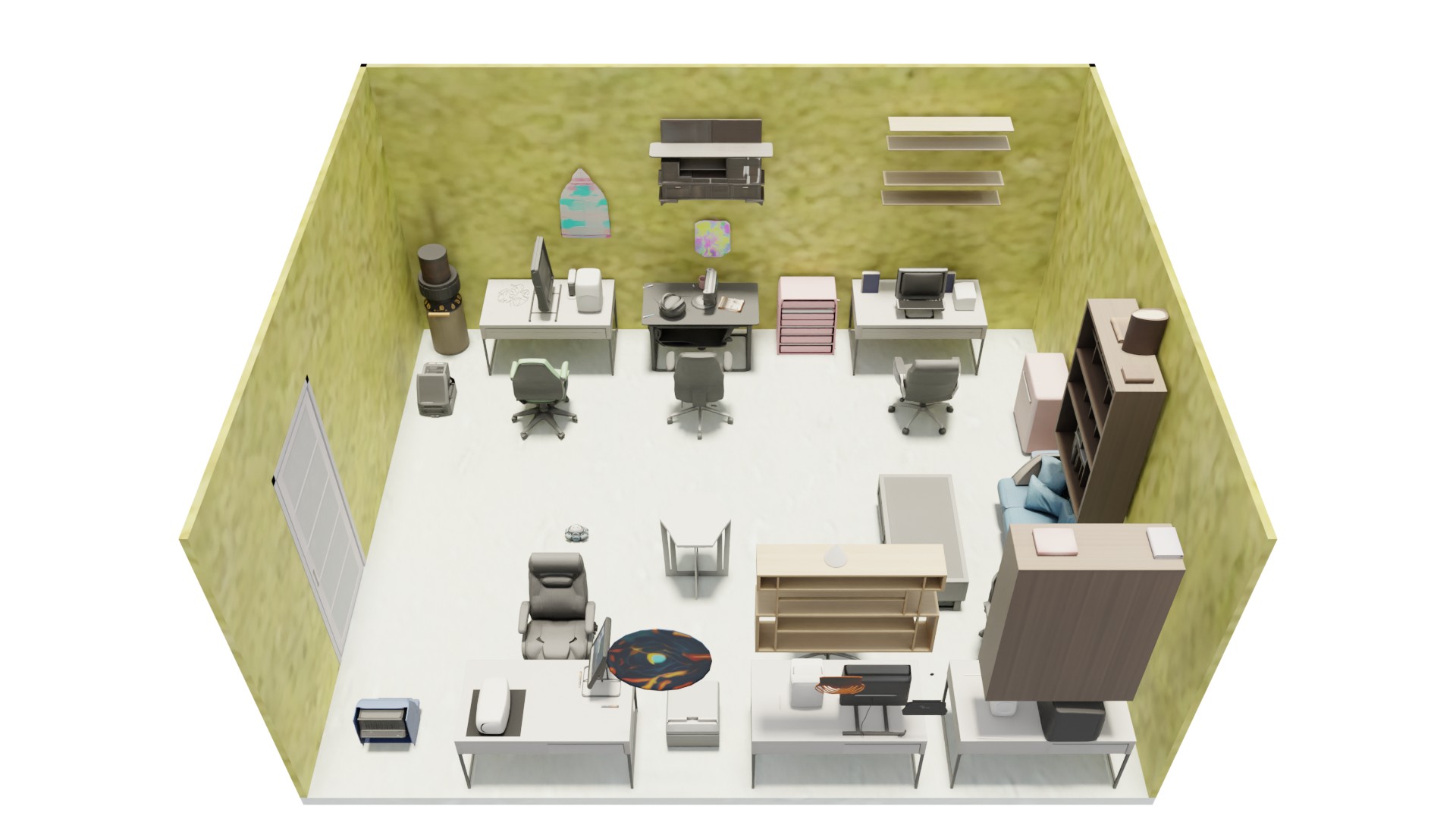}} &
\adjustbox{valign=c}{\includegraphics[width=0.35\linewidth,trim=120 0 120 0, clip]{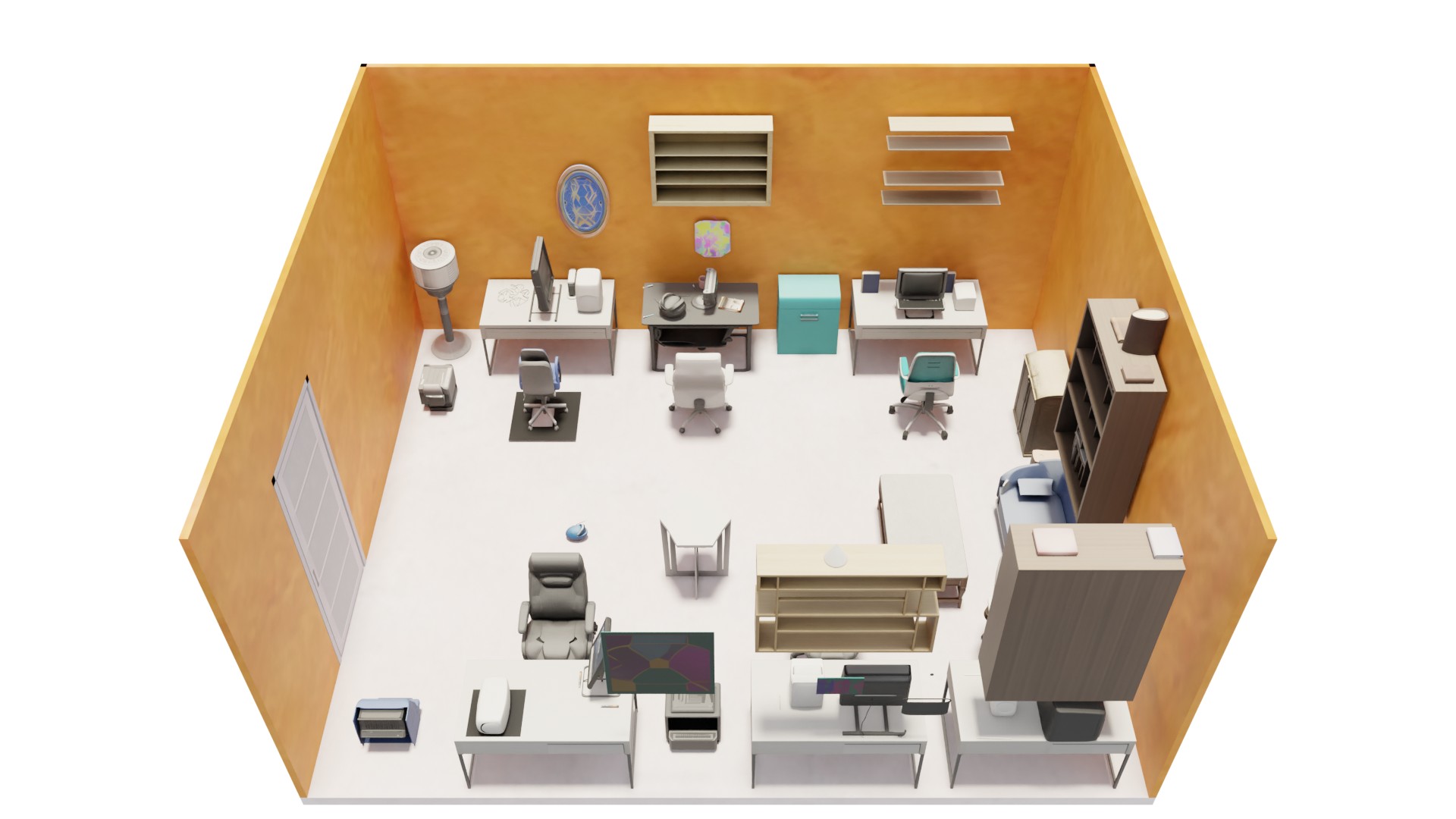}} \\

\adjustbox{valign=c}{\includegraphics[width=0.35\linewidth,trim=120 0 120 0, clip]{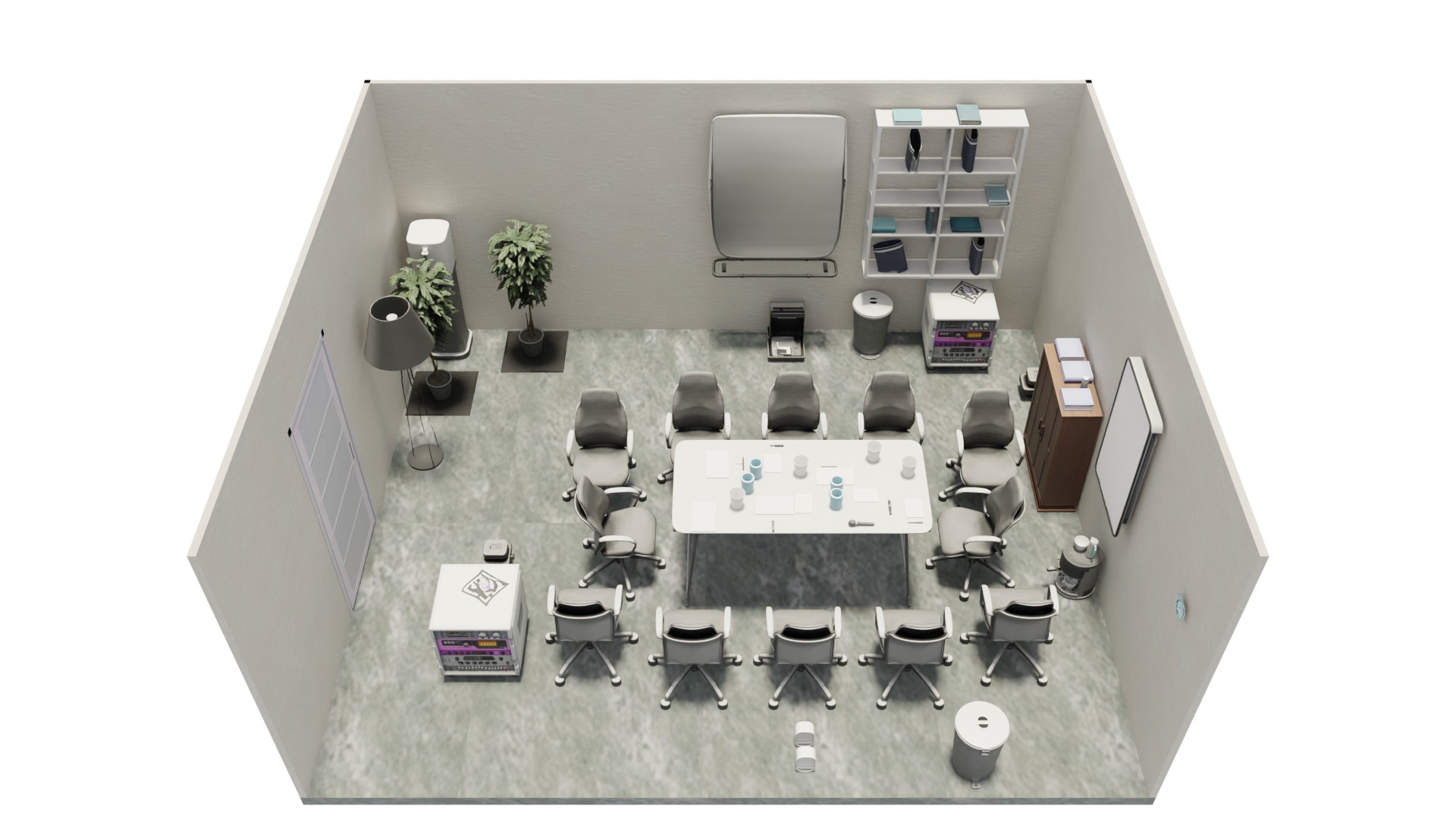}} &
\adjustbox{valign=c}{\includegraphics[width=0.35\linewidth,trim=120 0 120 0, clip]{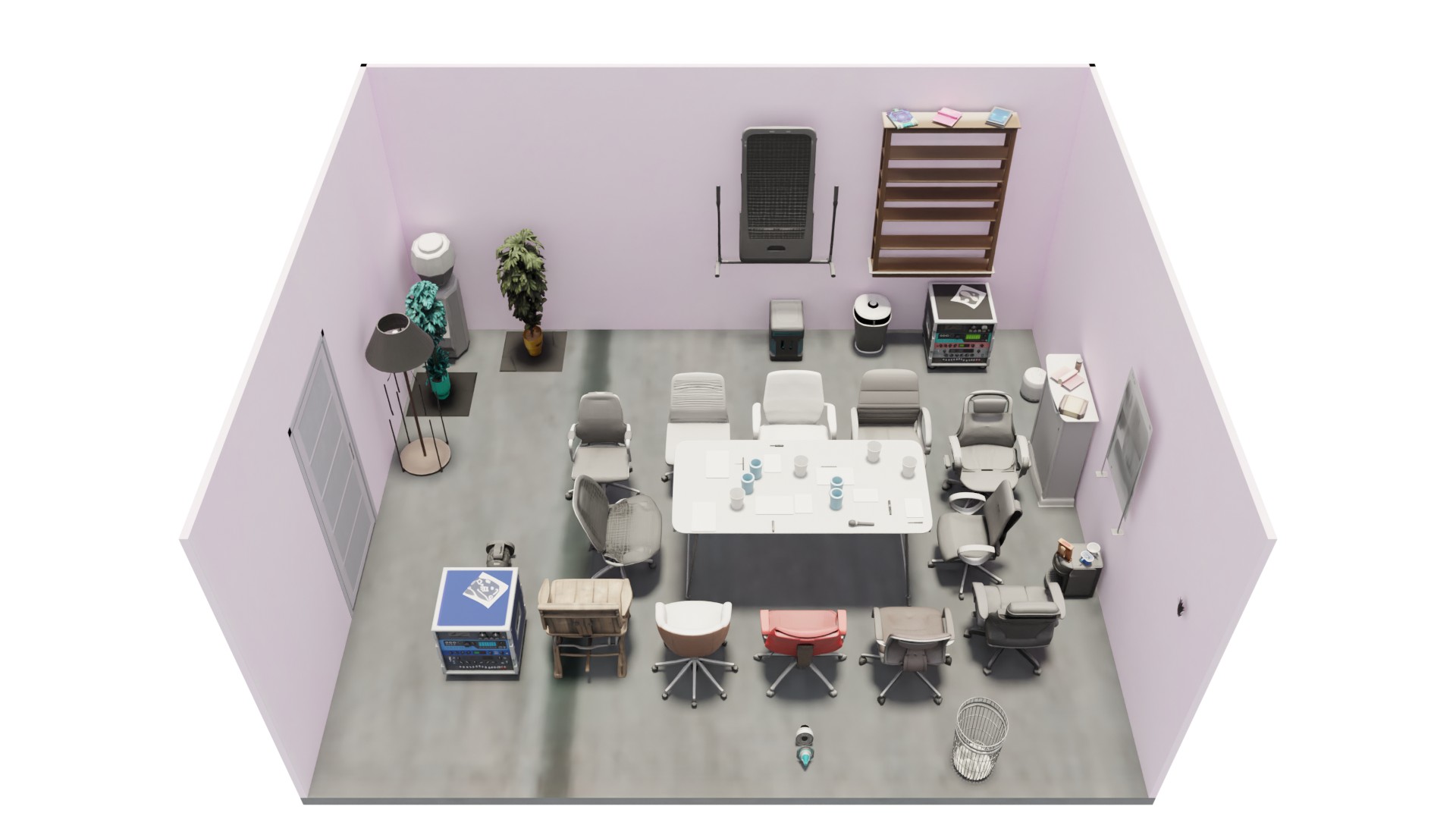}} &
\adjustbox{valign=c}{\includegraphics[width=0.35\linewidth,trim=120 0 120 0, clip]{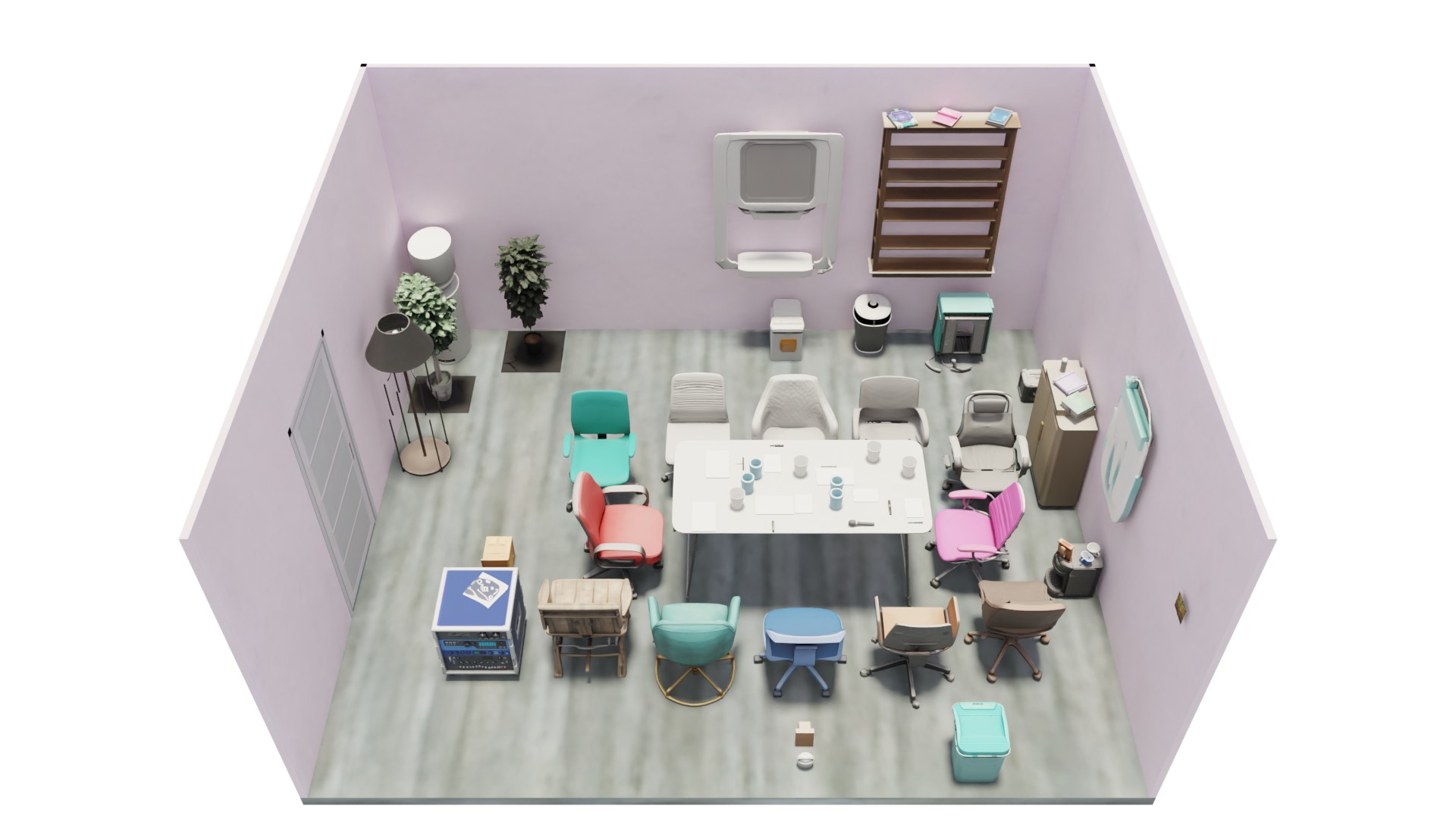}} &
\adjustbox{valign=c}{\includegraphics[width=0.35\linewidth,trim=120 0 120 0, clip]{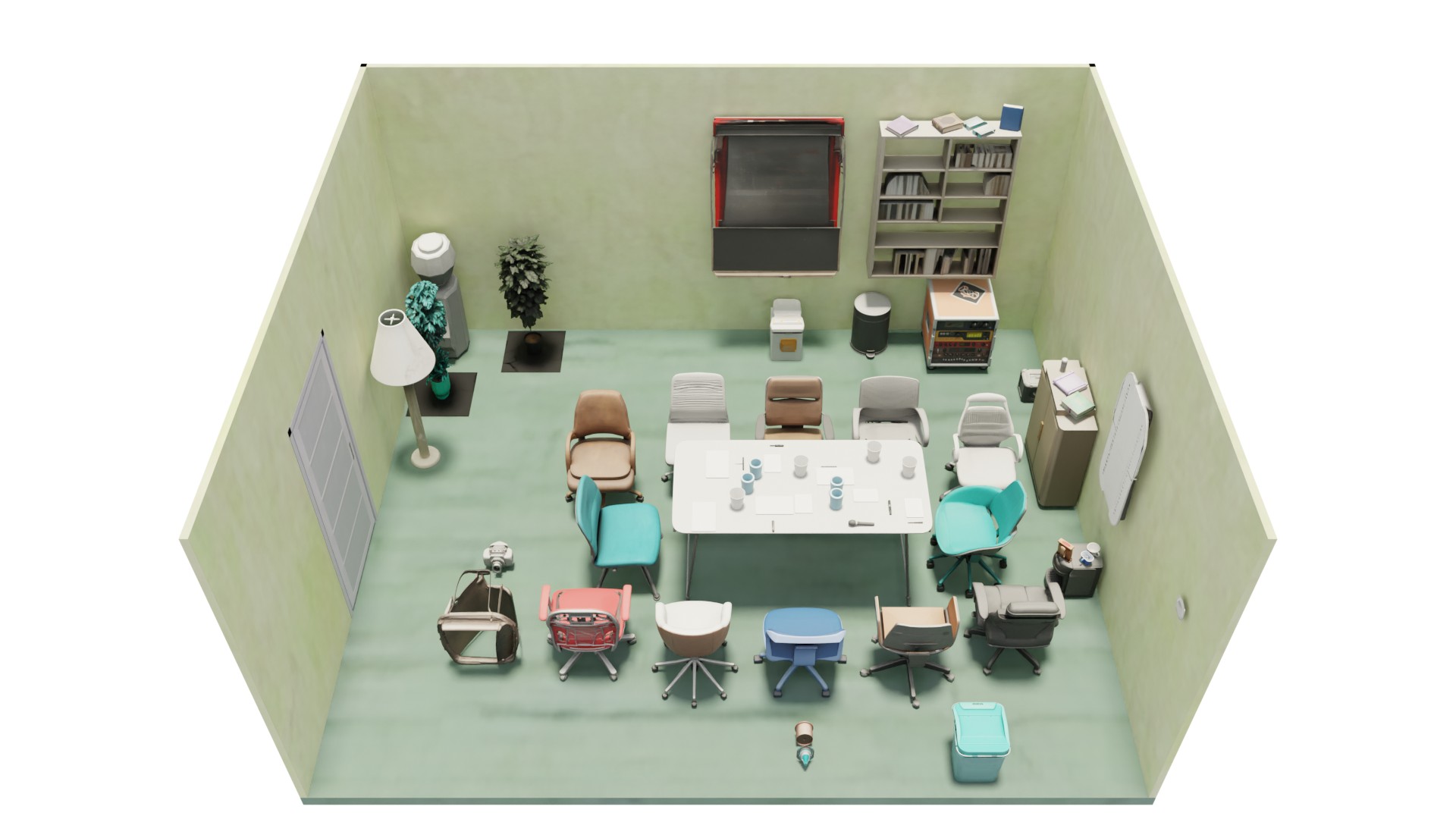}} \\
Base Scene & Aug. Scene 1 &  Aug. Scene 2 &  Aug. Scene 3 \\

\end{tabular}
}
\captionof{figure}{\textbf{Object Category-Level Augmentation.} Here we showcase the capability of our category augmentation method. We randomly select part of the objects in the scene for category augmentation. Given the text description of the selected object from the generation stage, we employ an LLM-based text augmentation to produce variations in geometry and texture (\eg, shape, color, material, or finish) while maintaining the original object category. We then use TRELLIS \cite{trellis} to synthesize corresponding 3D assets from these augmented descriptions, which are placed into the scene to enrich visual and physical diversity across instances. 
}
\vspace{-5mm}
\label{fig:cat_aug}
\end{table*}
\begin{table*}[t]
\centering
\resizebox{\textwidth}{!}{
\setlength{\tabcolsep}{0pt}
\begin{tabular}{@{}cccc@{}}
\adjustbox{valign=c}{\includegraphics[width=0.35\linewidth,trim=120 0 120 0, clip]{figs/qual_vis_cmp/ours/bedroom_2.jpg}} &
\adjustbox{valign=c}{\includegraphics[width=0.35\linewidth,trim=120 0 120 0, clip]{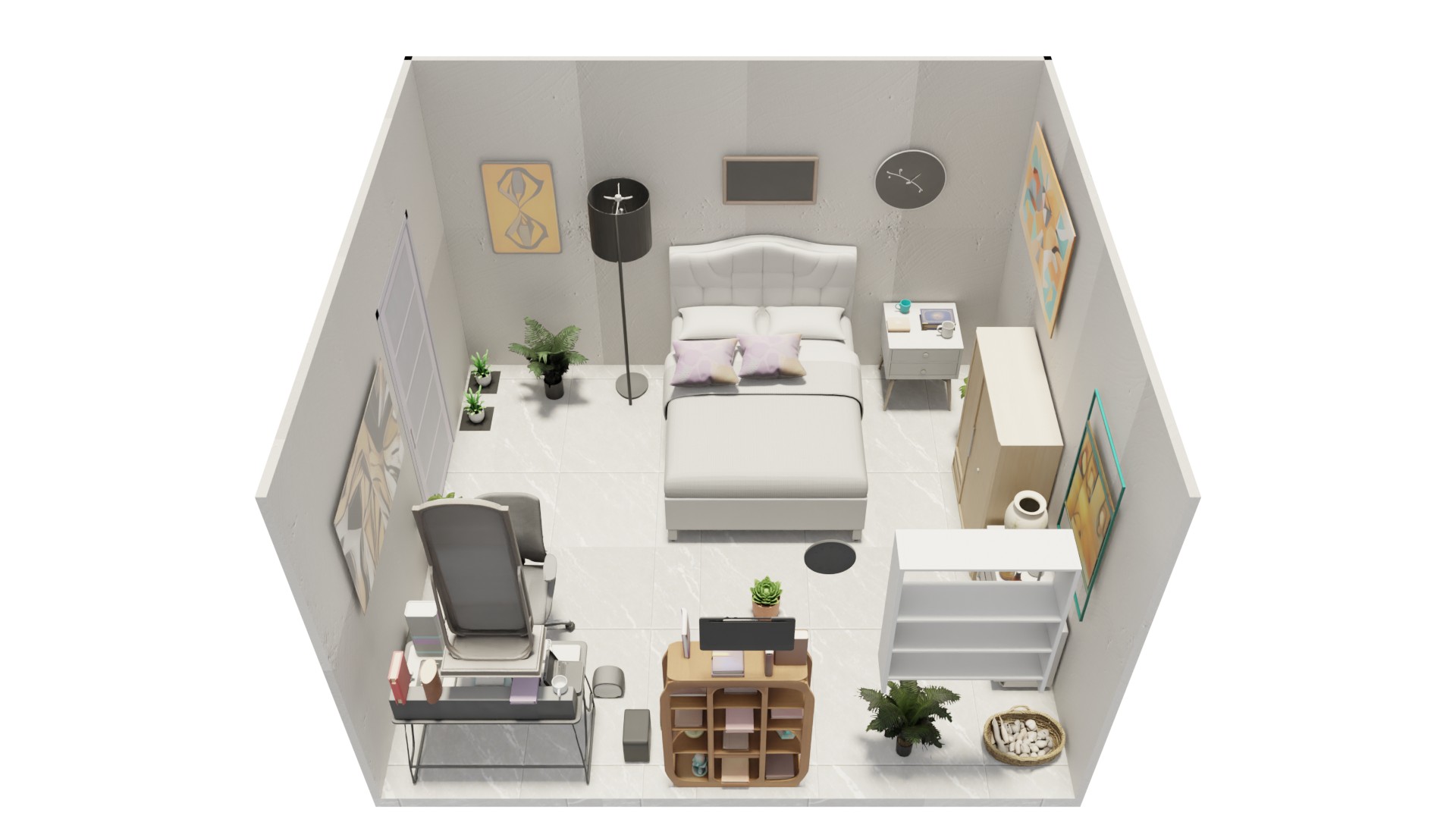}} &
\adjustbox{valign=c}{\includegraphics[width=0.35\linewidth,trim=120 0 120 0, clip]{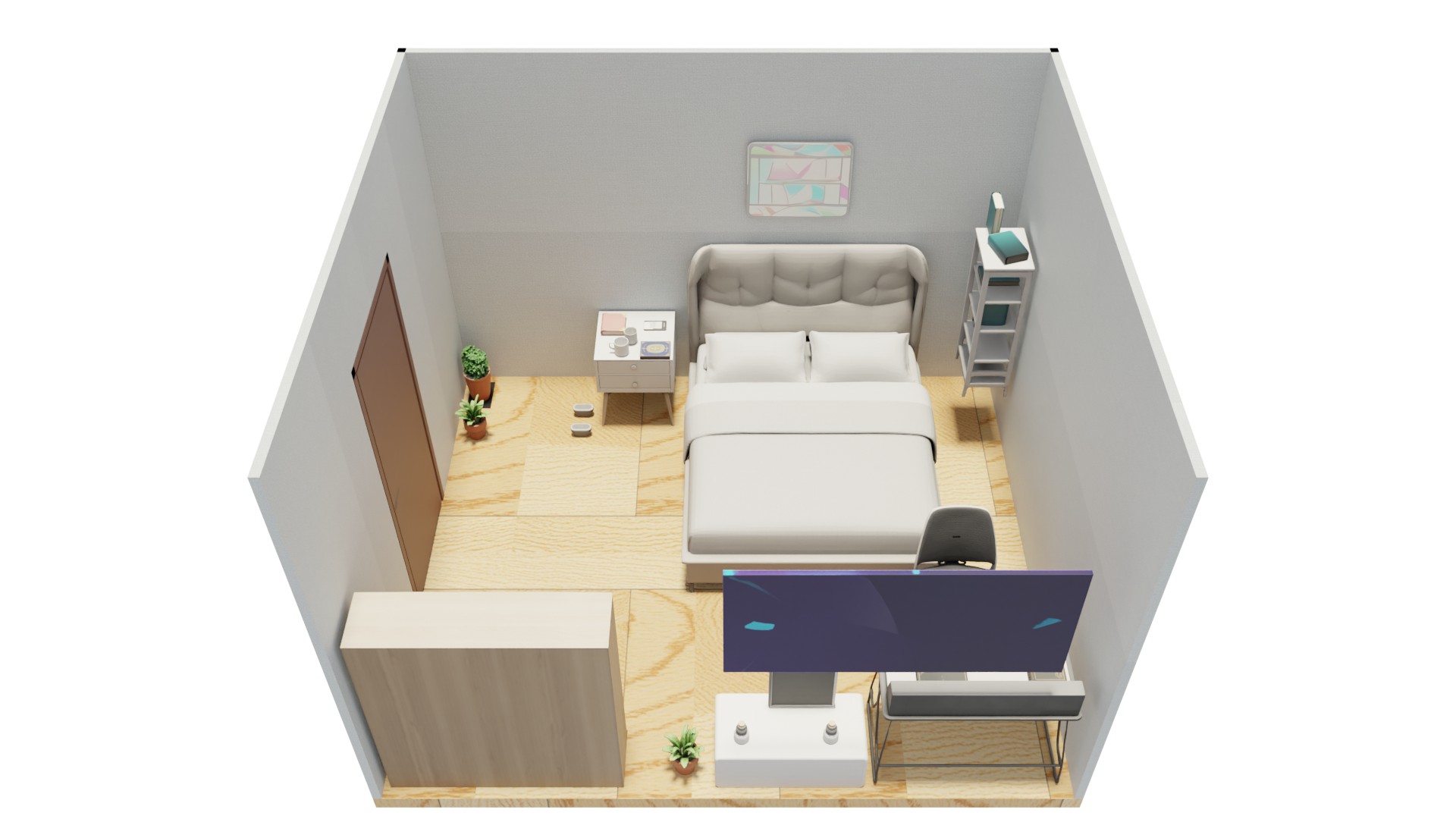}} &
\adjustbox{valign=c}{\includegraphics[width=0.35\linewidth,trim=120 0 120 0, clip]{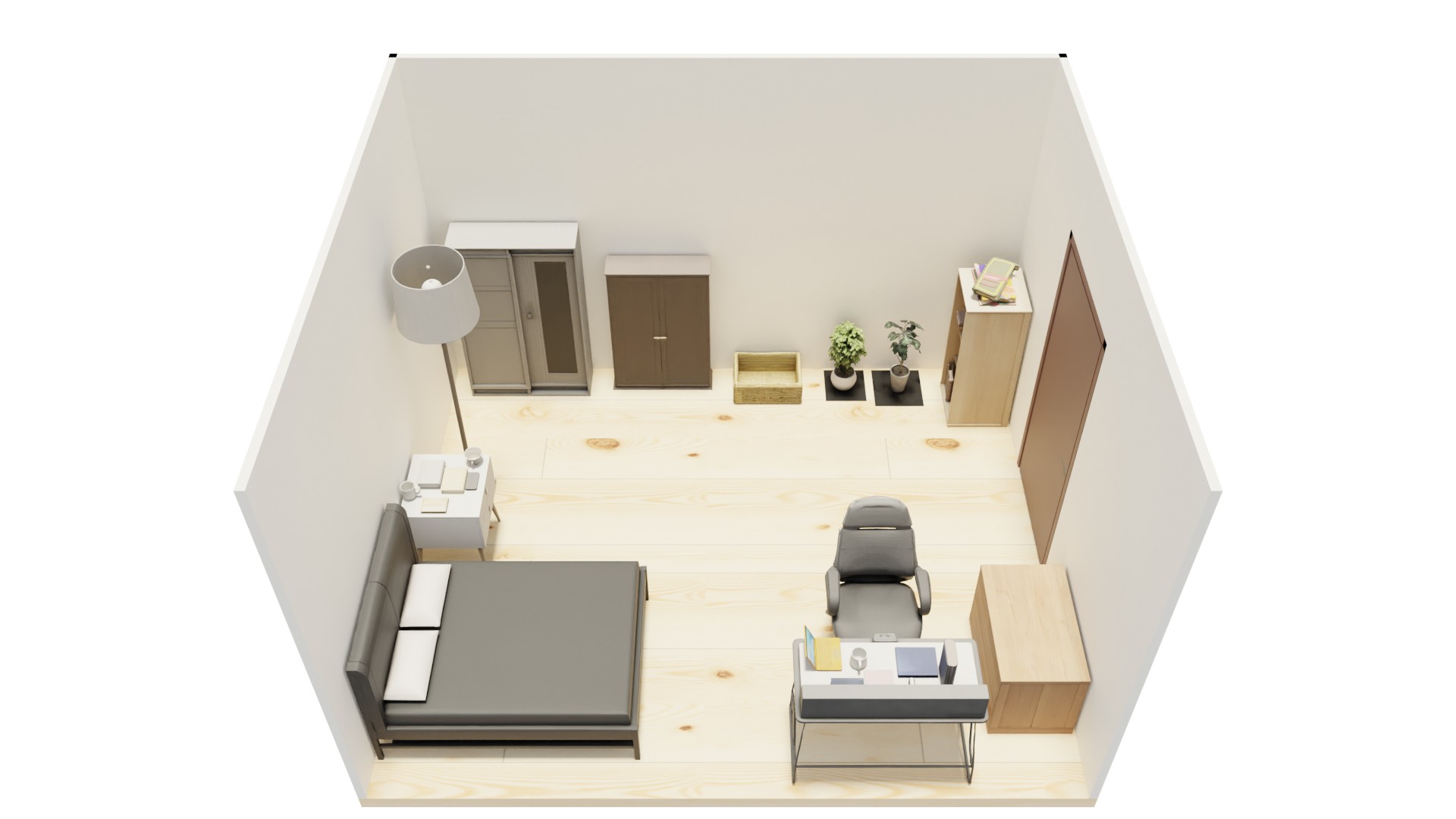}} \\

\adjustbox{valign=c}{\includegraphics[width=0.35\linewidth,trim=120 0 120 0, clip]{figs/qual_vis_cmp/ours/livingroom_2.jpg}} &
\adjustbox{valign=c}{\includegraphics[width=0.35\linewidth,trim=120 0 120 0, clip]{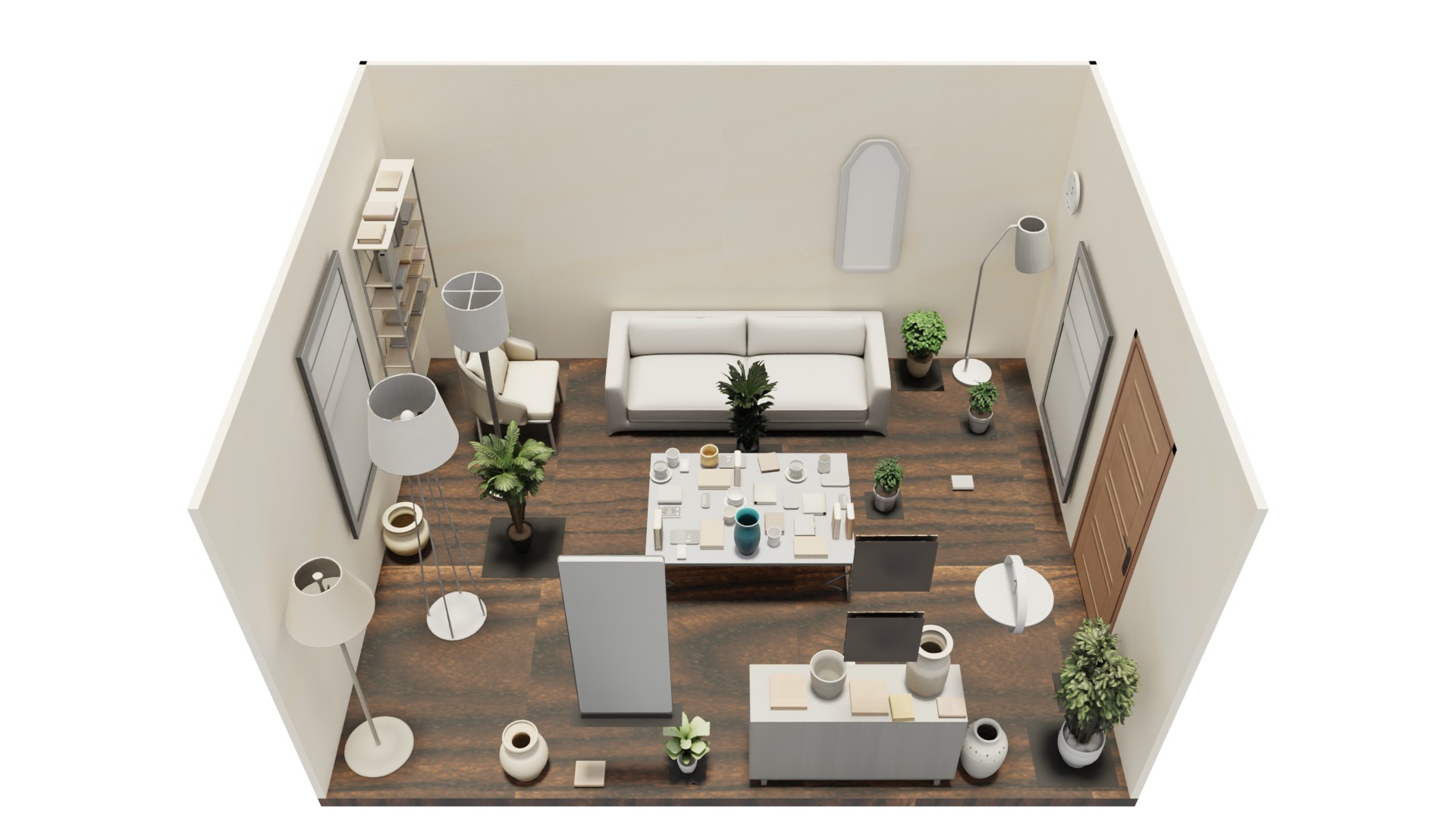}} &
\adjustbox{valign=c}{\includegraphics[width=0.35\linewidth,trim=120 0 120 0, clip]{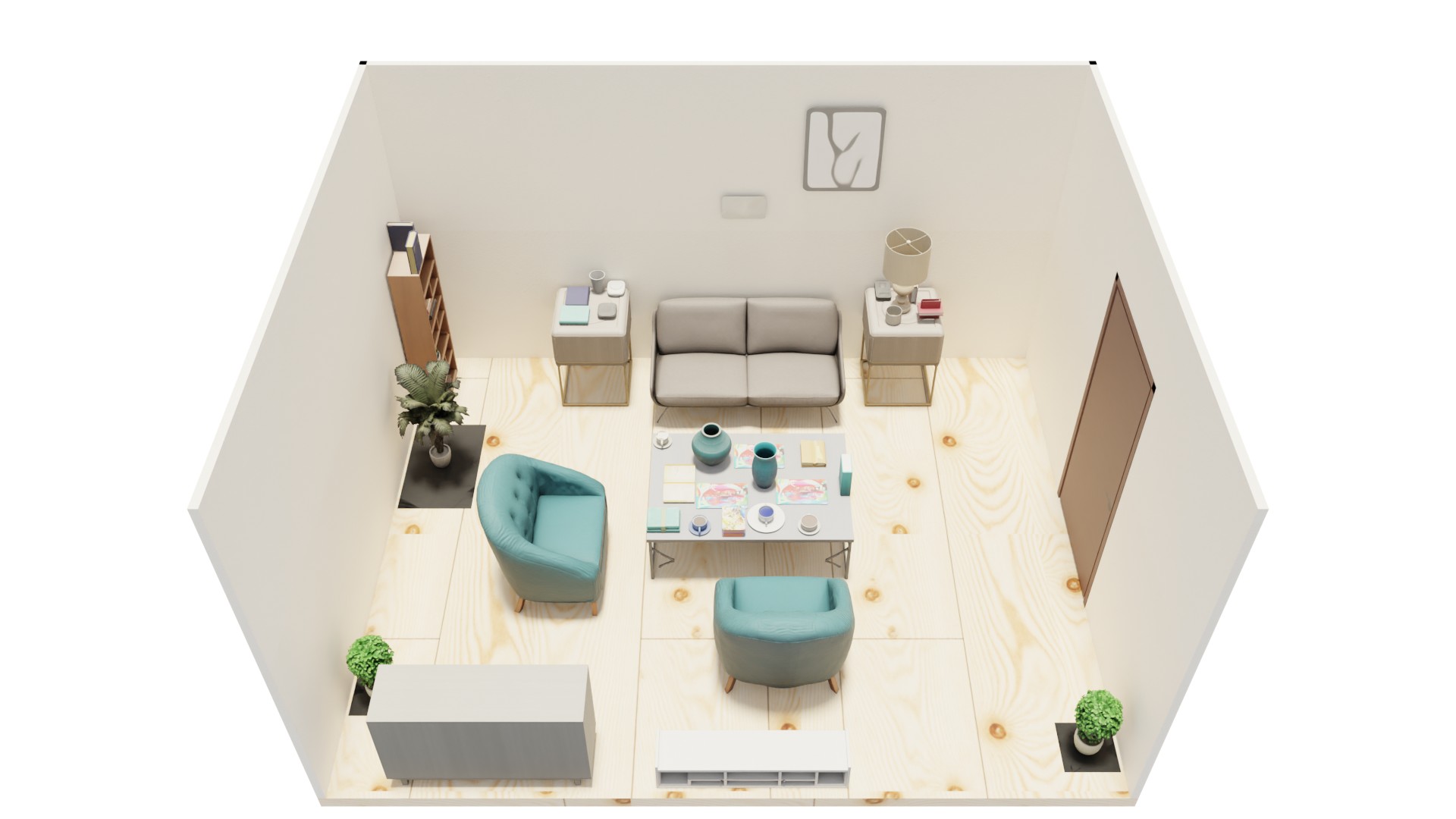}} &
\adjustbox{valign=c}{\includegraphics[width=0.35\linewidth,trim=120 0 120 0, clip]{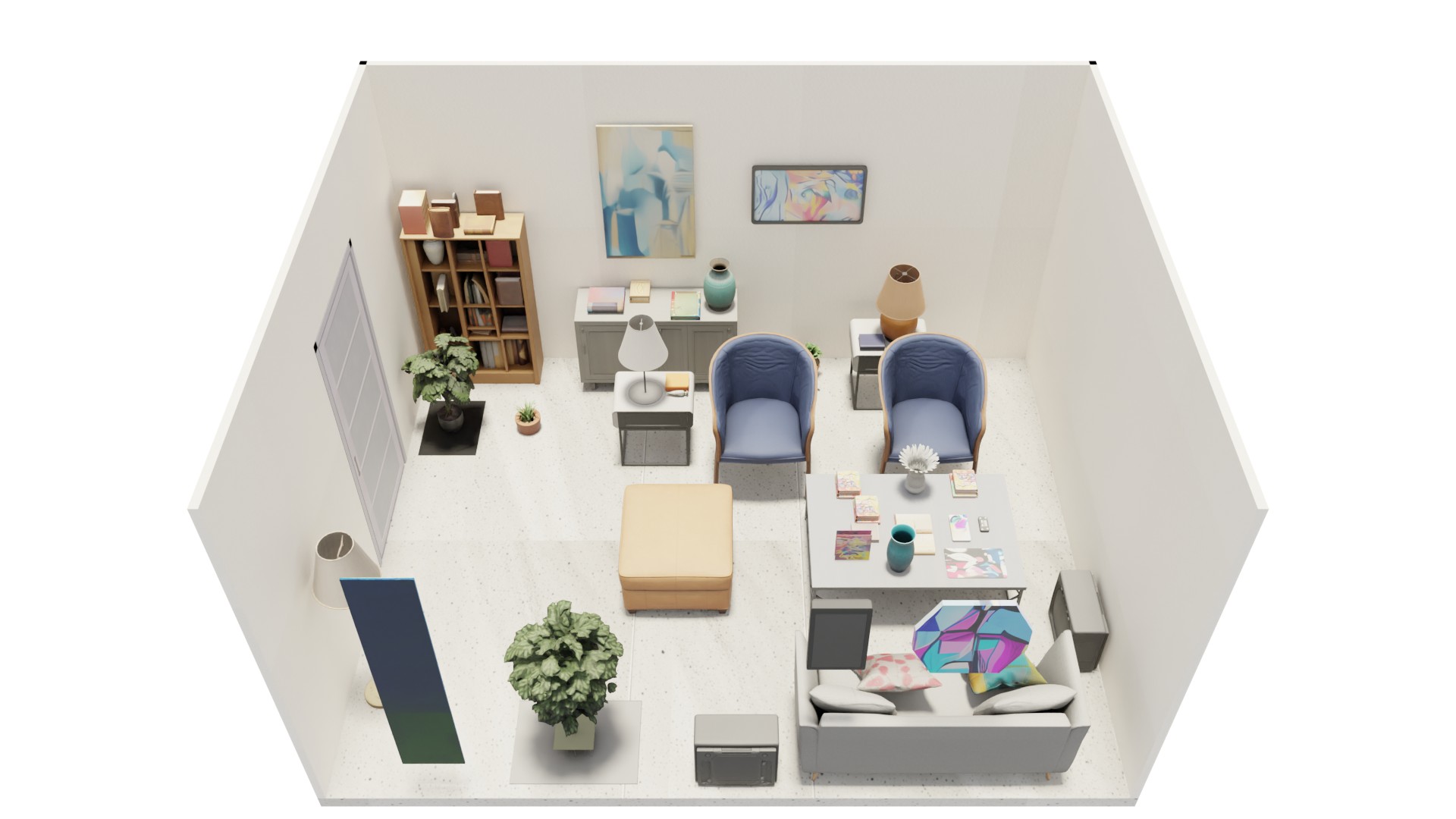}} \\

\adjustbox{valign=c}{\includegraphics[width=0.35\linewidth,trim=120 0 120 0, clip]{figs/object_aug/office_base.jpg}} &
\adjustbox{valign=c}{\includegraphics[width=0.35\linewidth,trim=120 0 120 0, clip]{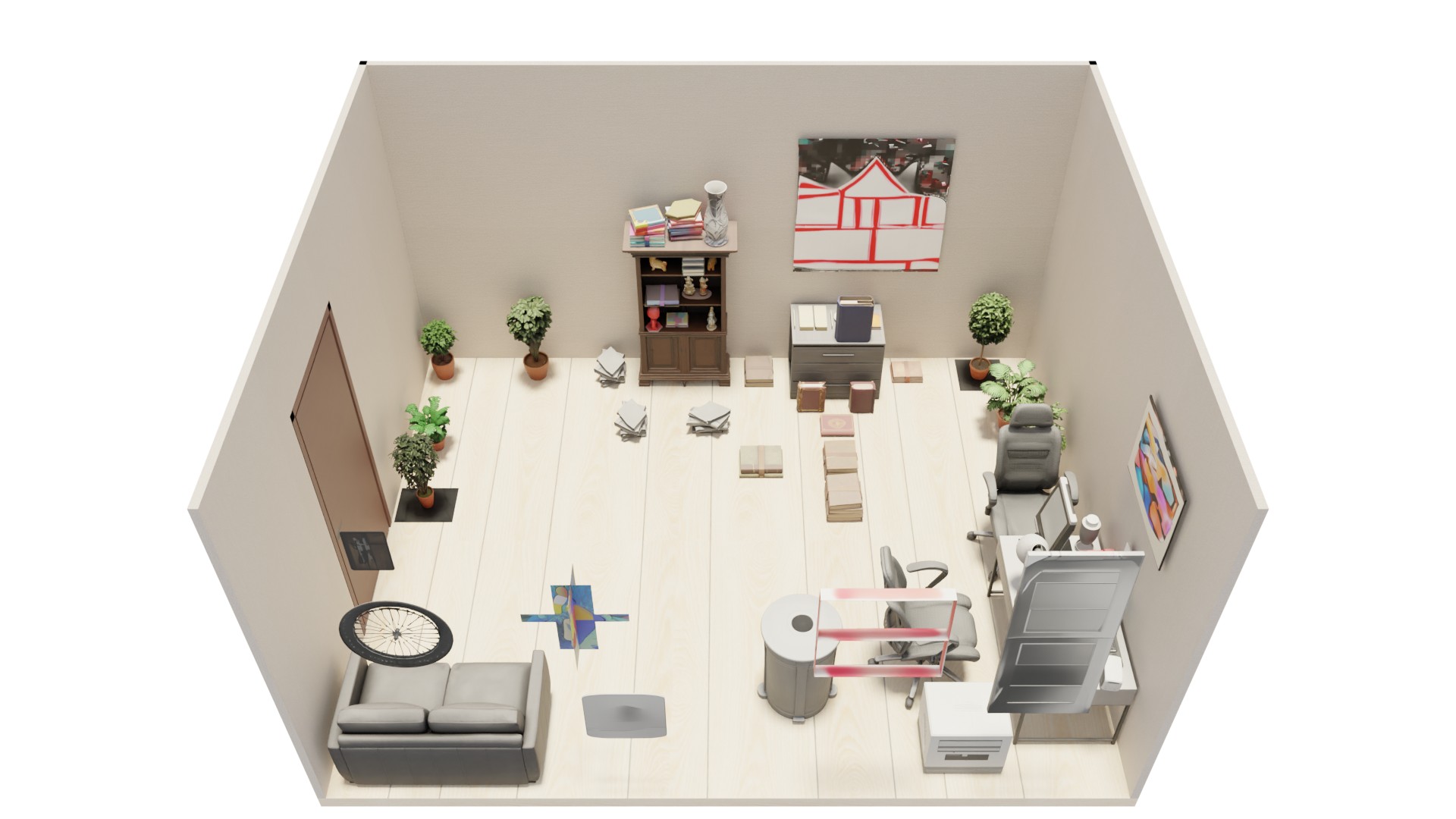}} &
\adjustbox{valign=c}{\includegraphics[width=0.35\linewidth,trim=120 0 120 0, clip]{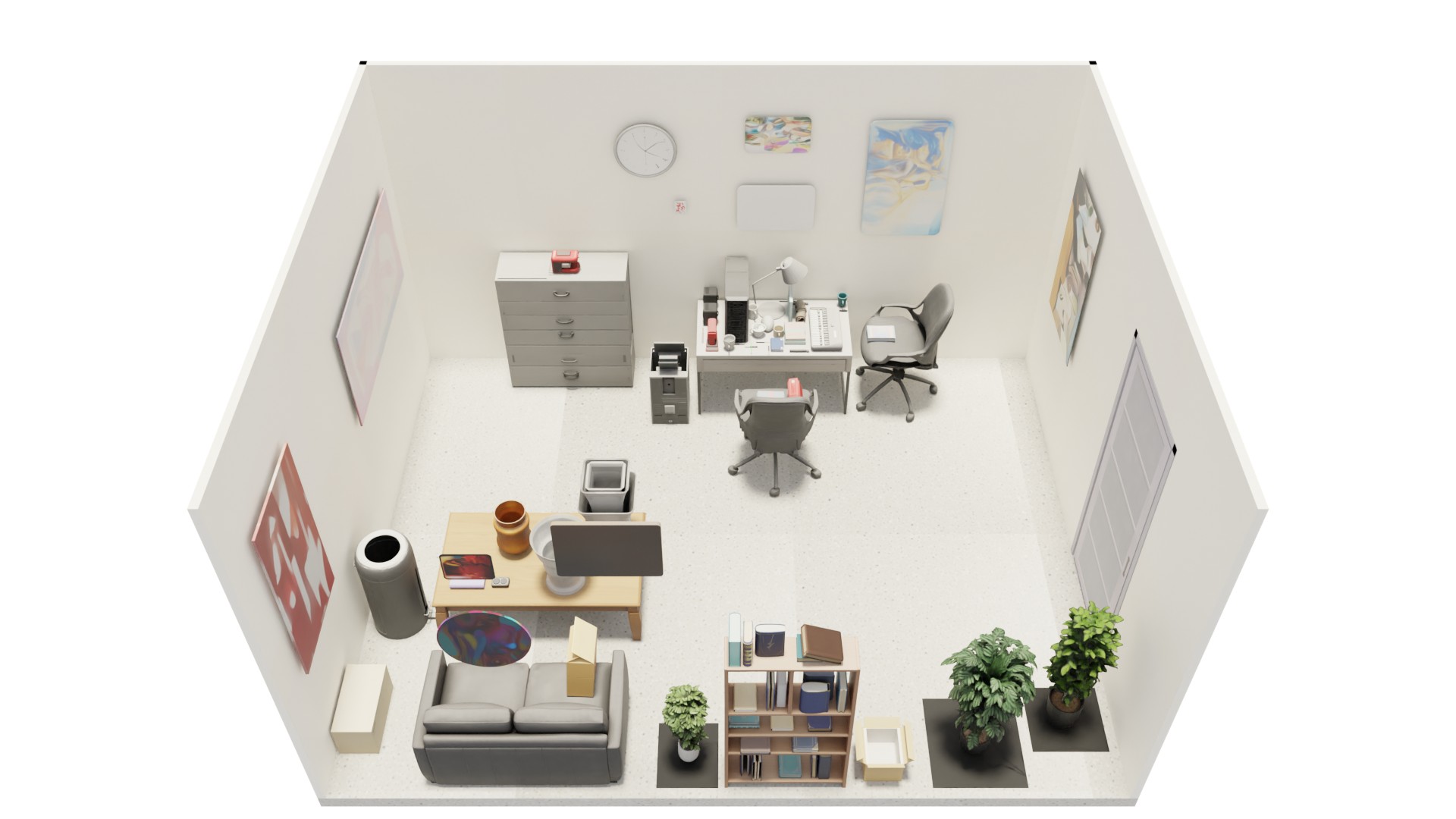}} &
\adjustbox{valign=c}{\includegraphics[width=0.35\linewidth,trim=120 0 120 0, clip]{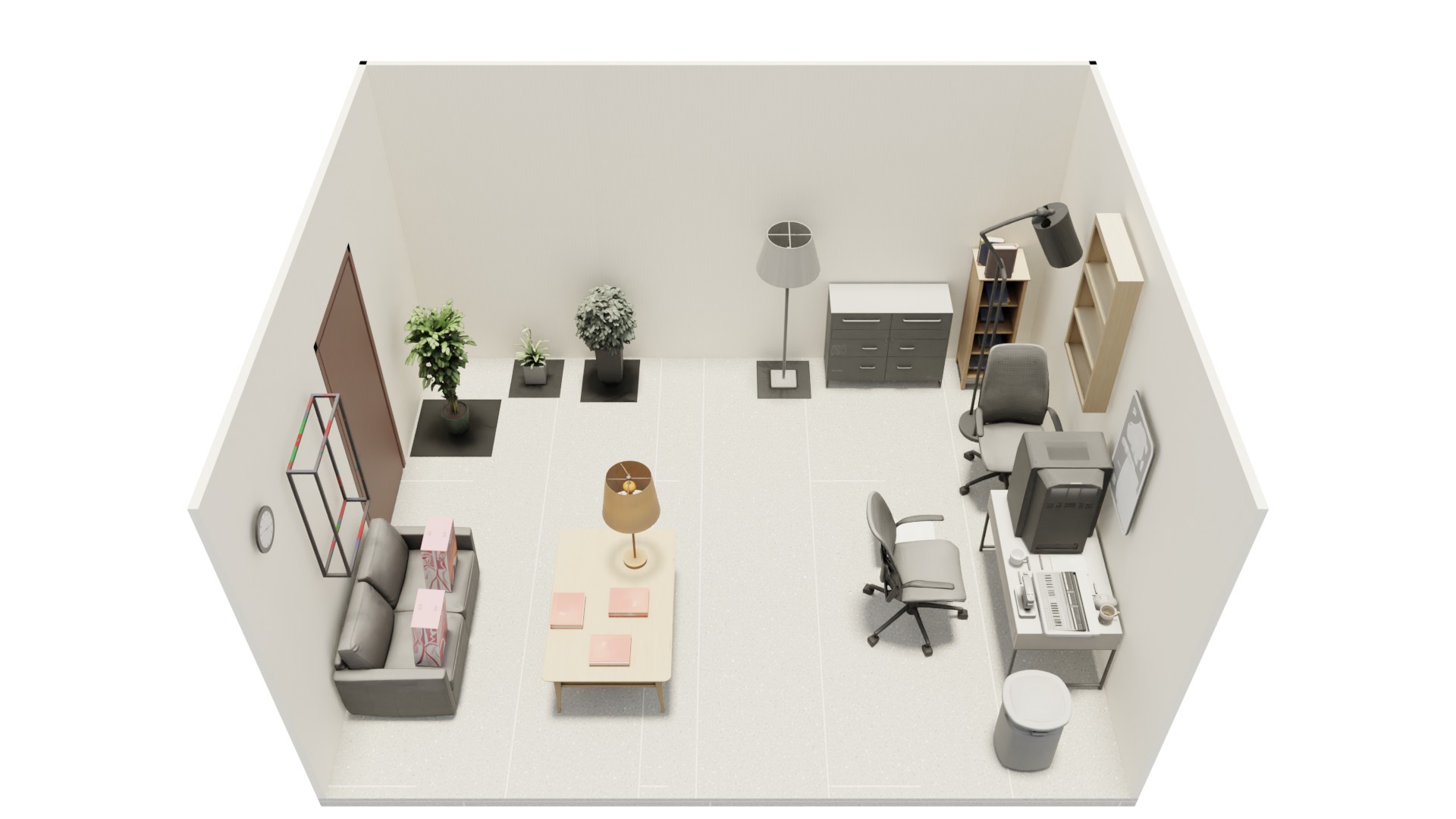}} \\

\adjustbox{valign=c}{\includegraphics[width=0.35\linewidth,trim=120 0 120 0, clip]{figs/object_aug/meeting_base.jpg}} &
\adjustbox{valign=c}{\includegraphics[width=0.35\linewidth,trim=120 0 120 0, clip]{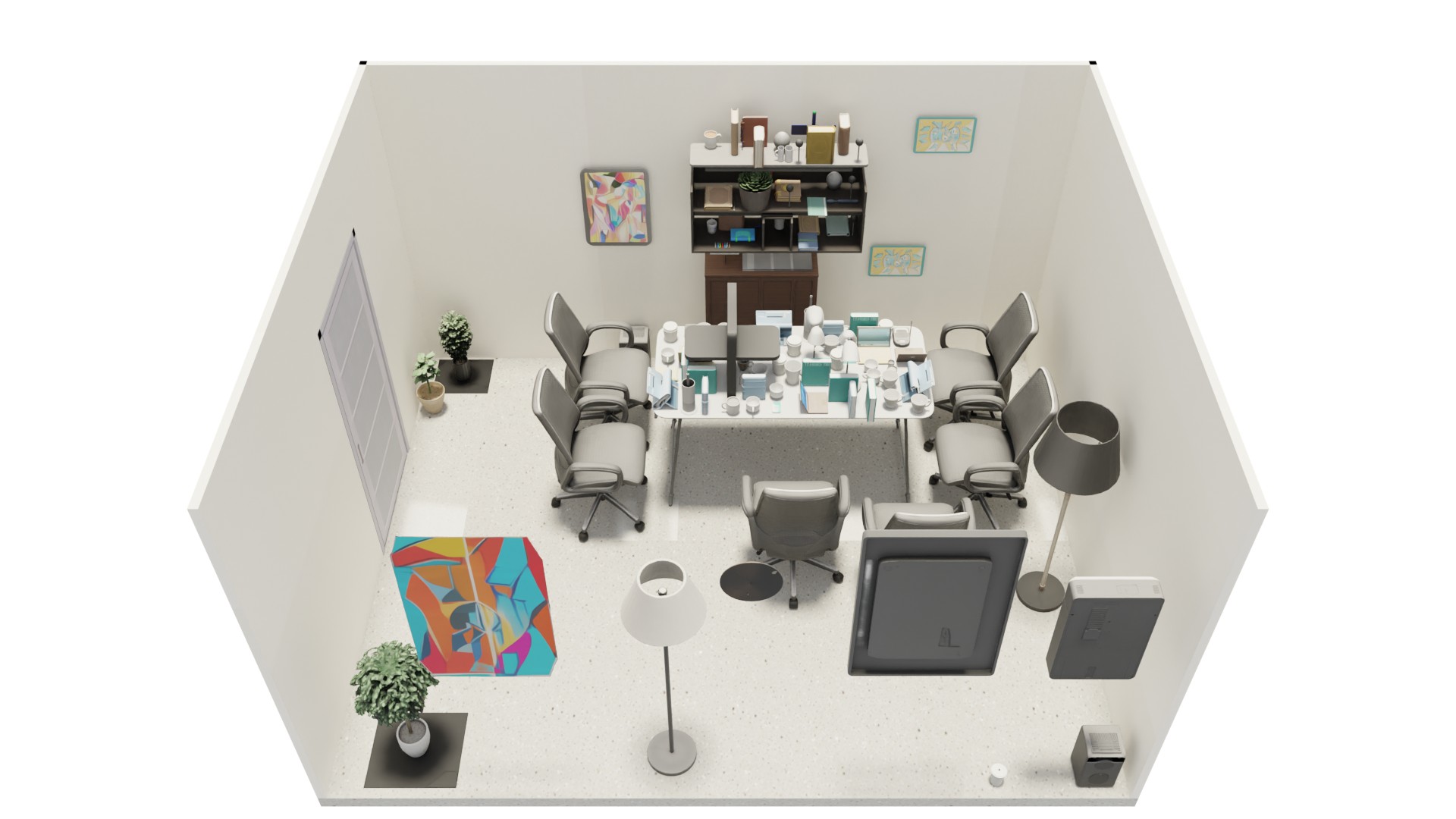}} &
\adjustbox{valign=c}{\includegraphics[width=0.35\linewidth,trim=120 0 120 0, clip]{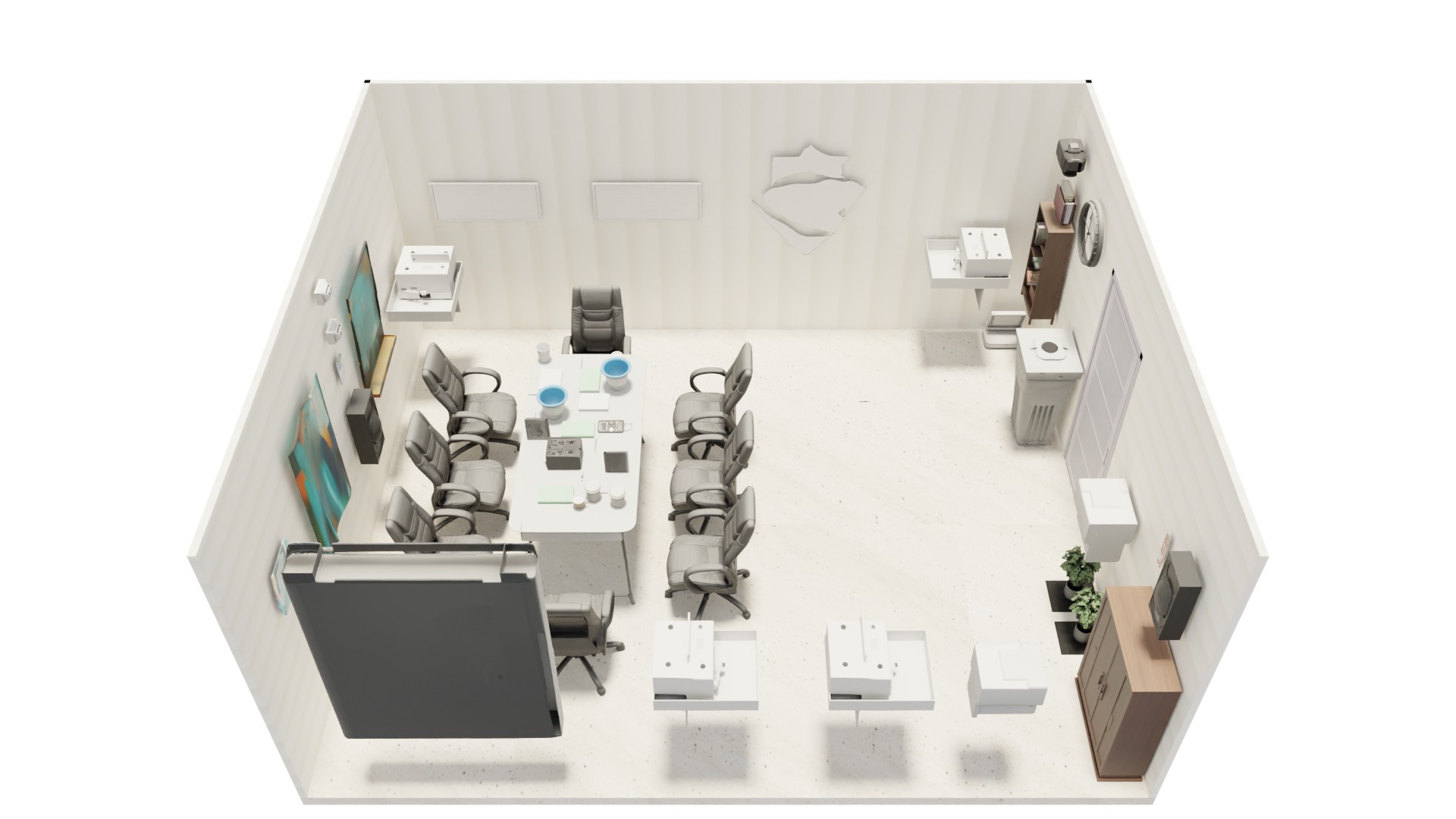}} &
\adjustbox{valign=c}{\includegraphics[width=0.35\linewidth,trim=120 0 120 0, clip]{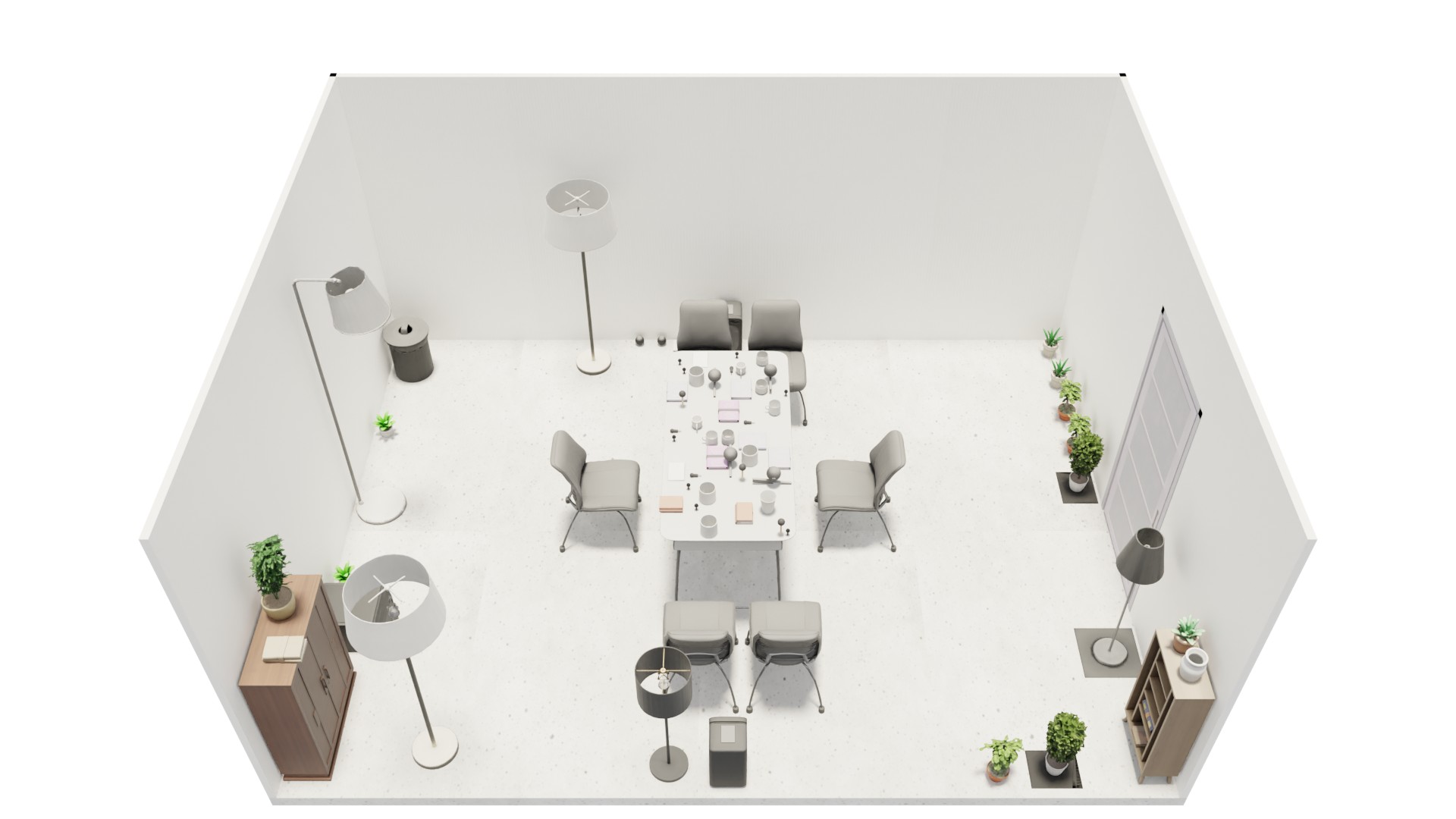}} \\
Base Scene & Aug. Scene 1 &  Aug. Scene 2 &  Aug. Scene 3 \\

\end{tabular}
}
\captionof{figure}{\textbf{Scene Layout-Level Augmentation.} 
Here we show more results of Scene Layout-Level augmentation, where the background scene, including room geometry and all task-irrelevant objects, is regenerated through the agent-driven scene generation. This process produces diverse scene layouts sharing the same task specification, enabling learning policies that generalize across spatial configurations.
\noindent{\textbf{Bedroom:}} We keep the objects of \textit{desk}, \textit{nightstand}, and \textit{mug on the nightstand} as the same.
\noindent{\textbf{Livingroom:}} We keep the objects of \textit{sideboard}, \textit{coffeetable}, and \textit{vase on the coffeetable} as the same.
\noindent{\textbf{Office:}} We keep the objects of \textit{sofa}, \textit{desk}, and \textit{pen on the desk} as the same.
\noindent{\textbf{Meeting room:}} We keep the objects of \textit{table}, \textit{cabinet}, and \textit{cup on the meeting table} as the same.
}
\vspace{-5mm}
\label{fig:layout_aug}
\end{table*}

\section{Additional Experiment Results}
In this section, we present additional visualization results and implementation details for both scene generation (Sec.~\ref{sec:supp-scene}) and robot learning (Sec.~\ref{sec:supp-robot}), followed by runtime analysis (Sec.~\ref{sec:supp-runtime}).
We encourage readers to visit our project webpage for additional visualization videos.

\subsection{Scene Augmentation}
\label{sec:supp-scene}
In this section, we will show more results related to our proposed scene augmentation method, including Object Category-level Augmentation and Scene Layout-level Augmentation.
\subsubsection{Object Category-level Augmentation}
For a previously generated single base scene produced by our agentic scene generation framework, we can augment the task relevant objects with different shapes and textures while keeping the same object category, so that robot can learn generalizable policy to unseen scenarios. 
Here we showcase the capability of our category augmentation method.
To illustrate the capacity of our augmentation method, we randomly select part of the objects in the scene for category augmentation. Following the method we described in the main paper method section, given the text description of the selected object from the generation stage, we employ an LLM-based text augmentation to produce variations in geometry and texture (\eg, shape, color, material, or finish) while maintaining the original object category. We then use TRELLIS \cite{trellis} to synthesize corresponding 3D assets from these augmented descriptions, which are placed into the scene to enrich visual and physical diversity across instances.
We show the object category-level augmentation results in different scene categories in Fig. \ref{fig:cat_aug}.

\subsubsection{Scene Layout-level Augmentation}
In addition to object category-level augmentation, to diversify the background environment as well besides the task-relevant objects, we proposed scene layout-level augmentation. The background scene, including room geometry and all task-irrelevant objects, is regenerated through the agent-driven scene generation, while we keep the task-relevant objects unchanged but repositioned according to the new scene layouts.
This process produces diverse scene layouts sharing the same task specification, enabling learning policies that generalize across spatial configurations.
We show the scene layout-level augmentation results in different scene categories in Fig. \ref{fig:layout_aug}.

\subsection{Robot Action Generation and Policy Inference}
\label{sec:supp-robot}
\subsubsection{Robot Action Generation}
We also showcase our parallel robot action generation process with the scaled training data in our project webpage. The collected robot action data is composed of robot end-effector poses with multiple camera views.
For Pick-and-Place task, the camera views input includes 3 perspective cameras located at left, right, and wrist of the Franka Emika Panda robot.
For Mobile Manipulation task, the camera views input includes 5 cameras.
Three of them are perspective cameras located at left, right, and wrist of the composed mobile manipulator robot.
The extra two cameras are fisheye cameras located on top of the robot, looking front and back separately for navigation purposes.
We input both RGB and depth from each view with the resolution of $128\times128$ to the policy network \cite{chi2024diffusionpolicyvisuomotorpolicy}.
Parallelized action generation is performed with 8 environments for Pick-and-Place task and 2 environments for Mobile Manipulation task in parallel per GPU in the IsaacSim \cite{isaacsim} simulator.
\subsubsection{Policy Inference}
For policy inference, we feed the trained policy network with the observed camera views and execute the inferred robot actions.
We showcase the policy inference demos in our project webpage, for both the Pick-and-Place task and the Mobile Manipulation task.
Failure cases are mostly due to the randomness of the policy inference, far-away objects, or difficult grasp pose location which is hard to reach and grasp by the robot.

\subsection{Runtime Analysis}
\label{sec:supp-runtime}

\paragraph{Scene Generation} Generation time for a single scene with \model varies according to user demands. The most time-consuming components are object generation and simulation-in-the-loop validation for physical stability, while agent reasoning and LLM/VLM inference are faster.
Object generation with TRELLIS \cite{trellis} takes 15 seconds per object. We parallelize this across 8 GPUs, reducing average generation time to 2-3 seconds per object. Simulation with Isaac Sim \cite{isaacsim} takes 1-2 seconds per placement candidate for stability validation.
To minimize simulation overhead, we simulate once after placing all floor and wall objects, then remove any unstable objects. This significantly improves efficiency by consolidating multiple simulations into one. For on-top objects, however, we simulate each placement individually since these objects are typically smaller and more prone to instability. We adopt an early-stopping strategy that accepts the first stable, collision-free placement validated by Isaac Sim \cite{isaacsim}. This greatly reduces computation time, as we typically find stable placements within a few trials despite having 30-50 candidate locations.
Overall, while generation time spans a broad range, a scene with 20 objects takes approximately 10 minutes, with time scaling linearly for additional objects.

\paragraph{Action Generation} 
Using Isaac Lab \cite{isaaclab} as our simulation platform, we leverage its parallelism to significantly reduce data collection time. For Pick-and-Place tasks, motion planning for each demonstration takes 8-10 seconds per environment when run sequentially. By parallelizing across 8 environments, the total planning time increases to 15-20 seconds, but this yields an average of 2-3 seconds per demonstration. For Mobile Manipulation tasks, which involve longer trajectories, the parallelized per-demonstration time is 8-10 seconds.
Our simulations scale readily across GPU clusters, enabling proportional speedup in data generation based on available GPU resources. This scalability facilitates efficient collection of large-scale datasets for training generalizable policies.

\paragraph{Policy Training}
We train a diffusion policy \cite{chi2024diffusionpolicyvisuomotorpolicy} using the Robomimic \cite{robomimic} framework, which takes several hours to converge on our robot action data. Note that diffusion policy is one approach to convert our generated actions into an executable policy, chosen here for its simplicity. Other methods, such as fine-tuning a VLA, may prove more efficient and we leave for future work.

\section{Additional Implementation Details}
In this section, we will elaborate the implementation details of our proposed agent-driven scene generation and robot action generation as well as policy learning.

\subsection{Scene Generation}
\subsubsection{Overview}
Our scene generation framework uses a carefully designed agent-driven system. The \model operates under the Model Context Protocol (MCP)~\cite{model_context_protocol}, a standardized protocol for seamless interaction with external tools~\cite{model_context_protocol}.
For the agent to understand each tool, we provide descriptions that include the tool's function, input argument types (specified as Python strings), and output format. We formulate all tool outputs as JSON dictionary strings, which the agent can easily parse and interpret.
At each iteration, the agent either specifies the next tool and its input arguments, or indicates that scene generation is complete. The server executes the requested operation and returns the result, which the agent uses as feedback to determine the next action. This setup enables adaptive, tool-driven scene generation without hard-coded logic.
For the sake of spaces in supplementary, we will include all the prompts we used in our code release.

\subsubsection{Generators}
The scene is constructed using a set of generator tools that the agent dynamically invokes via MCP. Each generator is an MCP tool the agent can call at each iteration. Below we describe the implementation of each generator.
\paragraph{Scene Initializer}
The scene initializer generates the floor plan and materials for the floor and walls. Floor plan generation proceeds in two steps. First, we prompt the LLM with the scene description, which returns the room types and sizes. For multi-room layouts, a second step generates connectivity between rooms and places connecting doors.
We generate materials from text descriptions of the floor and walls provided by the LLM using Matfuse \cite{matfuse}. In the supplementary material, we also use Flux.1-dev \cite{flux2024, labs2025flux1kontextflowmatching} to generate more detailed textures. We then assemble the floor plan with the generated materials to create the scene layout.

\paragraph{Asset Placer}
The asset placer takes a text string describing placement requirements for objects, generates them with TRELLIS~\cite{trellis}, and places them into the 3D scene.
After the text-to-3D inference with TRELLIS \cite{trellis}, we perform a series of mesh post-processing, including decimation, non-manifold correction, and hole filling, to ensure the watertight property of each generated mesh.
After asset generation, we use a VLM to estimate physical properties, including height, mass, and metallic/roughness values.
For each object, an LLM analyzes the input conditions and categorizes the placement as floor, wall, or on-top.
The placement logic varies by category.
For floor objects, we first use the LLM to generate constraints, including global position in the room and relative position/orientation to existing objects. These constraints guide our placement scoring.
We sample available positions using a grid-based approach, then apply depth-first search to find optimal positions and orientations while checking for collisions.
For wall objects, we similarly use depth-first search with grid-based sampling and collision checking against both floor and wall objects.
For on-top placement, unlike \cite{yang2024holodeck} which supports only single-layer relationships, our method enables multi-layer scene graphs.
We sample available locations by computing surface normals on the supporting object's mesh, selecting faces with normals close to the room's up axis.
This allows placement on shelves and surfaces, not just object tops as in \cite{yang2024holodeck}.
After collision checking, we obtain multiple candidate locations.
Finally, we validate stability using Isaac Sim \cite{isaacsim}.
In the simulation, all the wall objects are set as static since they are attached on the wall.
We simulate each candidate placement—if unstable (\eg, a pillow standing on a bed), we record the post-simulation pose and re-simulate with this adjusted pose.
If the second simulation is stable, we accept the placement; otherwise, we reject it.
If all candidates fail, the physics critic will report the failure to the agent and suggest alternatives such as smaller objects or different support surfaces.
Note that unlike \cite{yang2024holodeck}, which supports only single-iteration placement, our agentic framework enables iterative, adaptive placements across multiple tool calls to fully realize the scene and satisfy user requirements.

\paragraph{Asset Mover}
The Asset Mover locates and moves objects specified in the input text string. It first uses the integrated LLM to parse the target object and its destination. The object is temporarily removed from the current room, then repositioned using the same placement logic applied to other objects. If placement fails due to insufficient space, the object is restored to its original location and the failure is reported to the agent. The agent can then choose alternative actions, such as moving the object elsewhere or removing other objects to free up space. Movement instructions typically come from the visual critic.

\paragraph{Asset Remover}
The Asset Remover deletes objects described in the input text, using LLM reasoning to locate them in the scene. It is typically called when the critic provides feedback to remove an object.

\subsubsection{Critics}

\paragraph{Visual Critic}
The visual critic evaluates the current scene configuration and suggests new objects to place or adjustments to existing placements. When proposing new objects, it considers several factors. First, it identifies natural object combinations—for example, placing chairs around tables or books on bookshelves—using example combinations we provide to guide its reasoning. Second, when the room appears sparse, it suggests background elements like floor plants or decorative objects to fill empty spaces. Third, it verifies whether task-relevant objects specified by the user have been placed, notifying the agent if any are missing. For adjustments to existing objects, the critic analyzes rendered images to determine which items should be moved or removed. This feedback guides the agent in selecting the most appropriate generator to invoke next.
\paragraph{Physics Critic}
The physics critic operates at every stage of each generator, explicitly validating scene stability after object operations—addition, movement, and removal. During physical simulation, wall-mounted objects are treated as static since they attach to walls, while all other objects remain dynamic and respond to collisions and gravity. When operations fail due to physics instability, the physics critic detects these failures and returns them to the agent as feedback.

\subsection{Scene Augmentation}
We apply two types of augmentation to our agent-generated scenes: object category-level and scene layout-level. Implementation details are provided below.

\paragraph{Object Category-Level Augmentation}
We adopt TRELLIS \cite{trellis} to generate augmented objects given the augmented text descriptions of objects. The generated objects are then placed into the scene with the same supporting relationships as their originals, and physics validation is performed using Isaac Sim \cite{isaacsim}. Wall and floor textures are also able to be augmented during this process.

\paragraph{Scene Layout-Level Augmentation}
In layout-level augmentation, the background scene—including room geometry and all task-irrelevant objects—is regenerated through agent-driven scene generation, while previously generated task-relevant objects are preserved and reused.
To achieve this, we add a stage in the Asset Placer generator that excludes pre-generated objects from the generation list using LLM-based reasoning, preventing them from being regenerated.
This ensures that task-relevant objects are preserved while all task-irrelevant objects and the room layout are regenerated, producing diverse spatial configurations that enable policies to generalize better.
Note that scenes from layout-level augmentation can undergo further augmentation. For example, we can combine them with another round of category-level augmentation to create even greater diversity.

\subsection{Robot Action Generation}
\paragraph{Pick-and-Place} We use M2T2~\cite{yuan2023m2t2} to generate grasp pose candidates from rendered depth images. These are then transformed using the camera pose to obtain the actual 3D grasp pose in world coordinates for motion planning with inverse kinematics (IK).
Curobo~\cite{curobo} is integrated into the motion planning and IK pipeline by incorporating mesh geometries into its collision checking, ensuring feasible and stable grasp execution.
Once we have the grasp pose, we divide the grasp into multiple steps. First, we use IK to calculate the end-effector trajectory from the start pose to a position directly above the grasp pose. Next, the gripper lowers and closes to grasp the object. Finally, the gripper lifts to raise the object.
For placement, we use IK with Curobo collision avoidance. We simplify placement to a drop, which could be improved with additional motion planning steps to gently place the object down.
\paragraph{Mobile Manipulation} This task is composed of a few subtasks including object grasping, navigation, and placement. Object grasping follows the same motion planning as Pick-and-Place. For navigation, we use RRT~\cite{rrt} to plan collision-free trajectories. The algorithm explores possible paths from both start and target positions, ending when the paths meet. Collision checking is implemented using a 2D occupancy grid, which is faster and saves memory compared to 3D explicit mesh collision checking.
\paragraph{Parallelism} We leverage parallelism features in Isaac Sim and Isaac Lab~\cite{isaaclab, isaacsim} to simulate multiple environments in parallel per GPU (8 for Pick-and-Place and 2 for Mobile Manipulation). This scales easily to multiple GPUs in a cluster. We use NVIDIA L40S GPUs in our experiments.

\end{document}